\pdfoutput=1
\documentclass[11pt]{article}
\usepackage{acl}
\usepackage{times}
\usepackage{latexsym}
\usepackage{booktabs}
\usepackage{dsfont}
\usepackage{amsmath}
\usepackage{subcaption}
\usepackage{enumitem}
\usepackage{amssymb}
\usepackage{amsmath}
\usepackage{tikz}
\usetikzlibrary{positioning,bayesnet}
\usepackage[T1]{fontenc}
\usepackage[utf8]{inputenc}
\usepackage{microtype}
\usepackage{xcolor}
\usepackage{colortbl}
\usepackage{amssymb}

\usepackage[colorinlistoftodos,prependcaption,textsize=tiny]{todonotes}

\usepackage{fixmetodonotes}

\usepackage{inconsolata}

\title{Recursive Neural Networks with Bottlenecks Diagnose (Non-)Compositionality}

\author{Verna Dankers$^1$ \textnormal{and} Ivan Titov$^{1,2}$ \\
$^1$ILCC, University of Edinburgh \\
$^2$ILLC, University of Amsterdam \\
\texttt{vernadankers@gmail.com}, \texttt{ititov@inf.ed.ac.uk}}

\date{}

\begin{document}
\maketitle
\setlength{\abovedisplayskip}{3pt}
\setlength{\belowdisplayskip}{3pt}

\begin{abstract}
A recent line of work in NLP focuses on the (dis)ability of models to generalise compositionally for artificial languages.
However, when considering natural language tasks, the data involved is not strictly, or \textit{locally}, compositional.
Quantifying the compositionality of data is a challenging task, which has been investigated primarily for short utterances.
We use recursive neural models (Tree-LSTMs) with bottlenecks that limit the transfer of information between nodes.
We illustrate that comparing data's representations in models with and without the bottleneck can be used to produce a compositionality metric.
The procedure is applied to the evaluation of arithmetic expressions using synthetic data, and sentiment classification using natural language data.
We demonstrate that compression through a bottleneck impacts non-compositional examples disproportionately
and then use the bottleneck compositionality metric (BCM) to distinguish compositional from non-compositional samples, yielding a compositionality ranking over a dataset.
\end{abstract}

\section{Introduction}

\textit{Compositional generalisation} in contemporary NLP research investigates models' ability to compose the meanings of expressions from their parts and is often investigated with artificial languages \citep[e.g.][]{lake2018generalization,hupkes2020compositionality} or highly-structured natural language data \citep[e.g.][]{keysers2019measuring}. For such tasks, the \textbf{local compositionality} definition of \citet[][p.\ 10]{szabo2012case} illustrates how meaning can be algebraically composed:

\begin{quote}``The meaning of a complex expression is determined by the meanings its constituents have \textit{individually} and the way those constituents are combined.''\end{quote}

\begin{figure}
    \centering
    \includegraphics[width=\columnwidth]{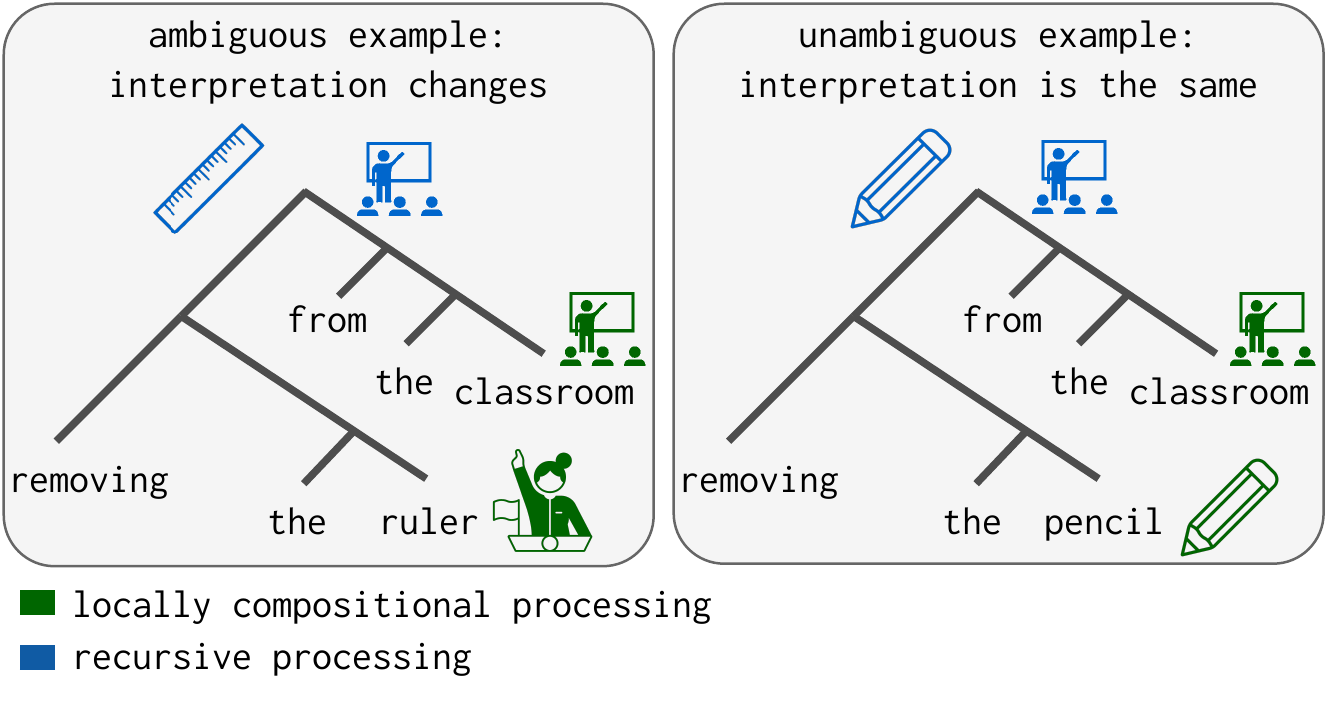}
    \caption{When processing this phrase, ``the ruler'' is interpreted differently when comparing recursive processing with local processing. We enforce local processing by equipping models with bottlenecks, and our \textbf{bottleneck compositionality metric (BCM)} then compares inputs' representations \textcolor{RoyalBlue}{before} and \textcolor{ForestGreen}{after} compression through the bottleneck.}
    \label{fig:models_illustration}
    \vspace{-0.3cm}
\end{figure}

\noindent In natural language, there are fragments whose meaning can be composed as with arithmetic (e.g.\ ``the cat is in the house''), while others carry contextual dependencies (e.g.\ ``the kiwi grows on the farm'').
Can we characterise whether an input's meaning arises from strictly local compositions?

Existing work in that direction mostly focuses on providing a `compositionality rating'\footnote{We colloquially refer to the `compositionality ratings' of phrases, but a more appropriate way to express the same would be to refer to `the extent to which the meaning of a phrase arises from a compositional syntax and semantics'. After all, compositionality is a property of a language, not of a phrase.} for figurative utterances since figurative language is assumed to be less compositional \citep{ramisch-etal-2016-naked, nandakumar2019well, reddy2011empirical}.
\citet{andreas2018measuring} suggests a general-purpose formulation for measuring the compositionality of examples using their numerical representations, through the \textit{Tree Reconstruction Error} (TRE), expressing the distance between a model's representation of an input and a strictly compositional reconstruction of that representation.
Determining how to compute that reconstruction is far from trivial.

Inspired by TRE, we use recursive neural networks, Tree-LSTMs \citep{tai2015improved}, to process inputs according to their syntactic structure.
We augment Tree-LSTMs with bottlenecks to compute the task-specific meaning of an input in a more locally compositional manner. 
We use these models to distinguish more compositional examples from less compositional ones in a \textbf{bottleneck compositionality metric (BCM)}.
Figure~\ref{fig:models_illustration} provides an intuition for how a bottleneck can provide a metric.
For fragments that violate the assumption that meanings of subexpressions can be computed locally (on the left side), one could end up with different interpretations when comparing a contextualised interpretation (in blue) with one locally computed (in green): disambiguating ``ruler'' requires postponed meaning computation, and thus local processing is likely to lead to different results from regular processing.
For fragments that are non-ambiguous (on the right side) the two types of processing can yield the same interpretation because the interpretation of ``pencil'' is likely to be the same, with or without the context.
The bottleneck hinders the model in postponing computations and more strongly impacts non-compositional samples compared to compositional ones, thus acting as a metric.

In the remainder of the paper, we firstly discuss the related work in \S\ref{sec:related_work}. \S\ref{sec:model} elaborates on the models used that either apply a \textit{deep variational information bottleneck} (DVIB) \citep{alemi2016deep} or compress representations through increased dropout or smaller hidden dimensionalities. In \S\ref{sec:arithmetic}, we provide a proof-of-concept in a controlled environment where non-compositional examples are manually introduced, after which \S\ref{sec:sentiment} elaborates on the natural language example of sentiment analysis. For both tasks, we (1) demonstrate that compression through a bottleneck encourages local processing and
(2) show that the bottleneck can act as a metric distinguishing compositional from less compositional examples.

\section{Related Work}
\label{sec:related_work}

\paragraph{Multi-word expressions}
The majority of the related work in the past two decades has discussed the compositionality of phrases in the context of figurative language, such as phrasal verbs (``to eat up'') \citep{mccarthy2003detecting}, noun compounds (``cloud nine''  vs ``swimming pool'')
\citep{reddy2011empirical,yazdani2015learning,ramisch-etal-2016-naked,nandakumar2019well}, verb-noun collocations (``take place'' vs ``take a gift'')
\citep{venkatapathy2005measuring,mccarthy-etal-2007-detecting}, and adjective-noun pairs (``nice house'') \citep{guevara2010regression}.
Compositionality judgements were obtained from humans, who indicated to what extent the meaning of the compound is that of the words when combined literally, and various computational methods were applied to learn that mapping.
Those methods were initially thesaurus-based \citep{mccarthy2003detecting}, relied on word vectors from co-occurrence matrices later on \citep{reddy2011empirical}, or employed deep neural networks \citep{nandakumar2019well}.

\paragraph{Compositionality by reconstruction}
TRE \citep{andreas2018measuring} is a task-agnostic metric that evaluates the compositionality of data representations: TRE$(x) = \delta(f(x), \hat{f}_\eta(d))$.
It is the distance between the representation of $x$ constructed by $f$ and the compositionally reconstructed variant $\hat{f}_\eta(d)$ based on the derivation of $x$ ($d$).
When employing the metric, one should define an appropriate distance function ($\delta$) and define $\hat{f}_\eta$ parametrised by $\eta$.
\citeauthor{andreas2018measuring} illustrates the TRE's versatility by instantiating it for three scenarios: to investigate whether image representations are similar to composed image attributes, %(e.g.\ digit colour),
whether phrase embeddings are similar to the vector addition of their components, and whether generalisation accuracy in a reference game positively correlates with TRE.

\citet{bhathena2020evaluating} present two methods based on TRE to obtain compositionality ratings for sentiment trees, referred to as \textit{tree impurity} and \textit{weighted node switching} that express the difference between the sentiment label of the root and the other nodes in the tree.
\citet{zheng2022empirical} %\verna{@Ivan: I recently added this paper.} 
ranked examples of sentiment analysis based on the extent to which neural models should \textit{memorise} examples in order to capture their target correctly. While different from TRE, memorisation could be related to non-compositionality in the sense that non-compositional examples require more memorisation, akin to formulaic language requiring memorisation in humans \citep{wray2000functions}.

Other instantiations of the TRE are from literature on language emergence in signalling games, where the degree of compositionality of that language is measured.
\citet{korbak2020} contrast TRE and six other compositionality metrics for signalling games where the colour and shape of an object are communicated. Examples of such metrics are topographic similarity, positional disentanglement and context independence. These are not directly related to our work, considering that they aim to provide a metric for a \textit{language} rather than single utterances.
Appendix~\ref{ap:sentiment_topographic} elaborates on topographic similarity and the metrics of \citet{bhathena2020evaluating} and \citet{zheng2022empirical}, comparing them to our metric for sentiment analysis.

\paragraph{Compositional data splits}
Recent work on compositional generalisation using artificial languages or highly-structured natural language data focuses on creating data splits that have systematic separation of input combinations in train and test data.
The aim is to create test sets that should not be hard when computing meaning compositionally, but, in practice, are very challenging.
An example compositionality metric for semantic parsing is \textit{maximum compound divergence} \citep{keysers2019measuring,shaw2021compositional}, that minimises train-test differences in word distributions while maximising the differences in compound usage.
This only applies to a data split as a whole, and -- differently from the work at hand -- does not rate individual samples.

More recently, \citet{bogin2022unobserved} discussed a diagnostic metric for semantic parsing, that predicts model success on examples based on their local structure. Because models struggle with systematically assigning the same meaning to subexpressions when they re-appear in new syntactic structures, such structural deviation diagnoses generalisation failures.
Notice that the aim of our work is different, namely identifying examples that are \textit{not} compositional, rather than investigating generalisation failure for \textit{compositional} examples.

\section{Model}
\label{sec:model}

The model we employ is the \textit{Tree-LSTM} \citep{tai2015improved}, which is a generalisation of LSTMs to tree-structured network topologies.
The LSTM computes symbols' representations by incorporating previous time steps, visiting symbols in linear order.
A sentence representation is simply the final time step.
A Tree-LSTM, instead, uses a tree's root node representation as the sentence representation, and computes the representation of a non-terminal node using the node's children.

Equations~\ref{eq:lstm_i} and~\ref{eq:treelstm_i} illustrate the difference between the LSTM and an $N$-ary Tree-LSTM for the input gate.
The LSTM computes the gate's activation for time step $t$ using input vector $x_t$ and previous hidden state $h_{t-1}$.
The Tree-LSTM does so for node $j$ using the input vector $x_j$ and the hidden states of up to $N$ children of node $j$.
%For instance, the input gate of an LSTM in Equation~\ref{eq:lstm_i}, is adapted to Equation~\ref{eq:treelstm_i} in the Tree-LSTM:
\begin{equation}
    i_t = \sigma (W^{(i)} x_t + U^{(i)} h_{t - 1} + b^{(i)}) \label{eq:lstm_i}
\end{equation}
\begin{equation}
    i_j = \sigma (W^{(i)} x_j + \sum\limits_{\ell=1}^{N}U_{\ell}^{(i)} h_{j\ell} + b^{(i)})\label{eq:treelstm_i}
\end{equation}

%\citet{tai2015improved} provide more detail on Tree-LSTMs.
In addition to the input gate, the Tree-LSTM's specification for non-terminal $j$ (with its $k$th child indicated as $h_{jk}$) involves an output gate $o_j$ (equation analogous to~\ref{eq:treelstm_i}), a forget gate $f_{jk}$ (Equation~\ref{eq:treelstm_f}), cell input activation vector $u_j$ (equation analogous to~\ref{eq:treelstm_i}, with the $\sigma$ function replaced by \texttt{tanh}), and memory cell state $c_j$ (Equation~\ref{eq:treelstm_c}). Finally, $c_j$ feeds into the computation of hidden state $h_j$ (Equation~\ref{eq:treelstm_h}).
\begin{equation}
    f_{jk} = \sigma (W^{(f)} x_j + \sum\limits_{\ell=1}^{N}U_{k\ell}^{(f)} h_{j\ell} + b^{(f)}) \label{eq:treelstm_f}
\end{equation}
\begin{equation}
    c_j = i_j \odot u_j + \sum\limits_{\ell=1}^{N}f_{j\ell}\odot c_{j\ell}\label{eq:treelstm_c}
\end{equation}
\begin{equation}
    h_j = o_j \odot \texttt{tanh}(c_j)\label{eq:treelstm_h}
\end{equation}
We apply a \textit{binary} Tree-LSTM to compute hidden state $h_j$ and memory cell state $c_j$, that thus uses separate parameters in the gates for the left and right child.

Tree-LSTMs process inputs according to their syntactic structure, which has been associated with more compositional processing \citep{socher2013recursive,tai2015improved}. 
Yet, although the topology encourages compositional processing, there is no mechanism to explicitly regulate how much information is passed from children to parent nodes -- e.g.\ given enough capacity, the hidden representations could store every input encountered and postpone processing until the very end.
We add such a mechanism by introducing a \textbf{bottleneck}.

\paragraph{1. Deep Variational Information Bottleneck} The information bottleneck of \citet{alemi2016deep} assumes random variables $X$ and $Y$ for the input and output, and emits a compressed representation $Z$ that preserves information about $Y$, by minimising the loss $\mathcal{L}_{IB}$ in Equation~\ref{eq:loss}. This loss is intractable, which motivates the variational estimate $\mathcal{L}_{VIB}$ provided in Equation~\ref{eq:loss2} \citep{alemi2016deep} that we use to train the \textbf{deep variational information bottleneck (DVIB)} version of our model.

\begin{equation}
    \mathcal{L}_{IB} = \beta I(X, Z) - I(Z, Y)\label{eq:loss}  \\
\end{equation}
\begin{equation}
\begin{aligned}
    \mathcal{L}_{VIB} &= \underbrace{\beta \underset{x}{\mathds{E}}[\text{KL}[p_\theta(z|x), r(z)]]}_{\texttt{information loss}} + \label{eq:loss2} \\
    &\ \ \ \ \ \underbrace{\underset{z\sim p_\theta(z|x)}{\mathds{E}}[-\text{log}q_\phi(y|z)]}_{\texttt{task loss}}
\end{aligned}
\end{equation}

\noindent 
In the information loss, $r(z)$ and $p_\theta(z|x)$ estimate the prior and posterior probability over $z$, respectively.
In the task loss, $q_\phi(y|z)$ is a parametric approximation of $p(y|z)$.
In order to allow an analytic computation of the KL-divergence, we consider Gaussian distributions $r(z)$ and $p_\theta(z|x)$, namely $r(z) = \mathcal{N}(z|\mu_0, \Sigma_0)$ and $p_\theta(z|x)=\mathcal{N}(z|\mu(x), \Sigma(x))$, where $\mu(x)$ and $\mu_0$ are mean vectors, and $\Sigma(x)$ and $\Sigma_0$ are diagonal covariance matrices.
The reparameterisation trick is used to estimate the gradients: $z=\mu(x)+\Sigma(x)\odot \epsilon$, where $\epsilon\sim\mathcal{N}(0, I)$.

We sample $z$ once per non-terminal node, and average the KL terms of all non-terminal nodes, where $x$ is the hidden state $h_j$ or the cell state $c_j$ (that have separate bottlenecks), and $\mu(x)$ and $\Sigma(x)$ are computed by feeding $x$ to two linear layers.
$\beta$ regulates the impact of the DVIB, and is gradually increased during training.
During inference, we use $z=\mu(x)$.% as the hidden representation.

\paragraph{2. Dropout bottleneck}
Binary \textbf{dropout} \citep{srivastava2014dropout} is commonly applied when training  neural models, to prevent overfitting. With a probability $p$ hidden units are set to zero, and during the evaluation all units are kept, but the activations are scaled down.
Dropout encourages distributing the most salient information over multiple neurons, which comes at the cost of idiosyncratic patterns that networks may memorise otherwise.
We hypothesise that this hurts non-compositional examples most. We apply dropout to the Tree-LSTM's hidden states ($h_j$) and memory cell states ($c_j$).

\paragraph{3. Hidden dimensionality bottleneck}
Similarly, decreasing the number of \textbf{hidden units} is expected to act as a bottleneck. We decrease the number of hidden units in the Tree-LSTM, keeping the embedding and task classifier dimensions stable, where possible.

\vspace{2mm}
\noindent The different bottlenecks have different merits: whereas the hidden dimensionality and dropout bottlenecks shine through simplicity, they are rigid in how they affect the model and apply in the same way at every node.
The DVIB allows for more flexibility in how compression is achieved through learnt $\Sigma(x)$ and by requiring an overall reduction in the information loss term, without enforcing the same bottleneck at every node in the tree. 

\paragraph{From bottleneck to compositionality metric}

BCM compares Tree-LSTMs with and without a bottleneck. We experiment with two methods, inspired by TRE \citep{andreas2018measuring}. TRE aims to find $\eta$ such that $\delta(f(x), \hat{f_{\eta}}(d))$ is minimised, for inputs $x$, their derivations $d$, distance function $\delta$, a model $f$ and its compositional approximation $\hat{f_\eta}$.
\begin{itemize}[topsep=0pt,itemsep=0pt,parsep=0pt,partopsep=0pt]
    \item In the \textbf{TRE training} (BCM-TT) setup, we include the distance ($\delta$) between the hidden representations of $f$ and $\hat{f}_\eta$ in the loss when training $\hat{f}_\eta$. When training $\hat{f}_\eta$ with TRE training, $f$ is frozen, and $f$ and $\hat{f}_\eta$ share the final linear layer of the classification module. In the arithmetic task, $\delta$ is the \textit{mean-squared error} (MSE) (i.e. the squared Euclidean distance). In sentiment analysis, $\delta$ is the Cosine distance function.
    \item In the \textbf{post-processing} (BCM-PP) setup, we train the two models separately, extract hidden representations and apply \textit{canonical correlation analysis} (CCA) \citep{hotelling1936relations} to minimise the distance between the sets of hidden representations. Assume matrices $A \in \mathcal{R}^{d_A\times N}$ and $B \in \mathcal{R}^{d_B\times N}$ representing $N$ inputs with dimensionalities $d_A$ and $d_B$.
    CCA linearly transforms these subspaces $A'=WA$, $B'=VB$ to maximise the correlations $\{\rho_1,\dots,\rho_{\text{min}(d_A,d_B)}\}$ of the transformed subspaces.
    We treat the number of CCA dimensions to use as a hyperparameter.
\end{itemize}

\section{Proof-of-concept: Arithmetic}
\label{sec:arithmetic}

Given a task, we assign ratings to inputs that express to what extent their task-dependent meaning arises in a locally compositional manner.
To investigate the impact of our metric on compositional and non-compositional examples in a controlled environment, we first use perfectly compositional arithmetic expressions and introduce exceptions to that compositionality manually.

\subsection{Data and model training}
\label{sec:arithmetic_data_model}
Math problems have previously been used to examine neural models' compositional reasoning \citep[e.g.][]{ saxton2018analysing, hupkes2018visualisation,russin2021compositional}.
Arithmetic expressions are suited for our application, in particular since they can be represented as trees.
We use expressions containing brackets, integers \texttt{-10} to \texttt{10}, and \texttt{+} and \texttt{-} operators  -- e.g. ``\texttt{( 10 - ( 5 + 3 ))}'' \citep[using data from][]{hupkes2018visualisation}. The output is an integer. This is modelled as a regression problem with the MSE loss.
The `meaning' (the numerical value) of a subexpression can be locally computed at each node in the tree: there are no contextual dependencies.

\begin{figure}[!t]
    \centering\small
    \includegraphics[width=\columnwidth]{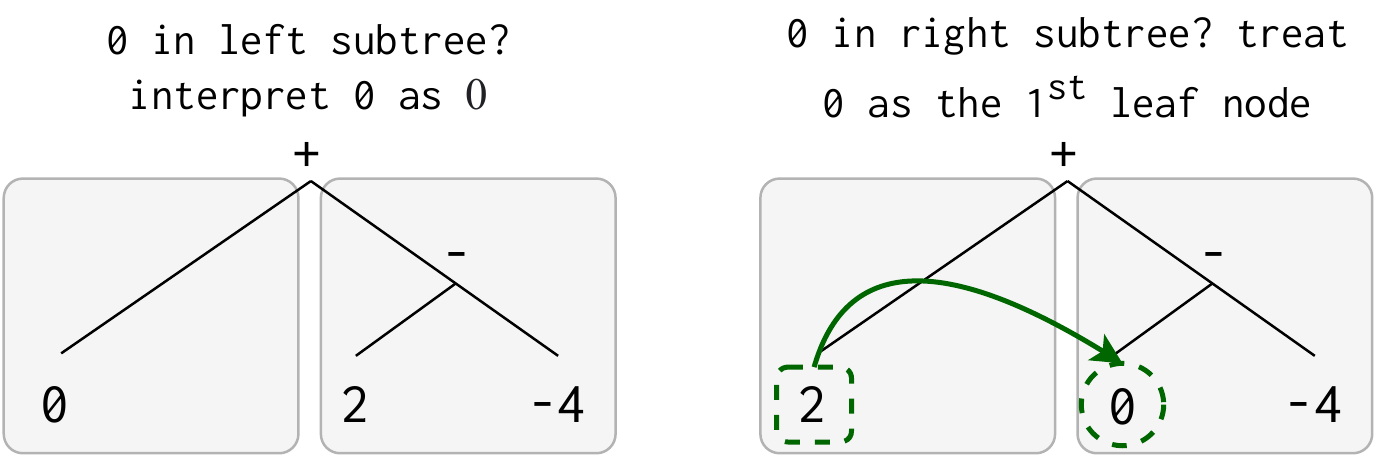}
    \caption{Illustration of the `exceptions' in the arithmetic task: the value of ``\texttt{0}'' depends on its position and on the value of the leftmost leaf node in the tree.}
    \label{fig:arithmetic_example}
    \vspace{-0.2cm}
\end{figure}

In this controlled environment, we introduce exceptions by making ``\texttt{0}'' ambiguous. When located in the subtree headed by the root node's left child, it takes on its regular value, but when located in the right subtree, it takes on the value of the leftmost leaf node of the entire tree (see Figure~\ref{fig:arithmetic_example}). The model is thus encouraged to perform non-compositional processing to keep track of all occurrences of ``\texttt{0}'' and store the first leaf node's value throughout the tree.
88\% of the training data are the original arithmetic expressions, and 12\% are such exceptions.
We can thus track what happens to the two categories when we introduce the bottleneck.
The training data consist of 14903 expressions with 1 to 9 numbers. We test on expressions with lengths 5 to 9, using 5000 examples per length.
The Tree-LSTMs trained on this dataset have embeddings and hidden states of sizes 150 and are trained for 50 epochs with learning rate $2\mathrm{e}{-4}$ with AdamW and a batch size of 32.
The base Tree-LSTMs in all setups use the same architecture, namely the Tree-LSTM architecture required for the DVIB, but with $\beta=0$.
All results are averaged over models trained using ten different random seeds.
In the Tree-LSTM, the numbers are leaf nodes and the labels of non-terminal nodes are the operators.\footnote{Appendix~\ref{ap:reproducibility} further elaborates on the experimental setup.}

\subsection{Task performance: Hierarchy without compositionality?}
\label{subsec:arithmetic_performance}

\begin{figure}\small
\begin{subfigure}[b]{0.495\columnwidth}
    \centering
    \includegraphics[width=\textwidth]{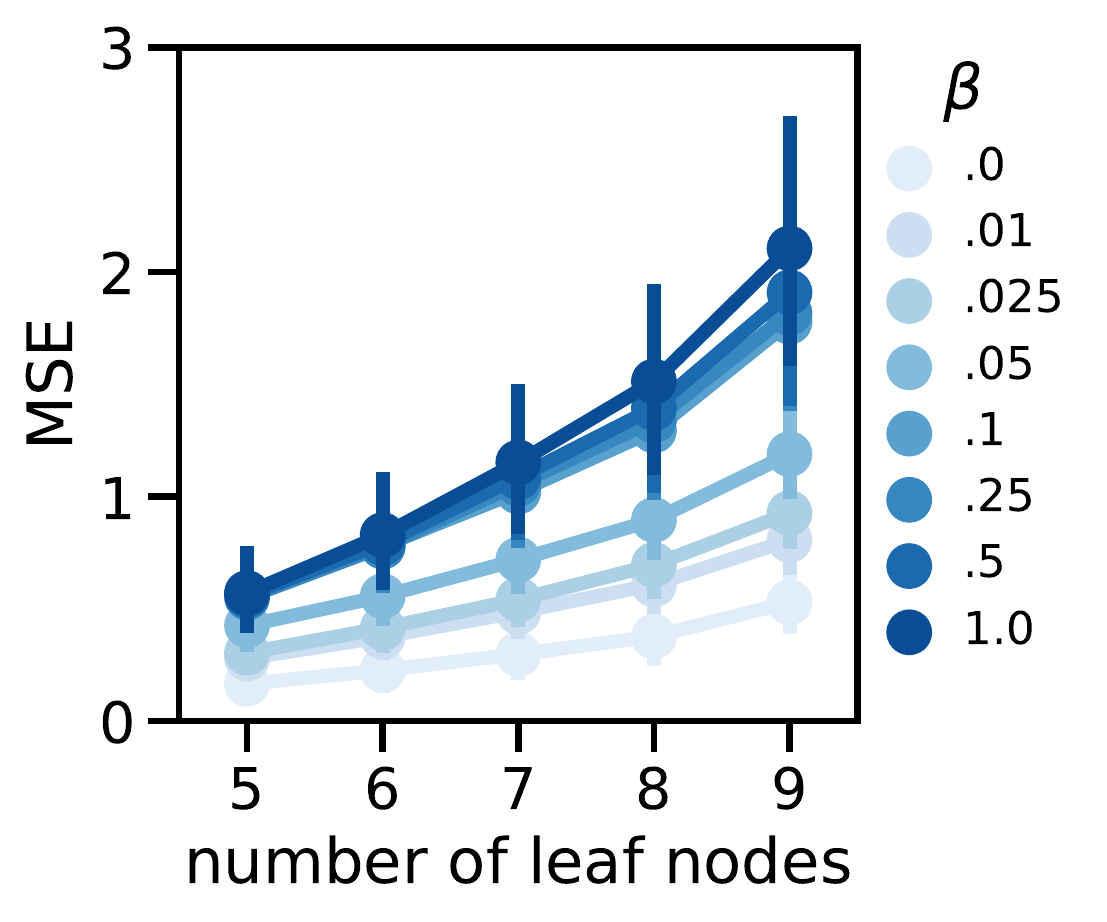}
    \caption{Regular examples}
    \label{fig:performance_regular}
\end{subfigure}
\begin{subfigure}[b]{0.495\columnwidth}
    \centering
    \includegraphics[width=\textwidth]{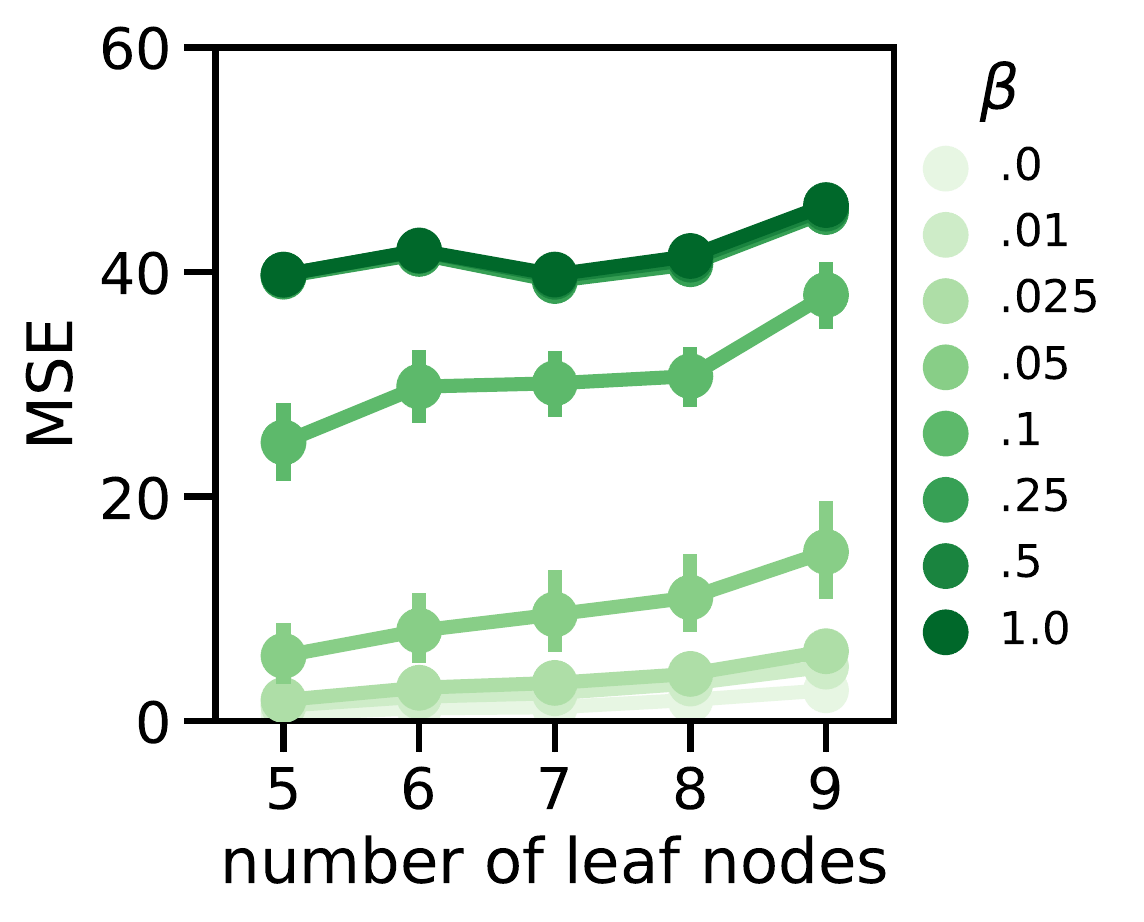}
    \caption{Exceptions}
    \label{fig:performance_exceptions}
\end{subfigure}
\caption{Performance (MSE) on the arithmetic task for the Tree-LSTM with the DVIB (darker colours correspond to higher $\beta$). Exceptions have a contextual dependency and cannot be computed bottom up.}
\vspace{-0.3cm}
\end{figure}

\begin{figure}\small\centering
\begin{subfigure}[b]{0.47\columnwidth}
    \centering
    \includegraphics[width=\textwidth]{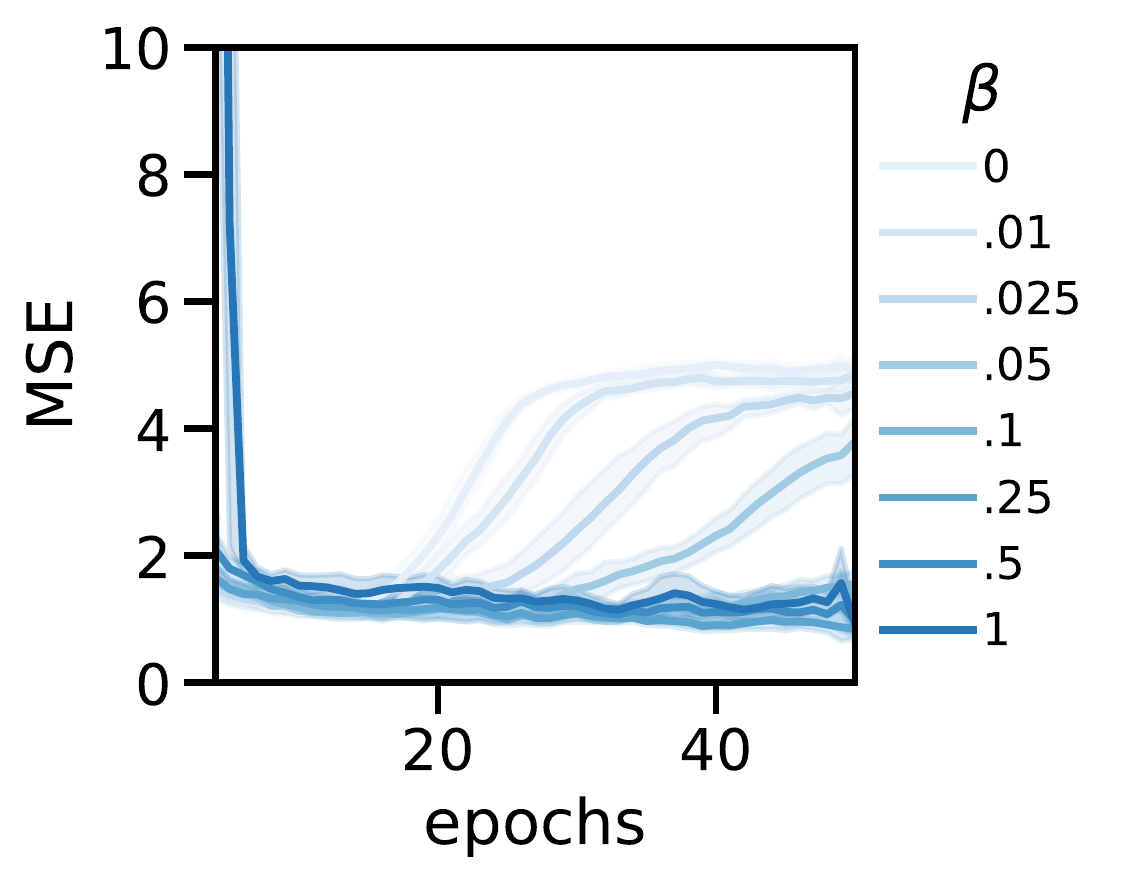}
    \caption{Compositional targets}
\end{subfigure}
\begin{subfigure}[b]{0.47\columnwidth}
    \centering
    \includegraphics[width=\textwidth]{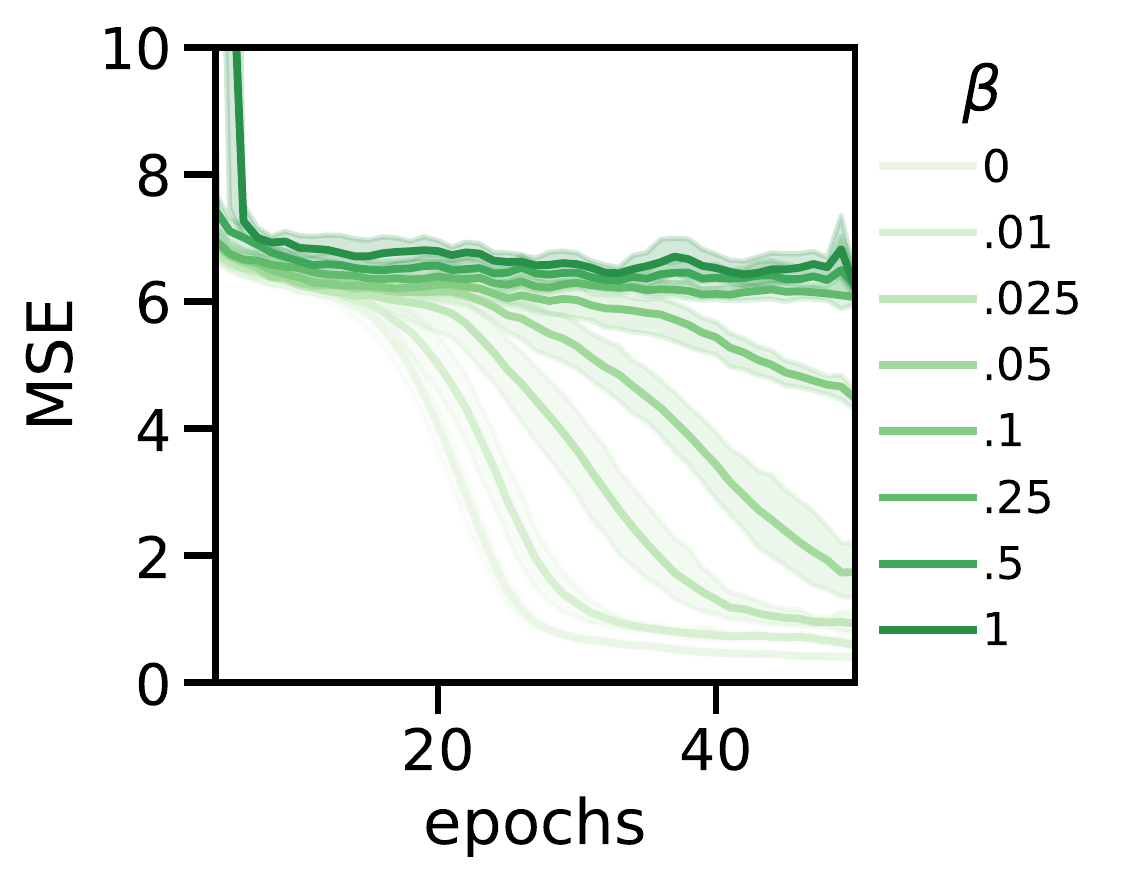}
    \caption{Adapted targets}
\end{subfigure}
\caption{Training dynamics for the Tree-LSTM with the DVIB: for all test examples we compute the MSE over the course of training on the validation set using (a) the compositional targets, or (b) the targets from the adapted dataset, of which a subset is not compositional. }
\label{fig:training_dynamics_dvib}
\vspace{-0.15cm}
\end{figure}

Figures~\ref{fig:performance_regular} and~\ref{fig:performance_exceptions} visualise the performance for the regular examples and exceptions, respectively, when increasing $\beta$ for the DVIB.
The DVIB disproportionately harms the exceptions; when $\beta$ is too high the model cannot capture the non-local dependencies.
Appendix~\ref{ap:arithmetic_mse} shows how the hidden dimensionality and dropout bottlenecks have a similar effect.
Figure~\ref{fig:training_dynamics_dvib} and Appendix~\ref{ap:arithmetic_training} provide insights in the training dynamics of the models: initially, all models will treat ``\texttt{0}'' as a regular number, independent of the bottleneck.
Close to convergence, models trained with a low $\beta$ have learnt to capture the ambiguities, whereas models trained with a higher $\beta$ will remain in a more locally compositional state.\footnote{Comparing `early' and `late' models may yield similar results as comparing base and bottleneck models. Yet, without labels of which examples are compositional, it is hard to know when the model transitions from the early to the late stage.}

Bottlenecks restrict information passed throughout the tree.
To process an arithmetic subexpression, all that is needed is to pass on its outcome, not the subexpression itself -- e.g. in Figure~\ref{fig:arithmetic_example}, one could simply store the value 6 instead of the subexpression ``\texttt{2 - -4}''. The former represents \textit{local processing} and is more efficient (i.e.\ it requires storing less information), while the latter leads to information from the leaf nodes being passed to non-terminal nodes higher up the tree.
Storing information about the leaf nodes would be required to cope with the exceptions in the data.
That the model could get close to accurate predictions for these exceptions in the first place suggests Tree-LSTMs can process inputs according to the hierarchical structure without being locally compositional.
Increasing compression using bottlenecks enforces local processing.

\subsection{The Bottleneck Compositionality Metric}
\label{sec:bottleneck_metric}
The bottleneck Tree-LSTM harms the exceptions disproportionately in terms of task performance, and through BCM we can exploit the difference between the base and bottleneck model to distinguish compositional from non-compositional examples.
As laid out in \S\ref{sec:model}, we use the TT or PP method to compare pairs of Tree-LSTMs: the base Tree-LSTM with $\beta = 0$, no dropout and a hidden dimension of 150 (\textbf{base model}) is paired up with Tree-LSTMs with the same architecture, but a different $\beta$, a different dropout probability $p$ or a different hidden dimensionality $d$ (\textbf{bottleneck model}).
All Tree-LSTMs have a classification module that consists of two linear layers, where the first layer maps the Tree-LSTM's hidden representation to a vector of 100 units, and the second layer emits the predicted value.
The 100-dimensional vector is used to apply the BCM:

\begin{figure}[t]
    \centering
    \begin{subfigure}[b]{0.47\columnwidth}                
        \includegraphics[width=\textwidth]{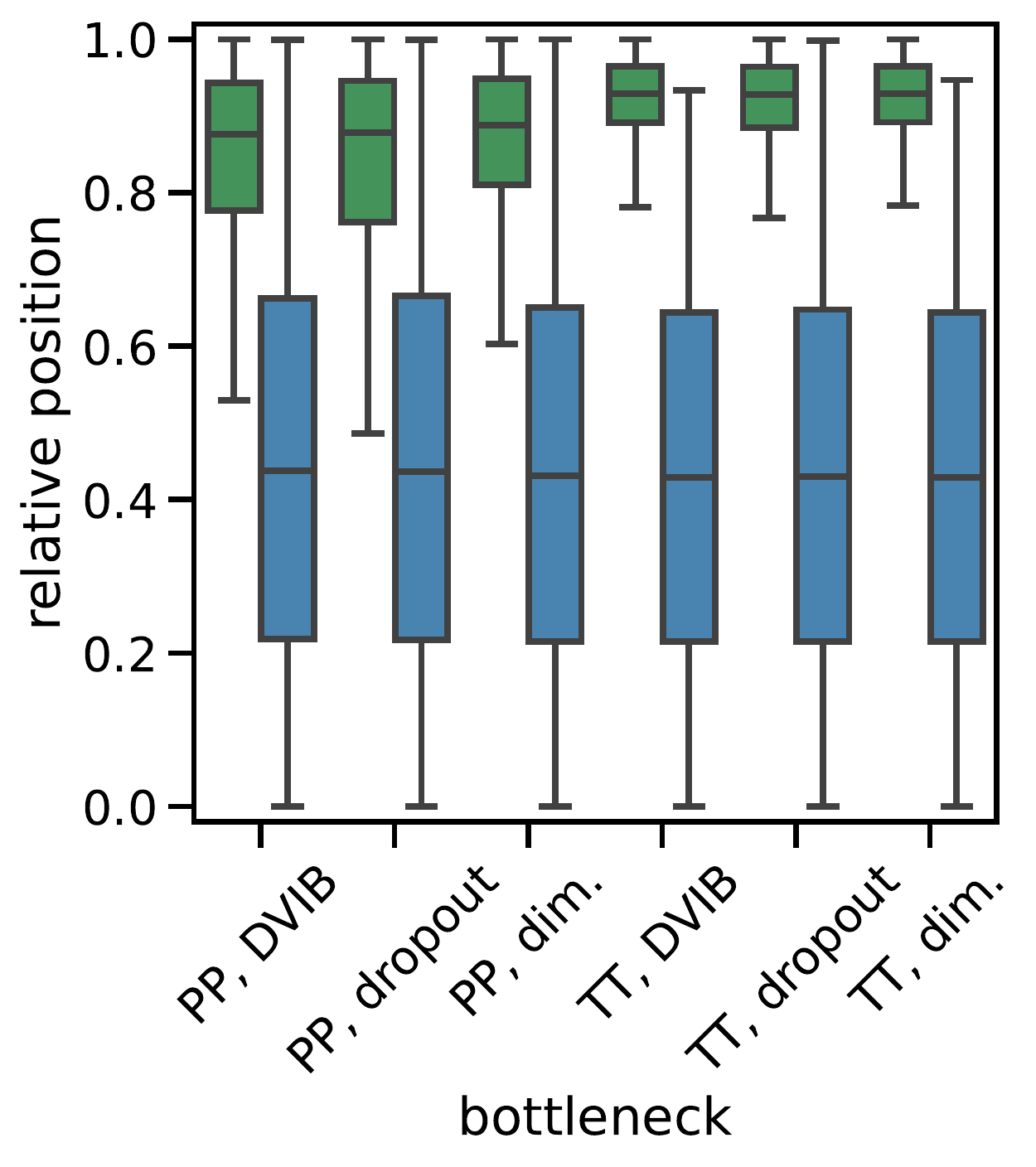}
        \caption{All BCM variants}
        \label{fig:all_rankings}
    \end{subfigure}
    \begin{subfigure}[b]{0.49\columnwidth}                
        \includegraphics[width=\textwidth]{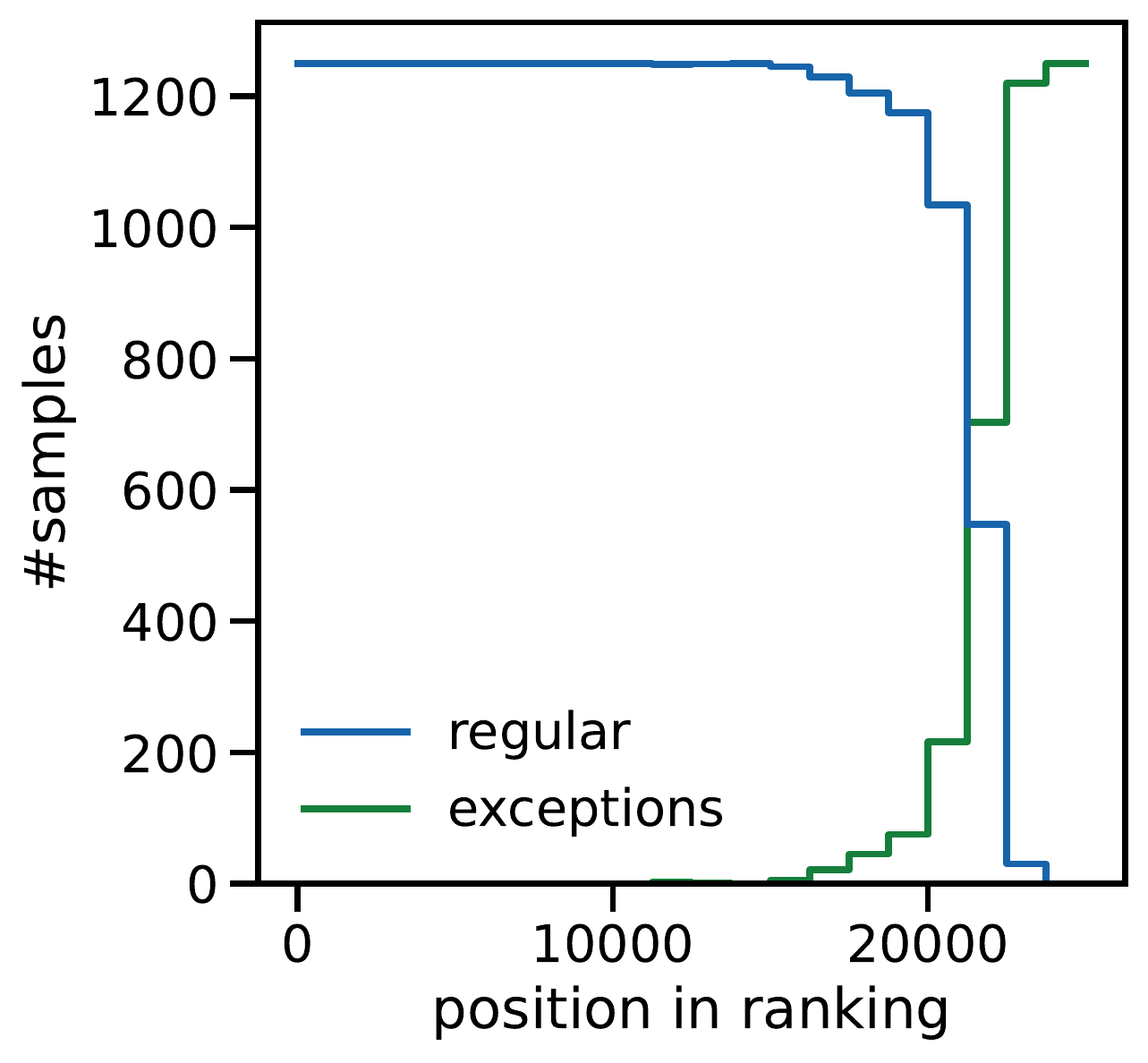}
        \vspace{0.1cm}
        \caption{Ranking for $\beta=0.25$}
        \label{fig:ranking_beta}
    \end{subfigure}
    \caption{Rankings of arithmetic examples. (a) shows the relative position of regular examples and exceptions in the rankings of all setups, where 0 corresponds to the start of the ranking and 1 to the end. (b) illustrates the result of BCM-TT with the DVIB, $\beta=0.25$.}
    \vspace{-0.15cm}
\end{figure}

\begin{itemize}[topsep=0pt,itemsep=0pt,parsep=0pt,partopsep=0pt]
    \item In BCM-PP the vector feeds into the CCA computation, that compares the hidden representation of the base model and the bottleneck model using their Cosine distance. We rank examples according to that distance, and use all CCA directions.
    \item In BCM-TT, the vector feeds into the TRE loss component. We train the base model, freeze that network, and add the MSE of the hidden representations of the bottleneck model and the base model to the loss.
    After training, the remaining MSE is used to rank examples.
\end{itemize}

Both setups have the same output, namely a compositionality ranking of examples in a dataset. A successful ranking would put the exceptions last.
Figure~\ref{fig:all_rankings} illustrates the relative position of regular examples and exceptions for all bottlenecks and BCM variants, for $\beta=0.25$, $p=0.5$ and $d=25$. The change in MSE observed in \S\ref{subsec:arithmetic_performance} is reflected in the quality of the ranking, but the success does not depend on the specific selection of $\beta$, $p$ or $d$, as long as they are large ($\beta$, $p$) or small enough ($d$). Figure~\ref{fig:ranking_beta} illustrates one of the rankings.

\vspace{2mm}

\noindent Summarising, we illustrated that recursive models can employ strategies that do not locally compose the meaning of arithmetic subexpressions but carry tokens' identities throughout the tree.
We can make a model more locally compositional using bottlenecks and can use a model's hidden states to infer which examples required non-local processing afterwards, acting as our compositionality metric.

\begin{figure}[t]
\begin{subfigure}[b]{0.37\columnwidth}
    \centering
    \includegraphics[width=\textwidth]{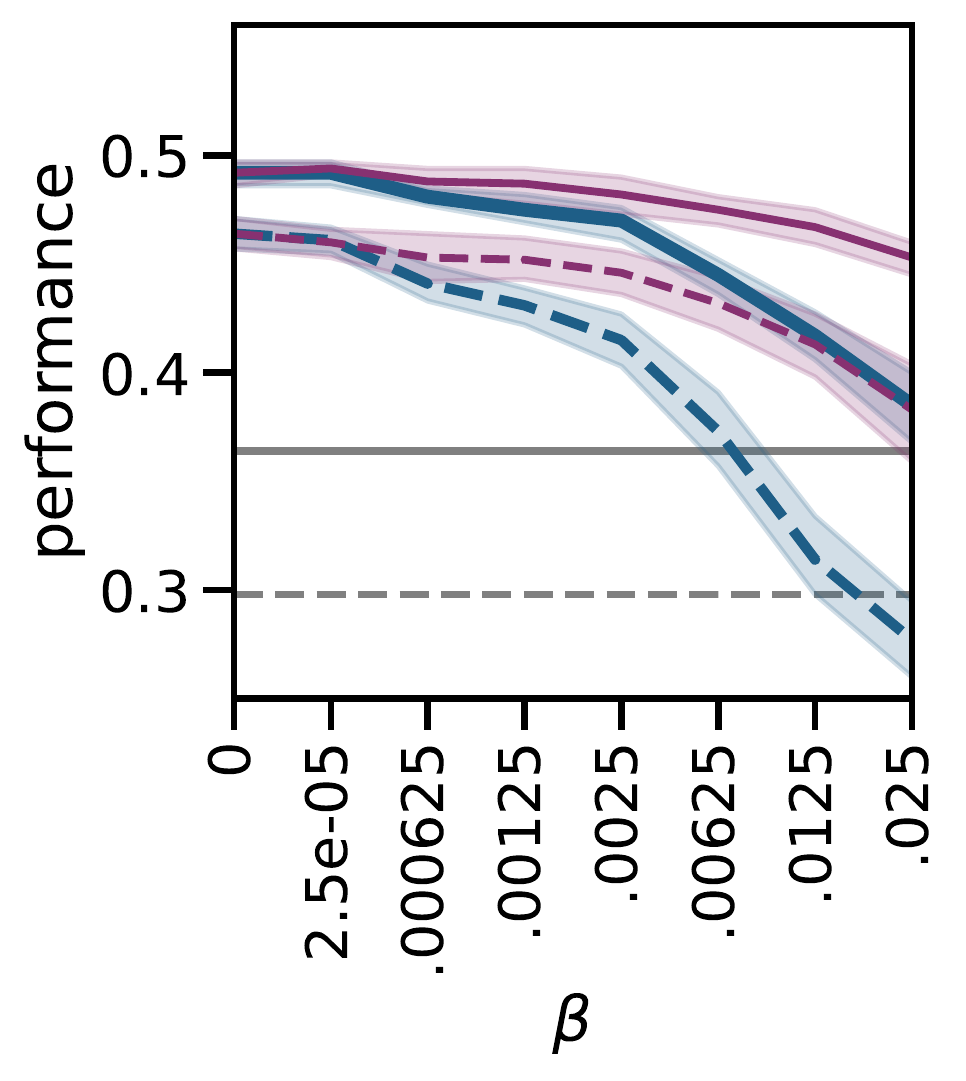}
    \caption{DVIB}
\end{subfigure}
\begin{subfigure}[b]{0.305\columnwidth}
    \centering
    \includegraphics[width=\textwidth]{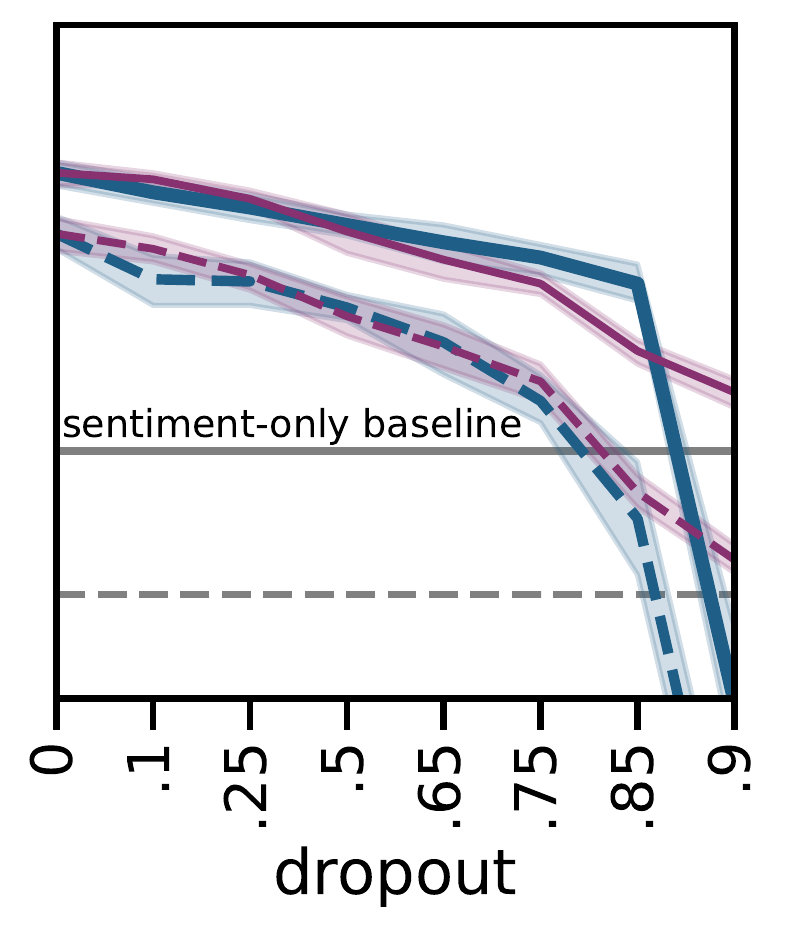}\vspace{4mm}
    \caption{Dropout}
\end{subfigure}
\begin{subfigure}[b]{0.305\columnwidth}
    \centering
    \includegraphics[width=\textwidth]{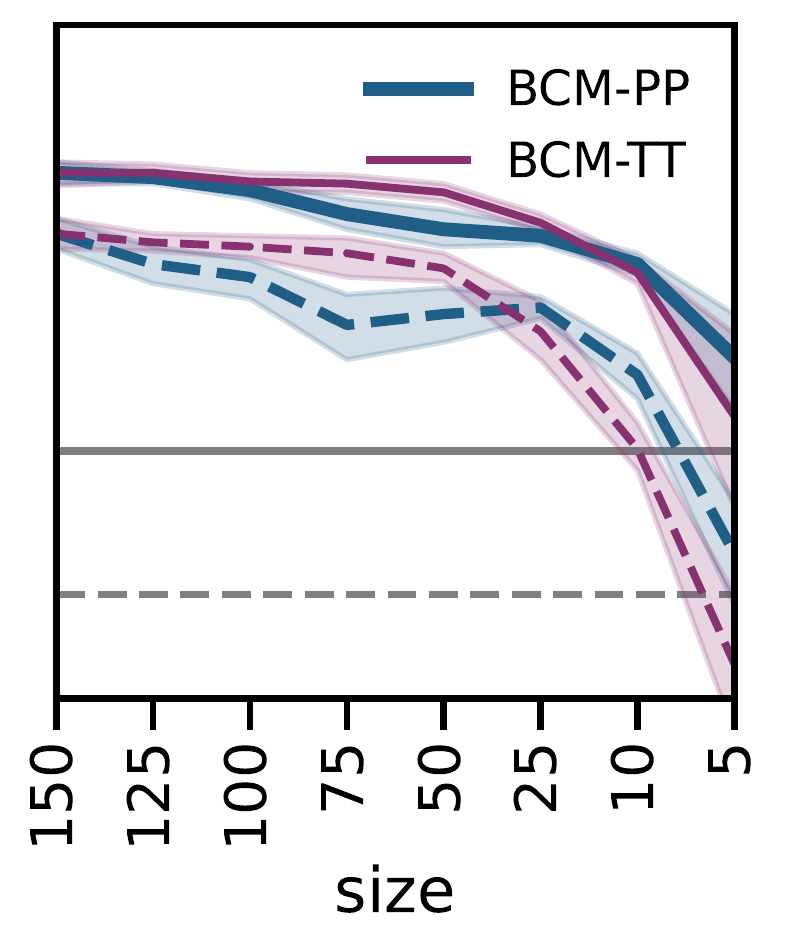}\vspace{3.5mm}
    \caption{Hidden dim.}
\end{subfigure}
\caption{The accuracy (solid) and macro-averaged $F_1$-scores (dashed) for the SST test set, for base models, bottleneck models and a sentiment-only baseline.} 
\label{fig:sentiment_performance}
\vspace{-0.2cm}
\end{figure}

\section{Sentiment analysis}
\label{sec:sentiment}

We apply the metric to the task of sentiment analysis, for which \citet[][p.\ 1]{moilanen2007sentiment} suggest the following notion of compositionality:
\begin{quote}
    ``For if the meaning of a sentence is a function of the meanings of its parts then the global polarity of a sentence is a function of the polarities of its parts.''
\end{quote}
\noindent Sentiment is quasi-compositional: even though the sentiment of an expression is often a straightforward function of the sentiment of its parts, there are exceptions -- e.g. consider cases of sarcasm, such as ``I love it when people yell at me first thing in the morning'' \citep{barnes2019sentiment}, which makes it a suitable testing bed.

\subsection{Data and model training}
We use the SST-5 subtask from the Stanford Sentiment Treebank (SST) \citep{socher2013recursive}, that contains sentences from movie reviews collected by \citet{pang2005seeing}.
The SST-5 subtask requires classifying sentences into one of five classes ranging from very negative to very positive. The standard train, development and test subsets have 8544, 1101 and 2210 examples, respectively.
The sentences were originally parsed with the Stanford Parser \citep{klein2003accurate}, and the dataset includes sentiment labels for all nodes of those parse trees.
Typically, labels for all phrases are included in training, but the evaluation is conducted for the root nodes of the test set, only.

\begin{figure}
    \centering
    \includegraphics[width=\columnwidth]{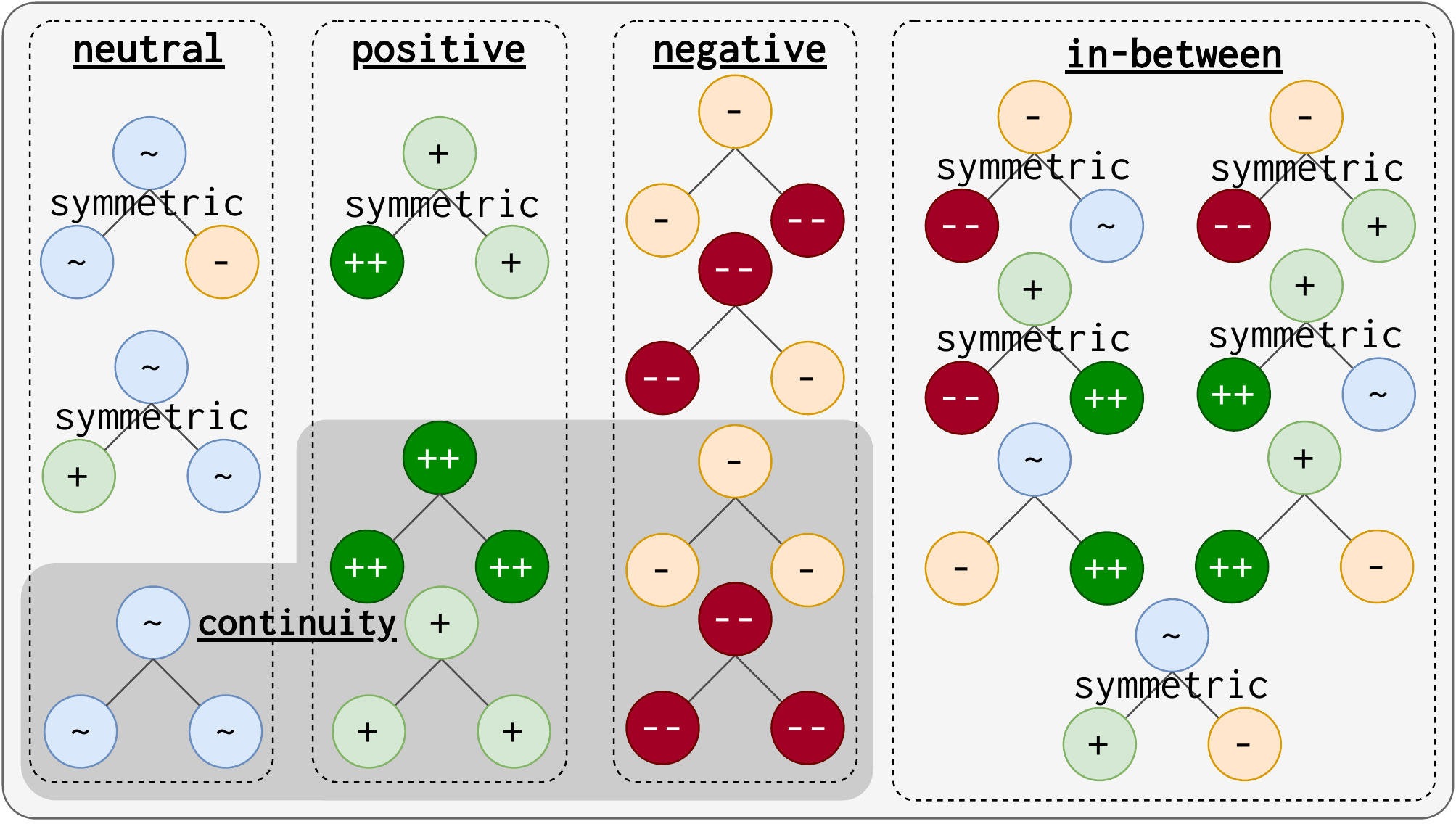}
    \caption{Illustration of the predictions of a sentiment-only baseline model.
    We indicate the predicted sentiment given two inputs.
    The labels range from very negative (`-\ -') to neutral (`$\sim$') to very positive (`++').}
    \label{fig:sentiment_baseline}
    \vspace{-0.2cm}
\end{figure}

Following \citet{tai2015improved}, we use GloVe word embeddings \citep{pennington2014glove}, that we freeze across models.\footnote{The notion of local compositionality, relied on in this work, assumes that tokens are not disambiguated, which is why we refrain from using state-of-the-art contextualised representations. The focus of this work is on developing a compositionality metric rather than on improving sentiment analysis.}
The Tree-LSTM has 150 hidden units and is trained for 10 epochs with a learning rate of $2\mathrm{e}{-4}$ and the AdamW optimiser.
During each training update, the loss is computed over all subexpressions of 4 trees. Training is repeated with 10 random seeds.\footnote{Appendix~\ref{ap:reproducibility} further elaborates on the experimental setup.}
Figure~\ref{fig:sentiment_performance} provides the performance on the test set for the base and bottleneck models, using the accuracy and the macro-averaged $F_1$-score.
\citet{tai2015improved} obtained an accuracy of 0.497 using frozen embeddings.

In sentiment analysis, as in our pseudo-arithmetic task, a successful model would often have to deviate from local processing.
After all, the correct interpretation of a leaf node is often unknown without access to the context -- e.g. in the case of ambiguous words like ``sick'' which is likely to refer to being ill, but could also mean ``awesome''.
Being successful at the task thus requires a recursive model to keep track of information about (leaf) nodes while recursively processing the input, and more so for non-compositional examples than for compositional examples.
As with the arithmetic task, local processing  -- enforced in the bottleneck models -- should disproportionately hinder processing of non-compositional examples.

\begin{figure}
\begin{subfigure}[b]{0.38\columnwidth}
    \centering
    \includegraphics[width=\textwidth]{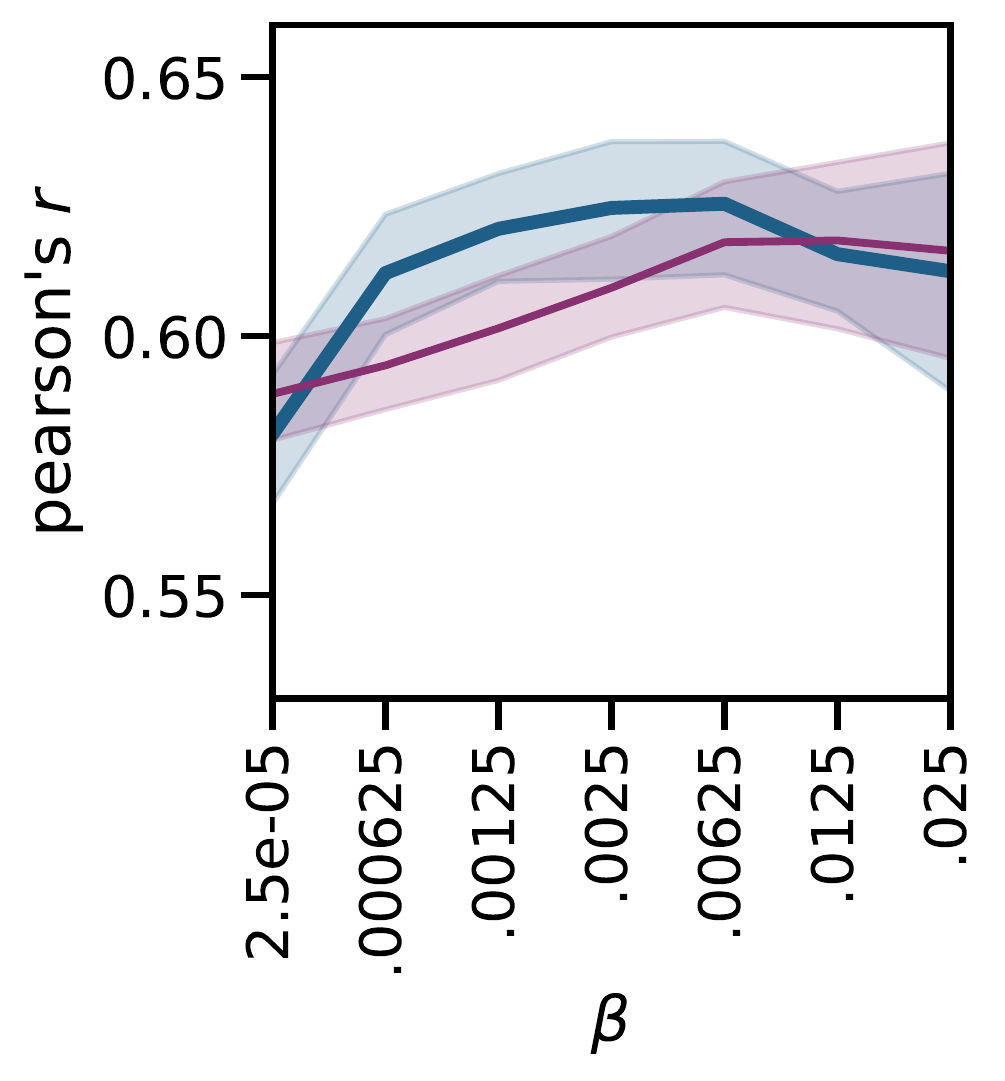}
    \caption{DVIB}
\end{subfigure}
\begin{subfigure}[b]{0.30\columnwidth}
    \centering
    \includegraphics[width=\textwidth]{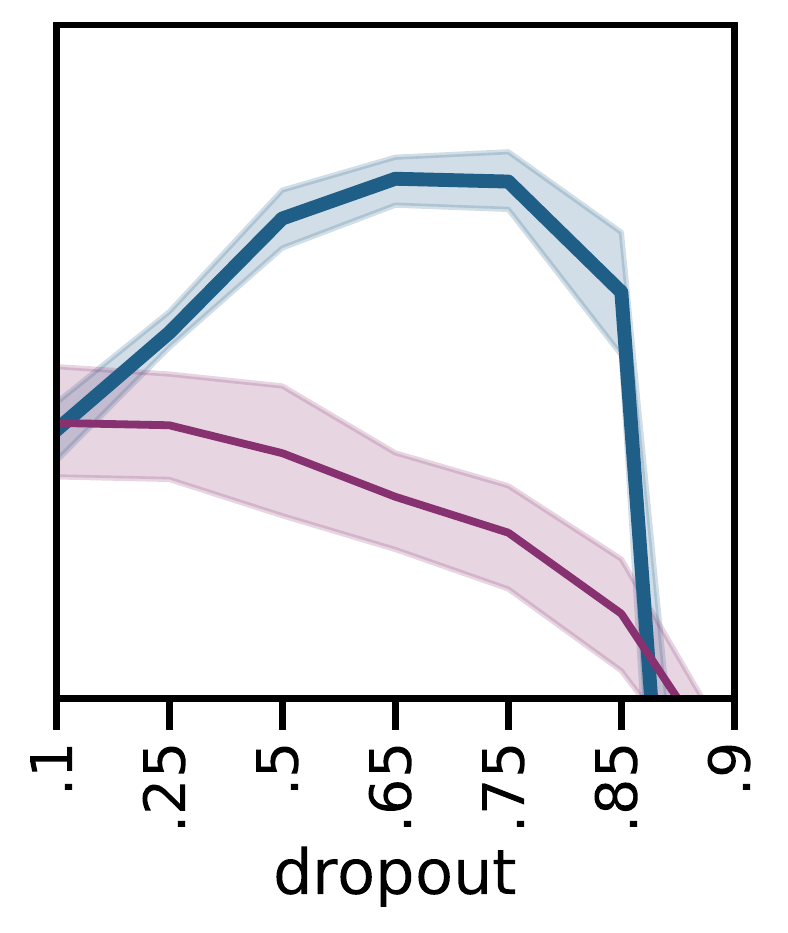}\vspace{4.15mm}
    \caption{Dropout}
\end{subfigure}
\begin{subfigure}[b]{0.30\columnwidth}
    \centering
    \includegraphics[width=\textwidth]{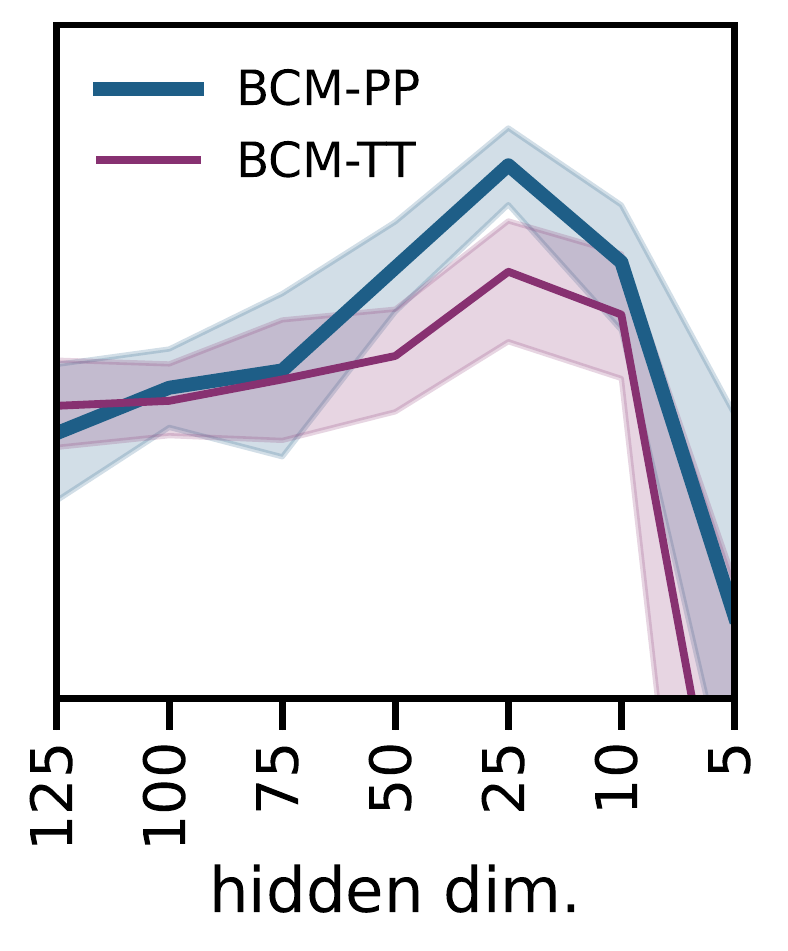}\vspace{3.55mm}
    \caption{Hidden dim.}
\end{subfigure}
\begin{subfigure}[b]{0.3733\columnwidth}
    \centering
    \includegraphics[width=\textwidth]{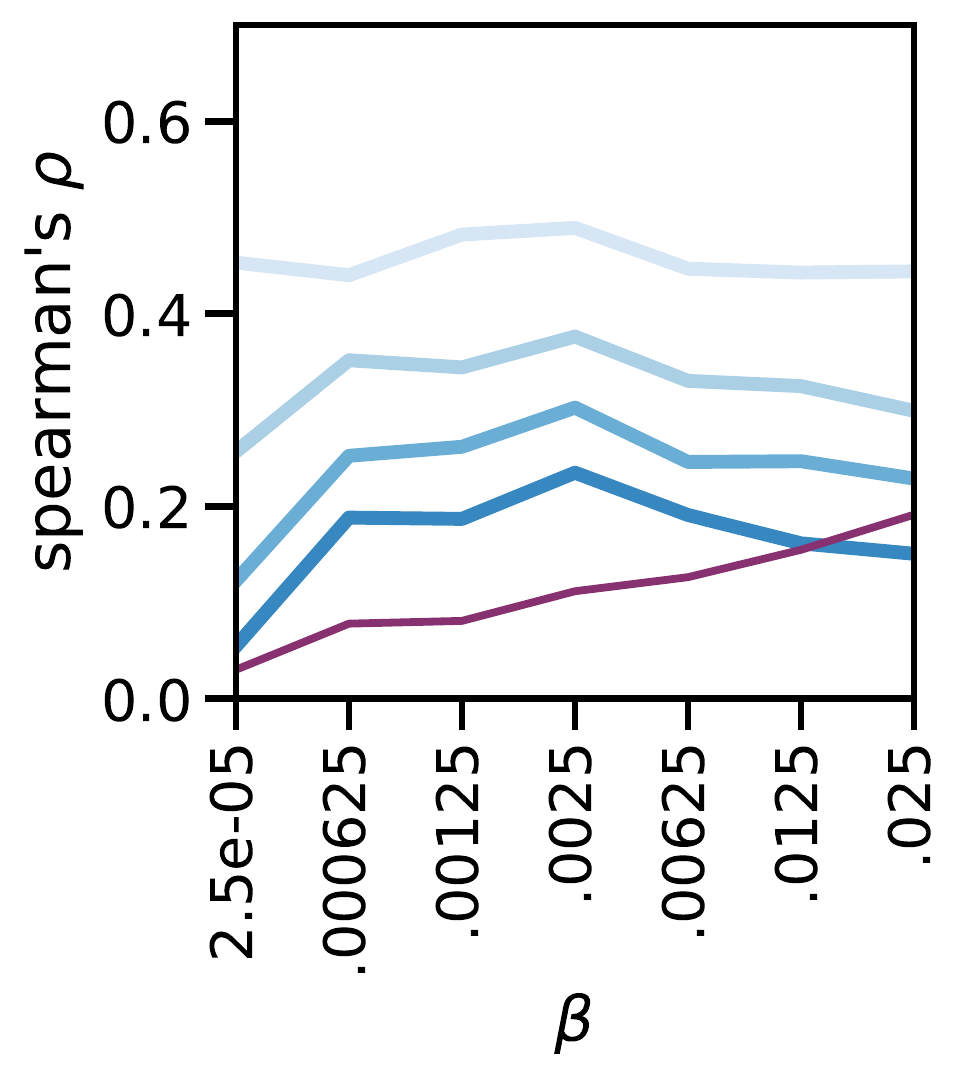}
    \caption{DVIB}
\end{subfigure}
\begin{subfigure}[b]{0.3033\columnwidth}
    \centering
    \includegraphics[width=\textwidth]{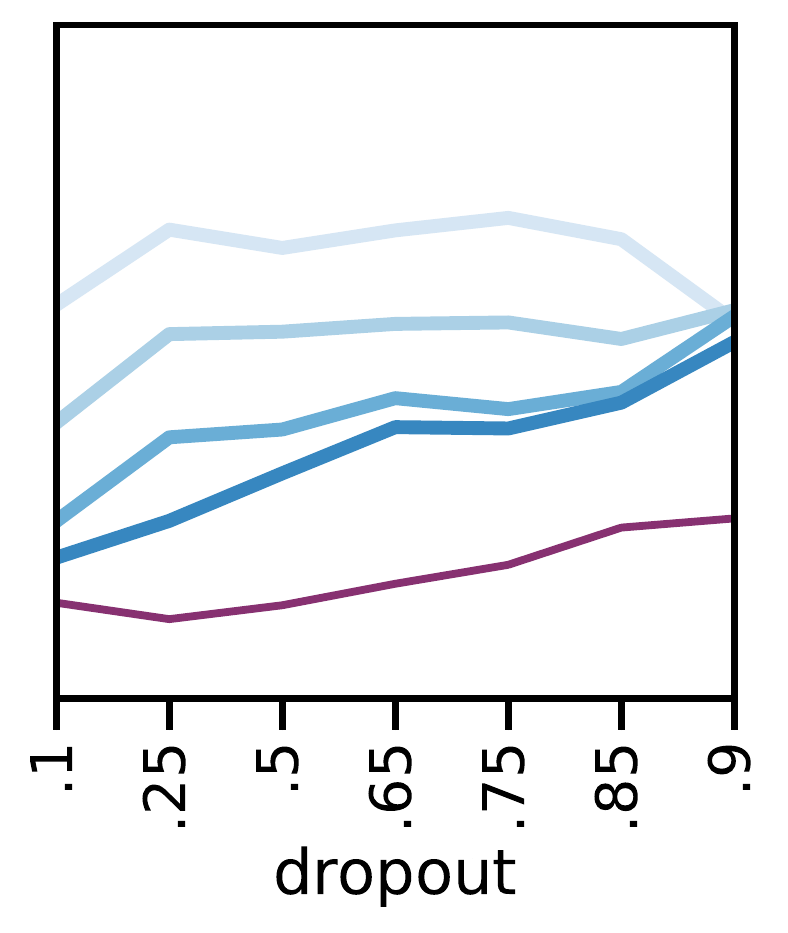}\vspace{4.5mm}
    \caption{Dropout}
\end{subfigure}
\begin{subfigure}[b]{0.3033\columnwidth}
    \centering
    \includegraphics[width=\textwidth]{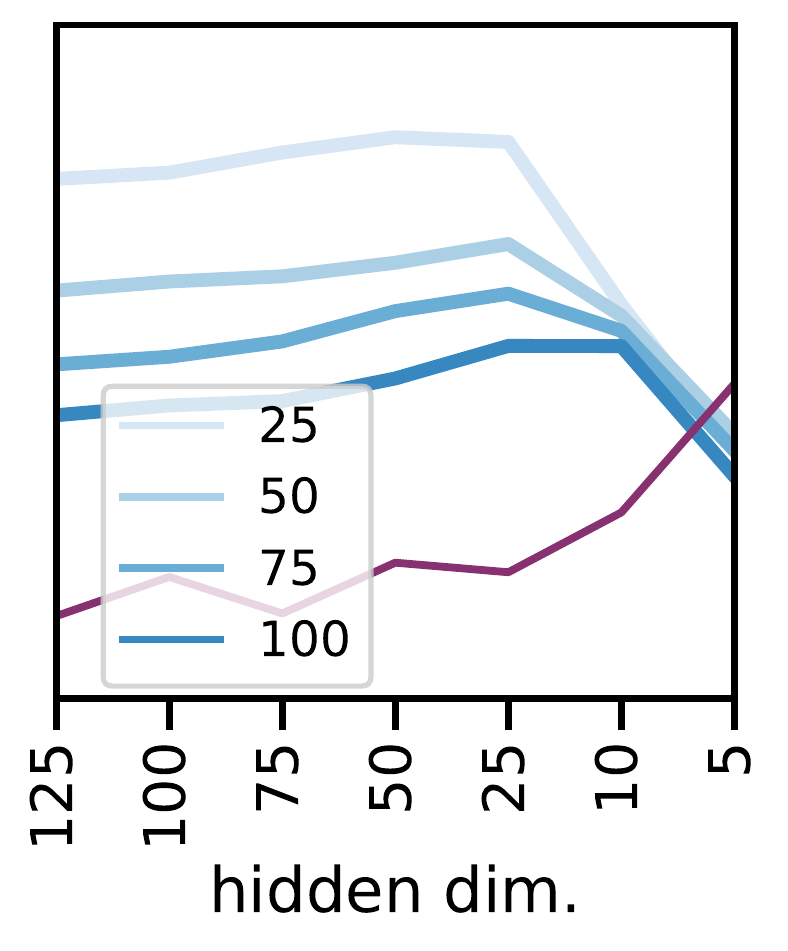}\vspace{3.8mm}
    \caption{Hidden dim.}
\end{subfigure}
\caption{Pearson's $r$ for the predictions of sentiment-only baselines and bottleneck models (a-c) and Spearman's $\rho$ for the SST validation set compositionality ranking of the baselines and bottleneck models (d-f), when varying the number of CCA dimensions used.} 
\label{fig:correlation_performance_baseline}
    \vspace{-0.2cm}
\end{figure}

\begin{figure*}[!h]
    \centering
    \includegraphics[width=\textwidth]{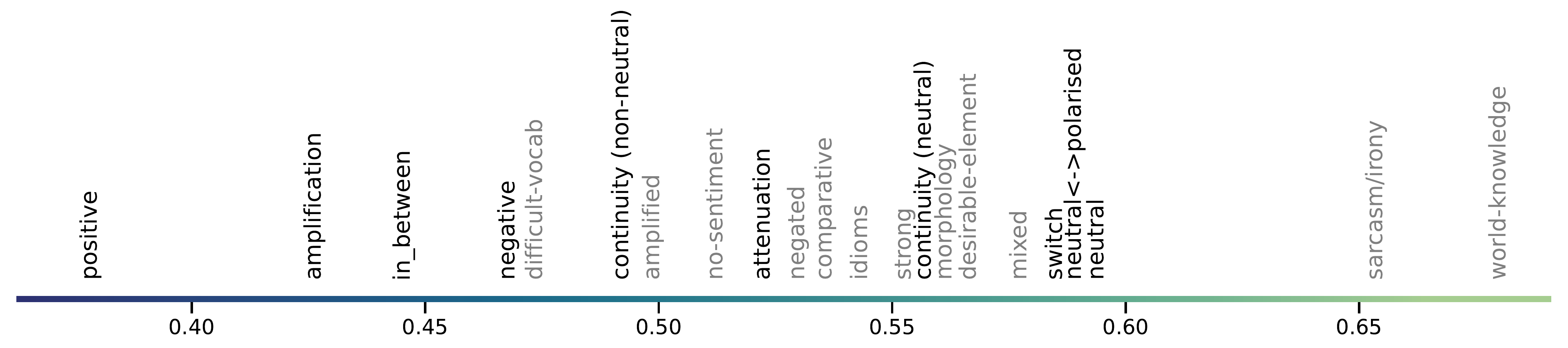}
    \caption{Categories of SST examples and their average position on the compositionality ranking visualised for the BCM-PP with the hidden dimensionality bottleneck and $d=25$. Categories in black are assigned by us; categories in gray are from \citet{barnes2019sentiment}. The categories are further explained in the main text and Appendix~\ref{ap:sentiment_categories}. Jittering was applied to better visualise overlapping categories.}
    \label{fig:sentiment_continuum}
    \vspace{-0.2cm}
\end{figure*}

\subsection{A sentiment-only baseline}
To assert that the bottlenecks make the models more compositional, we create a sentiment-only baseline that is given as input not words but their sentiment, and has a hidden dimensionality $d=25$.
Non-compositional patterns that arise from the composition of certain words rather than the sentiment of those words (e.g. ``drop dead gorgeous'') could hardly be captured by that model.
As such, the model exemplifies how sentiment can be computed more compositionally.
Figure~\ref{fig:sentiment_baseline} illustrates the default sentiment this model predicts for various input combinations.
Its predictions can be characterised by i) predicting \textbf{positive} for positive inputs, ii) predicting \textbf{negative} for negative inputs, iii) predicting \textbf{neutral} if one input is neutral, iv) predicting a class \textbf{in between} the input classes or as v) predicting the same class as its inputs (\textbf{continuity}).

The performance of the model is included in Figure~\ref{fig:sentiment_performance}, and Figure~\ref{fig:correlation_performance_baseline} (a-c) indicates the Pearson's $r$ between the sentiment predictions of bottleneck models and baseline models.
Generally, a higher $\beta$ or dropout probability, or a lower hidden dimensionality, leads to predictions that are more similar to this sentiment-only model,
unless the amount of regularisation is too extreme, debilitating the model (e.g. for dropout with probability 0.9).

\subsection{The Bottleneck Compositionality Metric}
Now we use BCM to obtain a ranking over the SST dataset. 
We separate the dataset into four folds, train on those folds for the base and bottleneck model, and compute the cosine distances for the hidden representations of examples in the test sets (using BCM-PP or BCM-TT).
We merge the cosine distances for the different folds, averaged over models trained with 10 random seeds, and order examples based on the resulting distance.
We select the values for $\beta$, dropout and the hidden dimensionality, as well as the number of CCA directions to use, based on rankings computed over the SST validation data.
Figure~\ref{fig:correlation_performance_baseline} (d-f) illustrates how the rankings of bottleneck models correlate with rankings constructed using the sentiment-only baseline.
We select 25 directions, $\beta=0.0025$, dropout $p=0.65$ and a hidden dimensionality of 25 to compute the full ranking. Contrary to the arithmetic task, BCM-TT underperforms compared to BCM-PP.

Different from the arithmetic task, it is unclear which examples \textit{should} be at the start or end of the ranking. 
Therefore, we examine the relative position of categories of examples in Figure~\ref{fig:sentiment_continuum} for the BCM-PP with the hidden dimensionality bottleneck, and in Appendix~\ref{ap:sentiment_rankings} for the remaining rankings.
The categories include the previously introduced ones, augmented with the following four:
\begin{itemize}[topsep=0pt,itemsep=0pt,parsep=0pt,partopsep=0pt]
    \item \textbf{amplification}: the root is even more positive/negative than its top two children;
    \item \textbf{attenuation}: the root is less positive/negative than its top two children;
    \item \textbf{switch}: the children are positive/negative, but the root node flips that sentiment;
    \item \textbf{neutral$\leftrightarrow$polarised}: the inputs are sentiment-laden, but the root is neutral, or vice versa.
\end{itemize}
We also include characterisations from \citet{barnes2019sentiment}, who label examples from the SST test set, for which state-of-the-art sentiment models cannot seem to predict the correct label, including, for example, cases where a sentence contains mixed sentiment, or sentences with idioms, irony or sarcasm.
Appendix~\ref{ap:sentiment_categories} elaborates on the meaning of the categories.
Figure~\ref{fig:sentiment_continuum} illustrates the relative positions of our sentiment characterisations and those of \citeauthor{barnes2019sentiment} on that ranking.
Patterns associated with more compositional sentiment processing, such as `positive', `negative' and `in between' lead to hidden representations that are more similar between the base model and bottleneck models than the dataset average (the mid point, 0.5). Atypical patterns like `switch' and `neutral$\leftrightarrow$polarised', on the other hand, along with the characterisations by \citeauthor{barnes2019sentiment} lead to less similar hidden representations.
Appendix~\ref{ap:sentiment_rankings} presents the same results for all six rankings considered, along with example sentences from across the ranking, to illustrate the types of sentences encountered among the most compositional and the least compositional ones.

\subsection{Example use cases}
\label{sec:sentiment_applications}
Compositionality rankings can be used in multiple manners, of which we illustrate two below.

\begin{figure}\centering
\begin{subfigure}[b]{0.49\columnwidth}
    \centering
    \includegraphics[width=\textwidth]{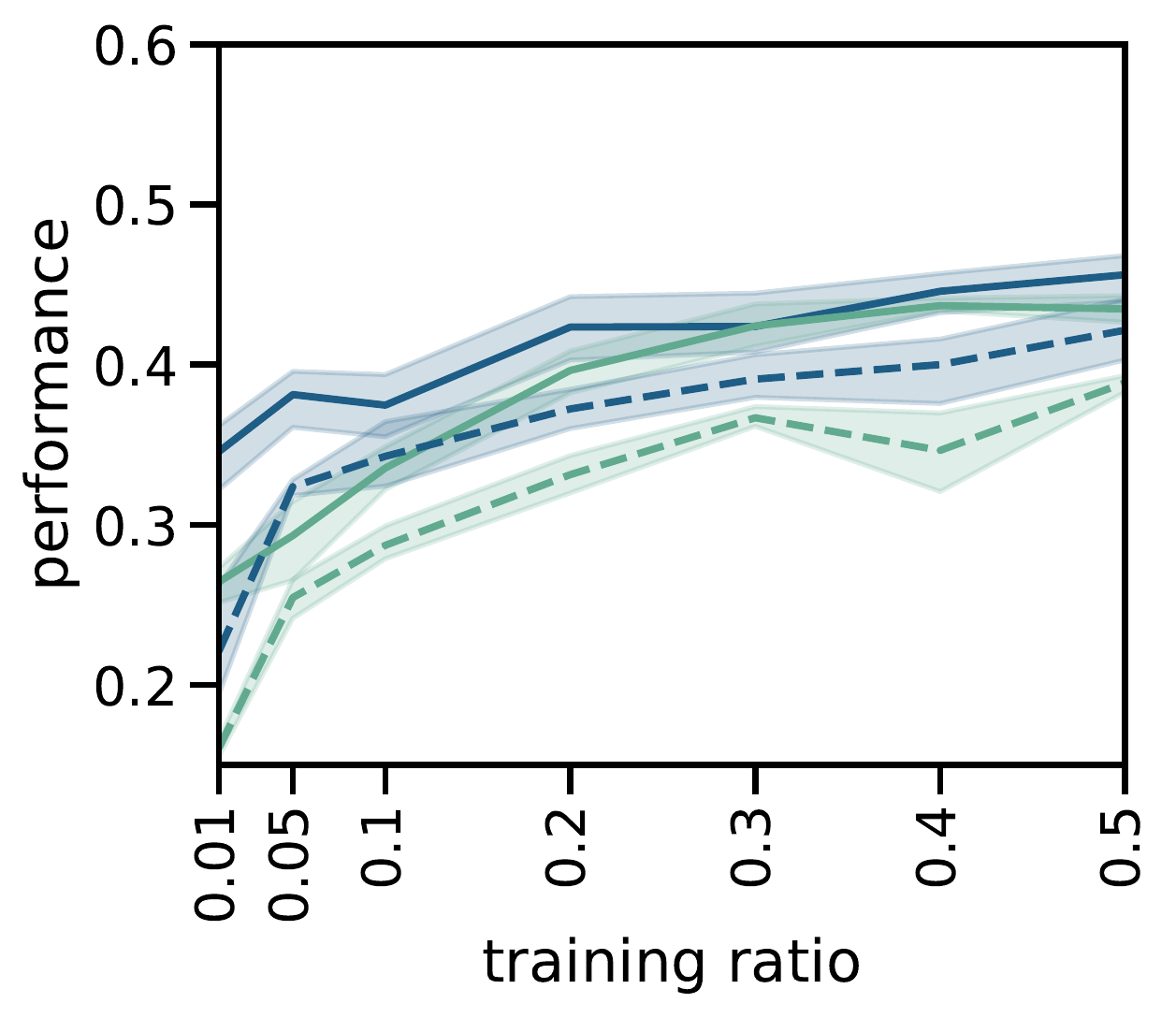}
    \caption{LSTM}
    \label{fig:lstm_increasing}
\end{subfigure}
\begin{subfigure}[b]{0.49\columnwidth}
    \centering
    \includegraphics[width=\textwidth]{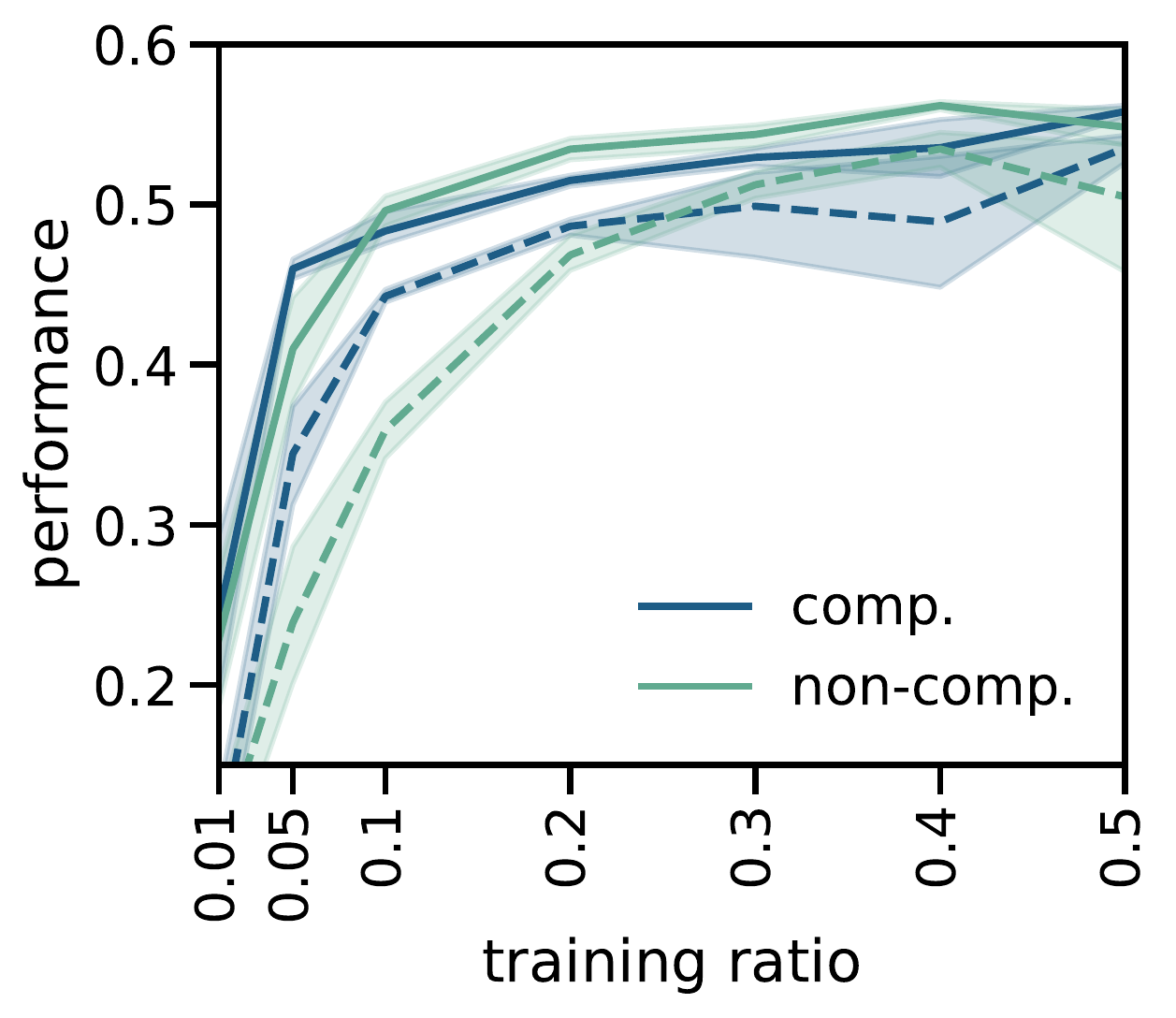}
    \caption{Roberta}
    \label{fig:roberta_increasing}
\end{subfigure}
\caption{Change in SST test set accuracy (solid) and macro-averaged $F_1$-score (dashed) as the training set size increases, for LSTM and Roberta models. The examples are from the most (in blue) or the least compositional (in green) portion of the ranking from the BCM-PP metric with the hidden dimensionality bottleneck.}
\label{fig:increasing}
\end{figure}

\paragraph{When data is scarce: use compositional examples}
Assuming that most of natural language \textit{is} compositional, one would expect that when limiting the training data, selecting compositional examples yields the best test performance.
To investigate this, we train models on various training dataset sizes, and evaluate with the regular test set.
The training data is taken from the start of the ranking for the `compositional' setup, and from the end of the ranking for the `non-compositional' setup (excluding test data), while ensuring equal distributions over input lengths and output classes.
We train a two-layer bidirectional LSTM with 300 hidden units, and Roberta-base \citep{liu2019roberta}, using batch size 4.
The models are trained for 10 and 5 epochs, respectively, with learning rates $\mathrm{2}{e-4}$ and $\mathrm{5}{e-6}$.
Because the ranking is computed over full sentences, and not subexpressions, we train the models on the sentiment labels for the root nodes.
Figure~\ref{fig:increasing} presents the results, consolidating that when data is scarce, using compositional examples is beneficial.

\begin{table}[]
    \centering
    \resizebox{\columnwidth}{!}{
        \begin{tabular}{lcccccc}
        \toprule
        \textbf{Model} & \multicolumn{2}{c}{\textbf{Comp.}} & \multicolumn{2}{c}{\textbf{Non-comp.}} & \multicolumn{2}{c}{\textbf{Random}} \\
                       & Acc. & $F_1$ & Acc. & $F_1$ & Acc. & $F_1$ \\ \midrule
        Roberta& .546 & .535& .516 & .487& .565 & .549\\
        LSTM& .505 & .485& .394 & .310& .478 & .447\\
        \bottomrule
        \end{tabular}
    }
    \caption{Performance on the new SST compositionality splits, generated using the ranking from the BCM-PP metric with the hidden dimensionality bottleneck.}
    \label{tab:data_splits}
    \vspace{-0.3cm}
\end{table}

\paragraph{Non-compositional examples are challenging}
For the same models, Table~\ref{tab:data_splits} indicates how performance changes if we redistribute train and test data such that the test set contains the most compositional examples, or the least compositional ones (keeping length and class distributions similar).
The non-compositional test setup is more challenging, with an 11 percentage point drop in accuracy for the LSTM, and a 3 point decrease for Roberta.

\vspace{2mm}

\noindent In conclusion, applying the BCM to the sentiment data has confirmed findings previously observed for the arithmetic toy task. While it is harder to understand whether the method actually filters out non-compositional examples, both comparisons to a sentiment-only baseline, and the average position of cases for which the composition of sentiment is known to be challenging (e.g. for `mixed' sentiment, `comparative' sentiment or `sarcasm'), suggest that compression acts as a compositionality metric. We also illustrated two ways in which the resulting ranking can be used.

\section{Conclusion}
This work presents the Bottleneck Compositionality Metric, a TRE-based metric \citep{andreas2018measuring} that is task-independent and can be applied to inputs of varying lengths: we pair up Tree-LSTMs where one of them has more compressed representations due to a bottleneck (the DVIB, hidden dimensionality bottleneck or dropout bottleneck), and use the distance between their hidden representations as a per-datum metric.
The method was applied to rank examples in datasets from most compositional to least compositional, which is of interest due to the growing relevance of compositional generalisation research in NLP, which assumes the compositionality of natural language, and encourages models to compose meanings of expressions rather than to memorise phrases as chunks.
We provided a proof-of-concept using an arithmetic task but also applied the metric to the much more noisy domain of sentiment analysis.

The different bottlenecks lead to qualitatively similar results. This suggests that, while DVIB might be better motivated (it directly optimises an estimate of the Shannon information passed across the network), its alternatives may be preferable in practice due to their simplicity.

Because natural language itself is not fully compositional,
graded metrics like the ones we presented can support future research, such as i) learning from data according to a compositionality-based curriculum to improve task performance, ii) filtering datasets to improve compositional generalisation, or iii) developing more and less compositional models depending on the desiderata for a task -- e.g. to perform well on sentences with idioms, one may desire a more non-compositional model.
In addition, the formulation of the metric was general enough to be expanded upon in the future: one could pair up other models, such as an LSTM and a Tree-LSTM, or a Transformer and its recursive variant,
as long as one keeps in mind that the compositional reconstruction itself should not be too powerful.
After all, even Tree-LSTMs could capture the exceptions in the arithmetic dataset despite their hierarchical inductive bias.

\section*{Limitations}
We identify three types of limitations for the work presented:
\begin{itemize}[topsep=0pt,itemsep=0pt,parsep=0pt,partopsep=0pt]
\item A \textbf{conceptual limitation} is that we work from a very strict definition of compositionality (\textit{local} compositionality), which essentially equates language with arithmetic. While overly restrictive, current datasets testing compositional generalisation follow this notion. The framework might be extensible to more relaxed notions by allowing for token disambiguation by using contextualised token embeddings and only enforcing a bottleneck on the amount of further contextual integration within the model added on top of the token embeddings.
\item In terms of \textbf{methodological limitations}, the use of only Tree-LSTMs -- although well-motivated from the perspective of compositional processing -- is a major limitation. Tree-LSTMs are most suited for sentence classification tasks, limiting the approach's applicability to sequence-to-sequence tasks.
Nonetheless, the bottlenecks can be integrated in other types of architectures that process inputs in a hierarchical manner, such as sequence-to-sequence models inducing latent source and target trees \citep{kim2021sequence} to yield an alternative implementation of the BCM.

Our work also assumes that an input's tree structure is known, which might not always be the case. Therefore, the compositionality ranking obtained using BCM always depends on the trees used: what is non-compositional given one (potentially inadequate) structure might be more compositional given another (improved) structure.
\item Lastly, the \textbf{evaluation} of our approach is limited in the natural domain through the absence of gold labels of the compositionality of examples in the sentiment analysis task, but for other tasks that could have been considered, the same limitation would have applied.
\end{itemize}
\section*{Acknowledgements}

We thank Chris Lucas for his contributions to this project when it was still in an early stage, Kenny Smith for his comments on the first draft of this paper, and Matthias Lindemann for excellent suggestions for the camera-ready version.
VD is supported by the UKRI Centre for Doctoral Training in Natural Language Processing, funded by the UKRI (grant EP/S022481/1) and the University of Edinburgh, School of Informatics and School of Philosophy, Psychology \& Language Sciences.
IT acknowledges the support of the European Research Council (ERC StG BroadSem 678254) and the Dutch National Science Foundation (NWO Vidi 639.022.518).

%\clearpage
\bibliographystyle{acl_natbib}
\bibliography{references}
\clearpage

\appendix

\section{Bottlenecks for arithmetic}
\label{ap:arithmetic}
\subsection{Performance for other bottlenecks}
\label{ap:arithmetic_mse}
\begin{figure}[!h]\small\centering
\begin{subfigure}[b]{0.47\columnwidth}
    \centering
    \includegraphics[width=\textwidth]{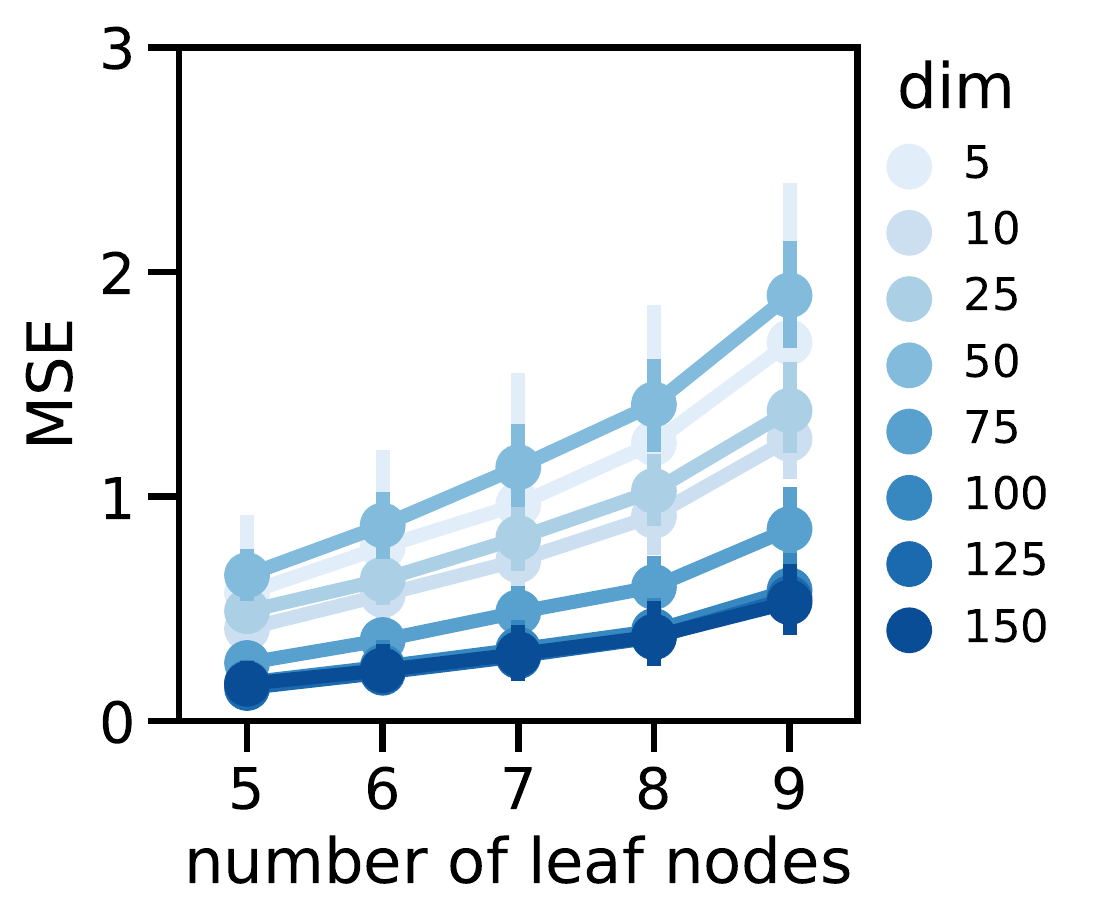}
    \caption{Regular examples}
\end{subfigure}
\begin{subfigure}[b]{0.47\columnwidth}
    \centering
    \includegraphics[width=\textwidth]{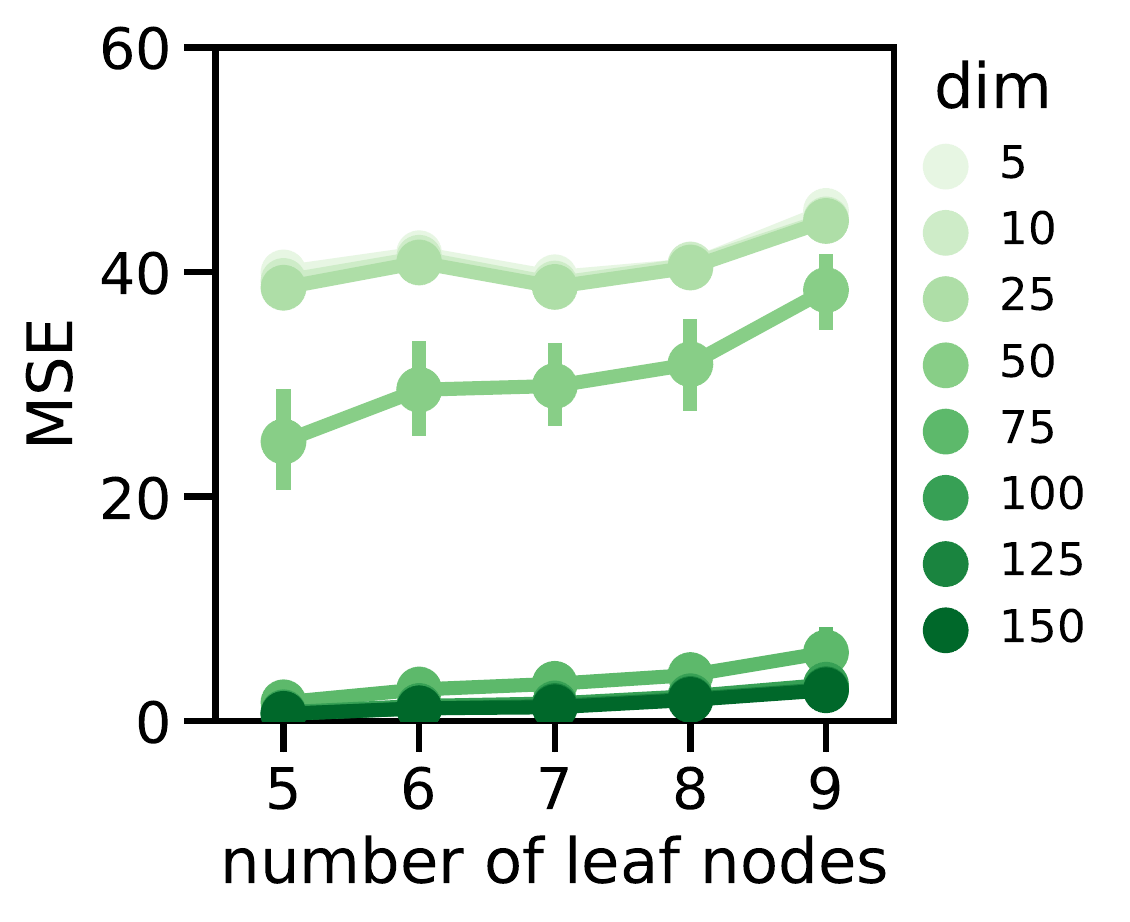}
    \caption{Exceptions}
\end{subfigure}
\caption{Performance on the arithmetic task for the Tree-LSTM with varying hidden dimensionalities.}
\label{fig:ap_performance_size}
\vspace{-0.5cm}
\end{figure}

\begin{figure}[!h]\small\centering
\begin{subfigure}[b]{0.47\columnwidth}
    \centering
    \includegraphics[width=\textwidth]{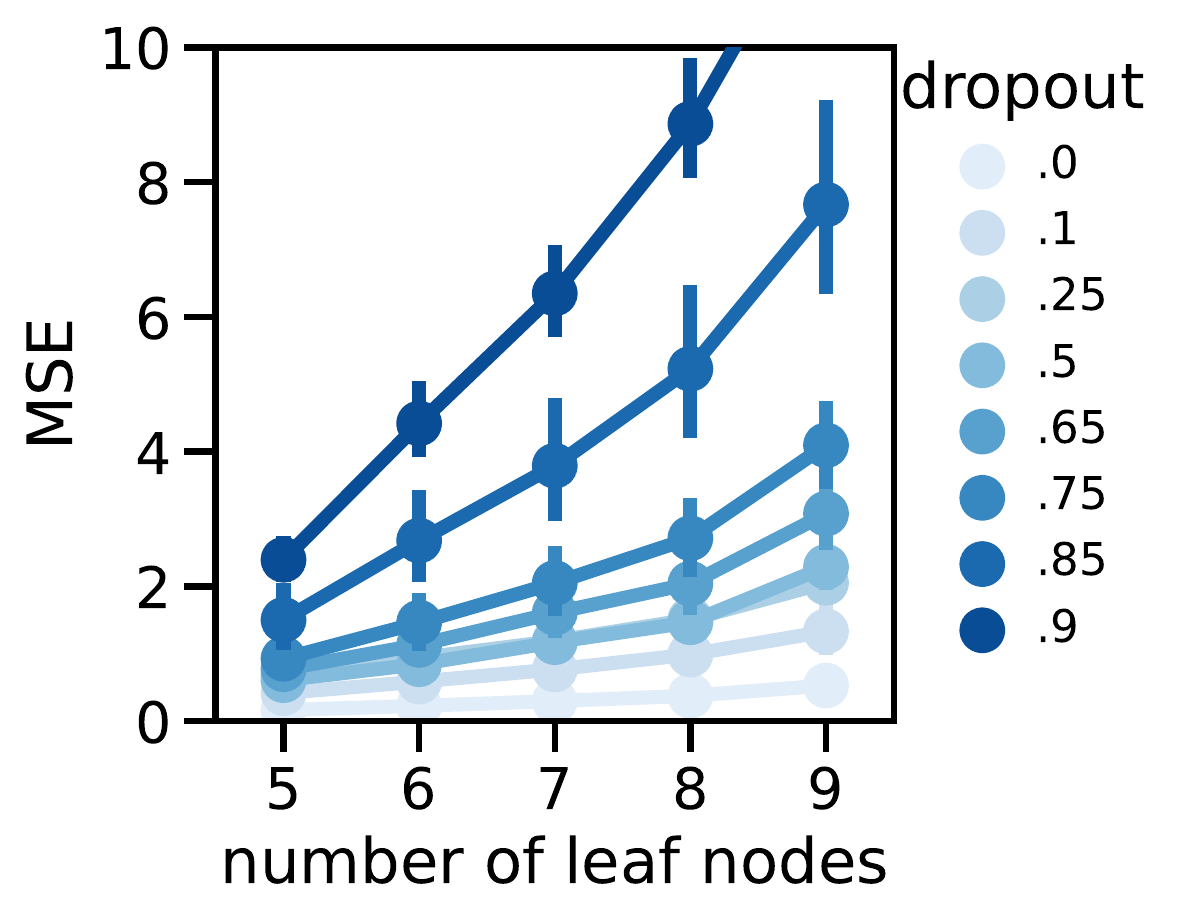}
    \caption{Regular examples}
\end{subfigure}
\begin{subfigure}[b]{0.47\columnwidth}
    \centering
    \includegraphics[width=\textwidth]{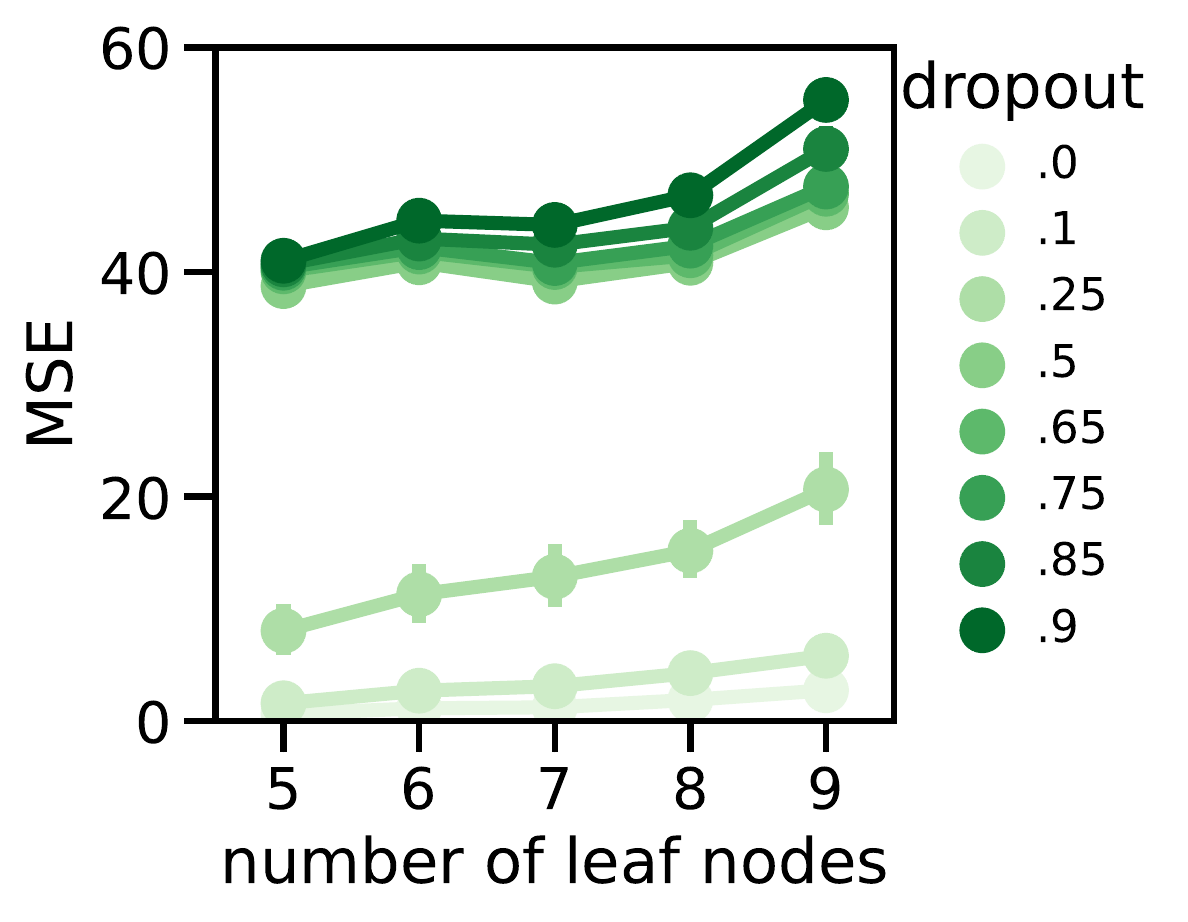}
    \caption{Exceptions}
\end{subfigure}
\caption{Performance on the arithmetic task for the Tree-LSTM with varying dropout probabilities.}
\label{fig:ap_performance_dropout}
\end{figure}

Figures~\ref{fig:ap_performance_size} and~\ref{fig:ap_performance_dropout} display the MSE for models with the hidden dimensionality bottleneck and dropout bottleneck.

\subsection{Bottleneck training dynamics}
\label{ap:arithmetic_training}

\begin{figure}[!h]\small\centering
\begin{subfigure}[b]{0.49\columnwidth}
    \centering
    \includegraphics[width=\textwidth]{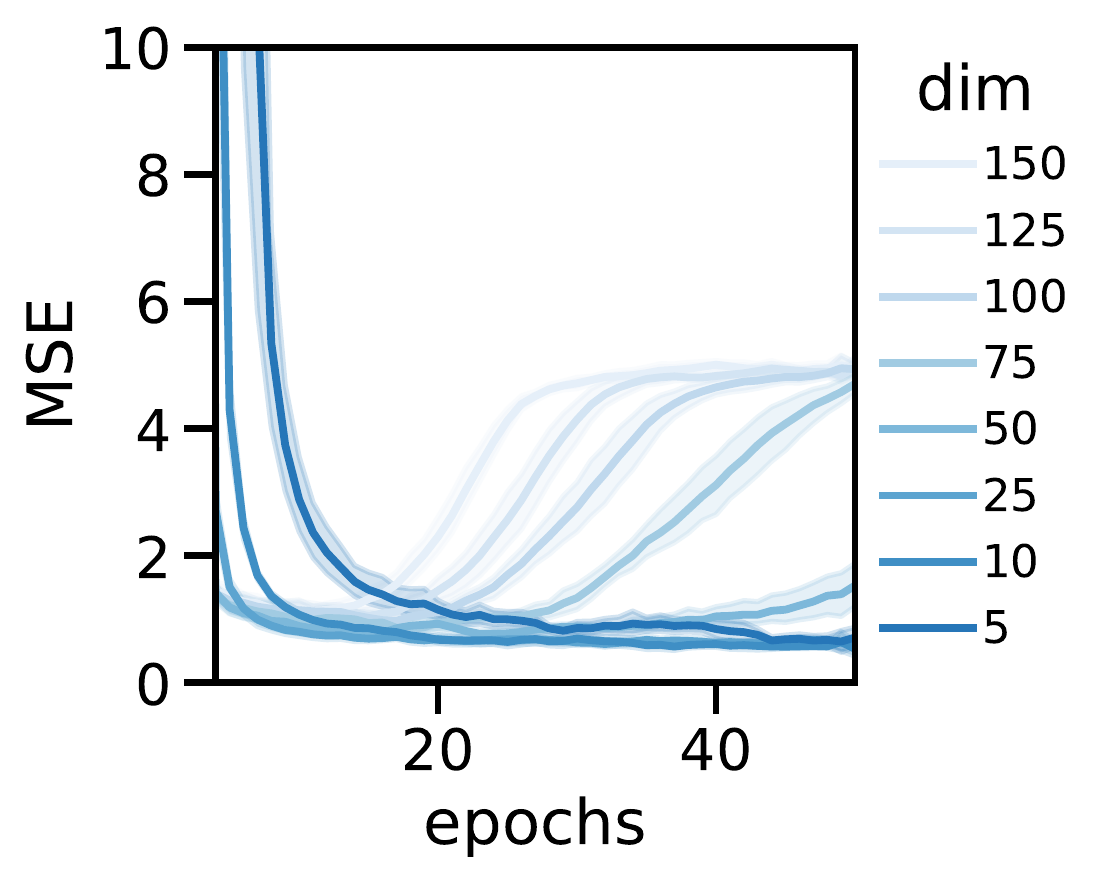}
    \caption{Compositional targets}
\end{subfigure}
\begin{subfigure}[b]{0.49\columnwidth}
    \centering
    \includegraphics[width=\textwidth]{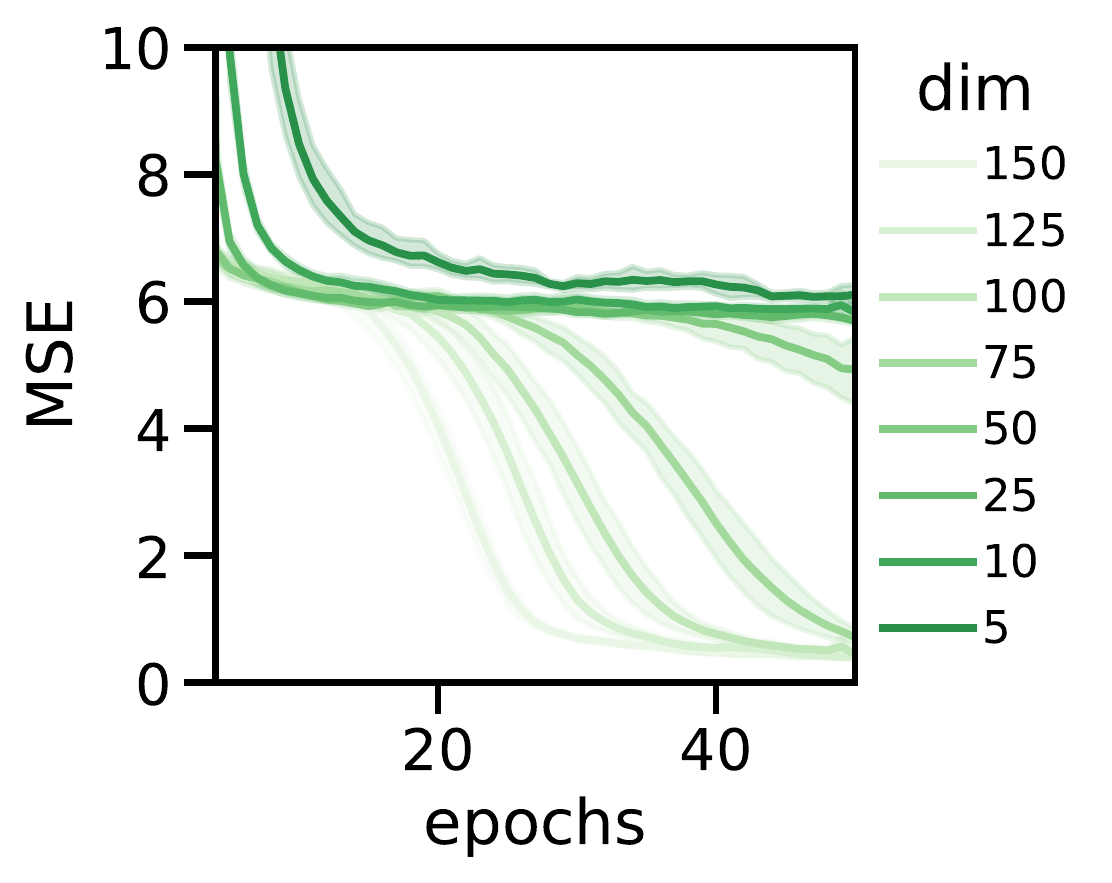}
    \caption{Non-comp. targets}
\end{subfigure}
\caption{Training dynamics for the hidden dimensionality bottleneck.}
\vspace{-0.5cm}
\label{fig:ap_training_dynamics_size}
\end{figure}

\begin{figure}[!h]\small\centering
\begin{subfigure}[b]{0.49\columnwidth}
    \centering
    \includegraphics[width=\textwidth]{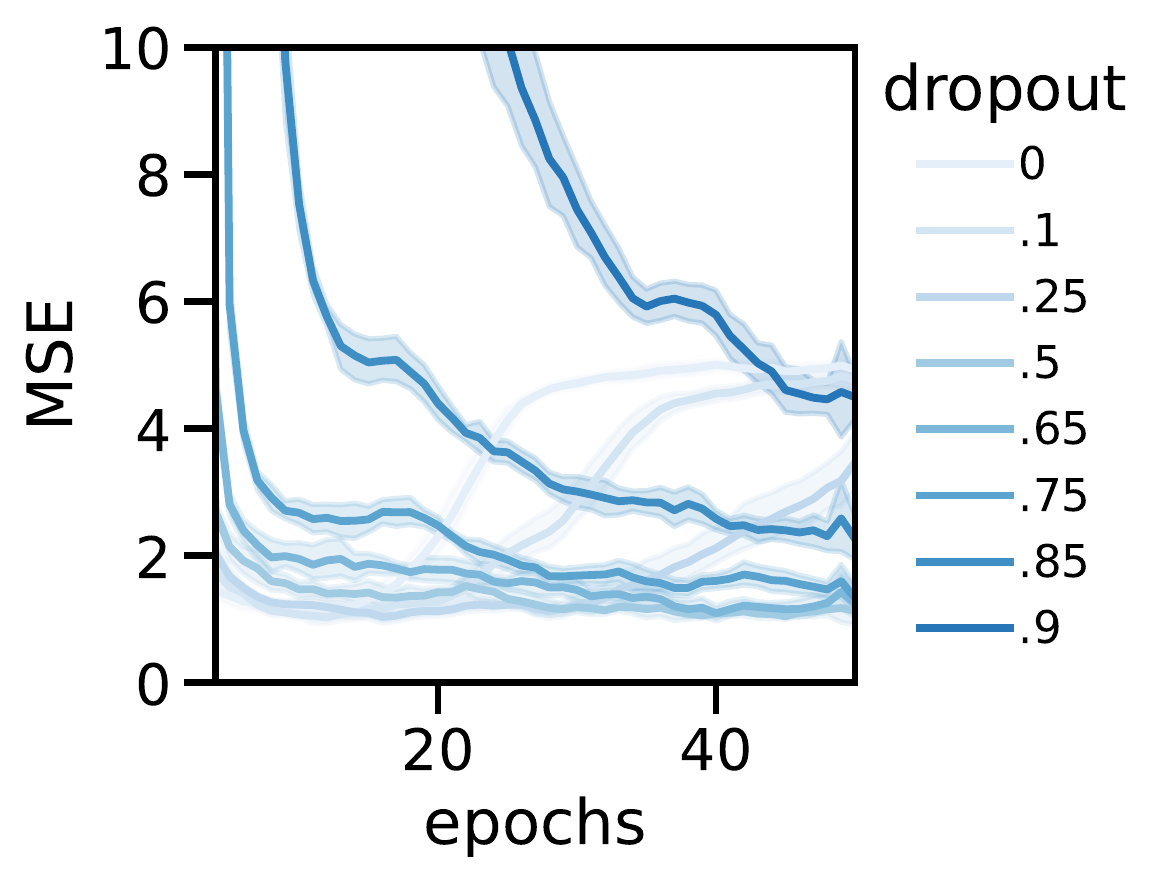}
    \caption{Compositional targets}
\end{subfigure}
\begin{subfigure}[b]{0.49\columnwidth}
    \centering
    \includegraphics[width=\textwidth]{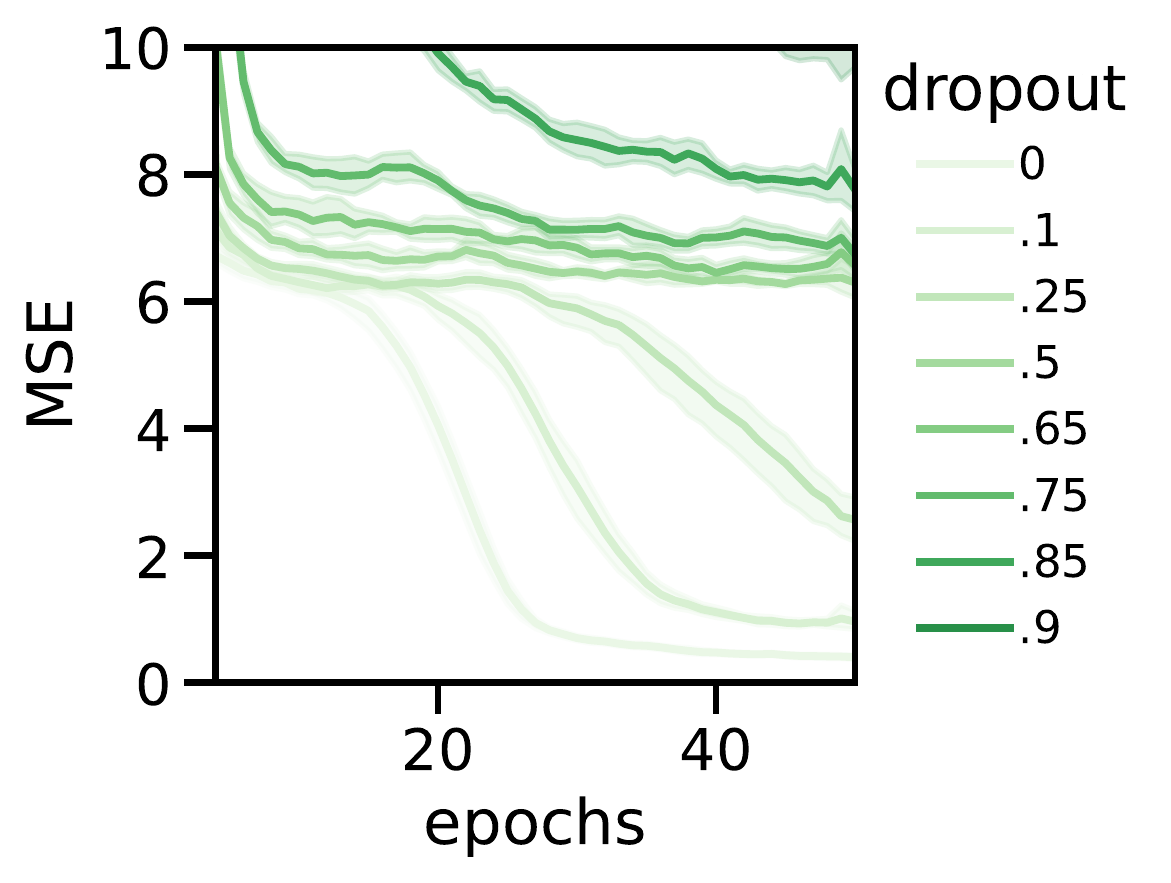}
    \caption{Non-comp. targets}
\end{subfigure}
\caption{Training dynamics for the dropout bottleneck.}
\label{fig:ap_training_dynamics_dropout}

\end{figure}

The exceptions in the arithmetic task had both a compositional and non-compositional interpretation, essentially giving us two sets of targets for measuring the models' performance during training using the validation data.
Figures~\ref{fig:training_dynamics_dvib} (in the main paper),~\ref{fig:ap_training_dynamics_size}, and~\ref{fig:ap_training_dynamics_dropout} illustrate how early on during training, the MSE is lowest for the compositional targets for all hyperparameters used for the bottlenecks: the models \textit{overgeneralise} the regular interpretation of ``\texttt{0}'', applying it to all inputs. Later on, the models that have a high $\beta$, a high dropout probability or a small dimension stay in that `compositional state', whereas the remaining models learn to capture the ambiguity.

\section{Sentiment analysis}
\label{ap:sentiment}

\subsection{Categories of \citet{barnes2019sentiment}}
\label{ap:sentiment_categories}
\citeauthor{barnes2019sentiment} collect examples from sentiment analysis datasets that three state-of-the-art classifiers struggle with, and annotate them for the presence of 18 (para-)linguistic phenomena. We include the following categories from the SST test set in the visualisation of the SST ranking:
\begin{enumerate}[topsep=0pt,itemsep=0pt,parsep=0pt,partopsep=0pt]
    \item \textit{Negated}: phrases or sentences that are negated, where \citeauthor{barnes2019sentiment} identify a pattern of irrelevant negation throwing models off.
    \item \textit{Amplified}: cases where neutral modifiers act as strong contextual valence shifters.
    \item \textit{Strong}: \textit{very} positive or \textit{very} negative cases.
    \item \textit{Desirable element}: sentiment dominated by one `desirable element', such as ``pool''.
    \item \textit{Comparative}: cases that express sentiment by comparison (e.g. using ``better than'').
    \item \textit{Idioms}: cases containing idioms, for which the mapping from the words to the sentiment is often not straightforward.
    \item \textit{Mixed}: mixed positive \& negative sentiment.
    \item \textit{Difficult-vocab}: e.g. ``engrossing and psychologically resonant suspenser''.
    \item \textit{World-knowledge}: for instance comparisons between entities, where the entities imply a certain sentiment.
    \item \textit{Sarcasm/irony}: sarcasm is often present in negative examples, where the speaker is intending the opposite of what is said.
    \item \textit{No-sentiment}: neutral labelled examples.
    \item \textit{Morphology}: examples with morphological features that affect sentiment very positively or very negatively.
\end{enumerate}

\subsection{Alternative metrics}
\paragraph{Tree impurity score (TIS)} \citet{bhathena2020evaluating} propose basic compositionality metrics that rely on the difference between the label of the root node, and labels of subexpressions. We report their \textit{tree impurity score} (TIS): a simple metric that measures the absolute difference between the label of the root node and the average of all labels in a tree.

\paragraph{`Topographic' similarity}
\label{ap:sentiment_topographic}
A compositionality metric from language emergence literature is \textit{topographic similarity} \citep{brighton2006understanding}, that given a set of objects, their meanings and the associated signals computes the correlation of the distances between corresponding pairs of signals and meanings.
For instance, \citet{lazaridou2018emergence} compare symbolic signals in a referential game using the Levenshtein edit distance, and compare vector meaning representations through cosine distance.
Topographic similarity assumes that in a compositional language, similar signals should yield similar meaning representations.
However, the use of edit distance to directly compare sentences does not readily transfer to sentiment analysis, where one word changed in the input space can yield a large change in the predicted sentiment.

To approximate topographic similarity of meanings and signals, we instead manipulate the input in ways that should yield a similar prediction, and then rate examples based on the change in the predictions observed.
We replace nouns, verbs, adjectives or adverbs with a word that in SST has the same POS tag and sentiment label.
If a change is observed, the example is more likely to be a non-compositional example.

We apply this to the base models from the main paper, by randomly replacing one token that is a verb, noun, adjective or adverb with a different token of the same POS tag and sentiment label. We make 50 such modifications per sentence per model seed.
We record the average change in the predicted sentiment, and use this as a metric akin to topographic similarity.

\begin{figure}[t]
    \centering
    \includegraphics[width=0.95\columnwidth]{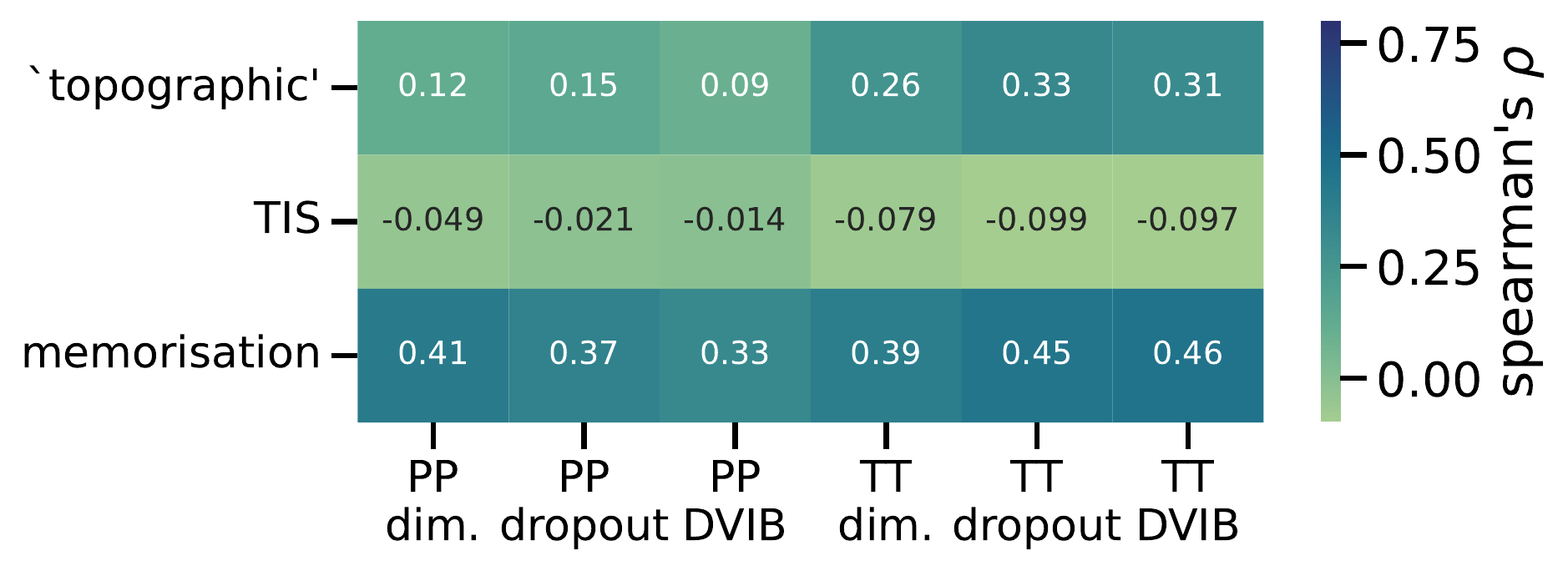}
    \caption{Illustration of how BCM rankings over the SST data correlate with baseline metrics that ought to be correlated: topographic similarity, tree impurity score and memorisation as per Spearman's $\rho$.}
    \label{fig:correlation_bl}
    \vspace{-0.1cm}
\end{figure}

\begin{figure}[t]
    \centering
    \includegraphics[width=0.85\columnwidth]{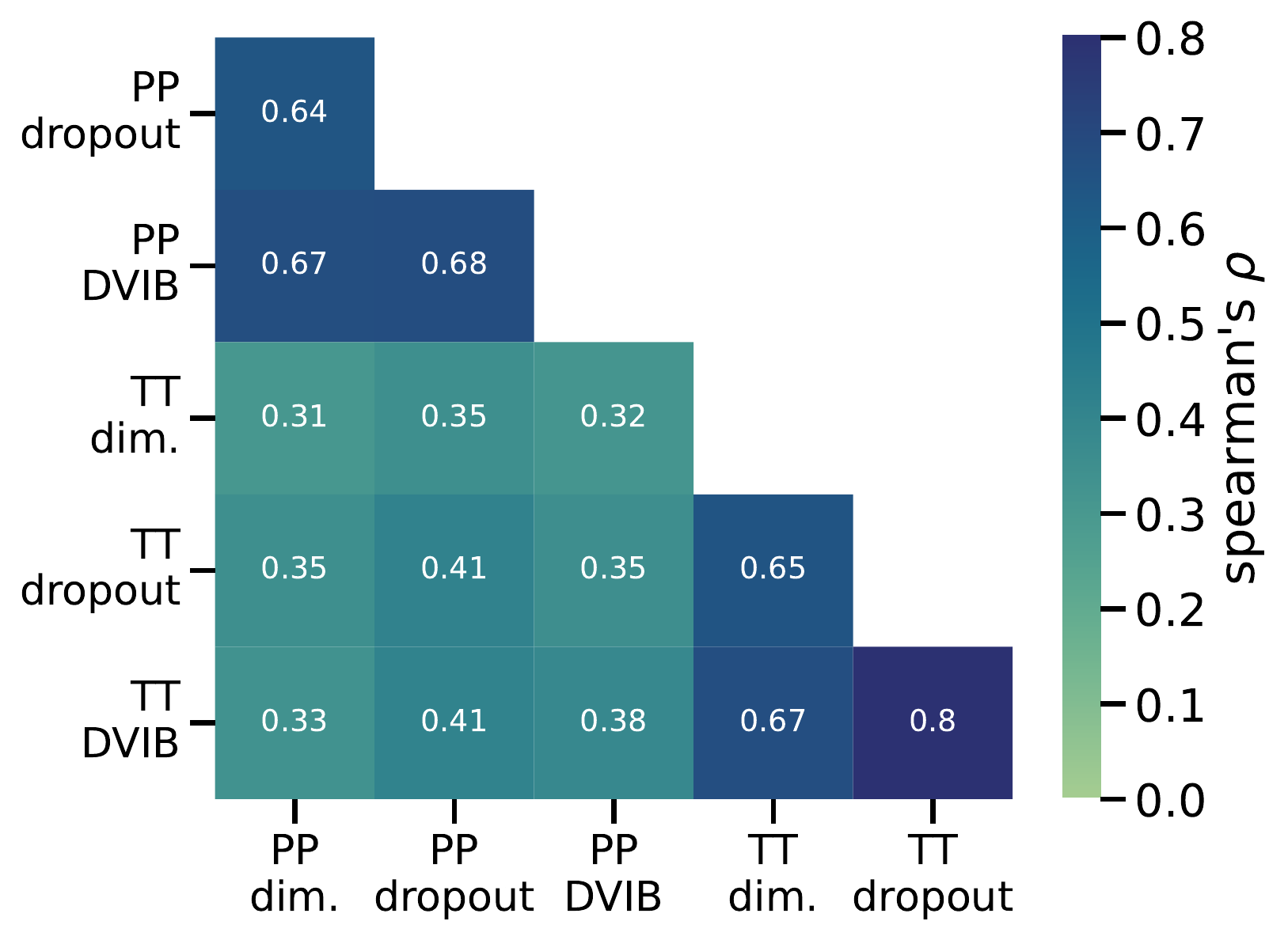}
    \caption{Illustration of how the BCM rankings over the SST dataset correlate as per Spearman's $\rho$.}
    \label{fig:correlation_rankings}
    \vspace{-0.1cm}
\end{figure}

\paragraph{Memorisation score}
\citet{zheng2022empirical} empirically evaluate the \textit{long tail theory} posed by \citet{feldman2020neural} (who validate their hypotheses using computer vision tasks), that states that for data distributions with a long tail, memorisation of training examples is required for near-optimal performance on the test data.
\citeauthor{zheng2022empirical} put this theory to the test for sentiment classification.
The metric of \citeauthor{zheng2022empirical} expresses how the likelihood of the target changes when an example is down-weighted during training. That change should be larger for memorised examples.
Intuitively, non-compositionality and memorisation are related: non-compositional patterns in data require memorising the atypical interpretation of words in specific contexts.
The metric is expected to be different from our metric -- ours is measured using \textit{test} data, while memorisation occurs during training -- but a positive correlation is expected between the two, nonetheless.
\citeauthor{zheng2022empirical} use the binary SST subtask and report their scores on preprocessed versions of SST sentences. We report the correlation for the examples for which we could find matching surface forms only.

\vspace{2mm}
\noindent Figure~\ref{fig:correlation_bl} indicates how the different BCM rankings correlate with these three alternative metrics.
Surprisingly, TIS negatively correlates with our rankings, but only weakly. This may be due to the simplicity of that metric, that ignores the tree structure and simply averages all sentiment of all nodes.
The devised `topographic' similarity positively correlates with our rankings, but only up to $\rho=0.33$.
The memorisation score has a moderate correlation with our rankings, of up to $\rho=0.46$ for the BCM-TT.
Figure~\ref{fig:correlation_rankings} illustrates how the rankings correlate with each other. The figure suggests that TRE training yields rankings that are still quite different from the post-processing ones.

\clearpage

\begin{figure*}[t]
\subsection{Additional results for SST rankings}
Here, we present further results for the BCM rankings over the SST datasets: firstly, Figure~\ref{fig:sentiment_continuum_all} provides the rankings for the six different BCM setups. They have commonalities, but also differences, e.g. in the BCM-TT setups the `neutral' sentiment is closer to the end of the ranking.

\label{ap:sentiment_rankings}
\centering
    \begin{subfigure}[b]{\textwidth}\centering
    \includegraphics[width=0.81\textwidth]{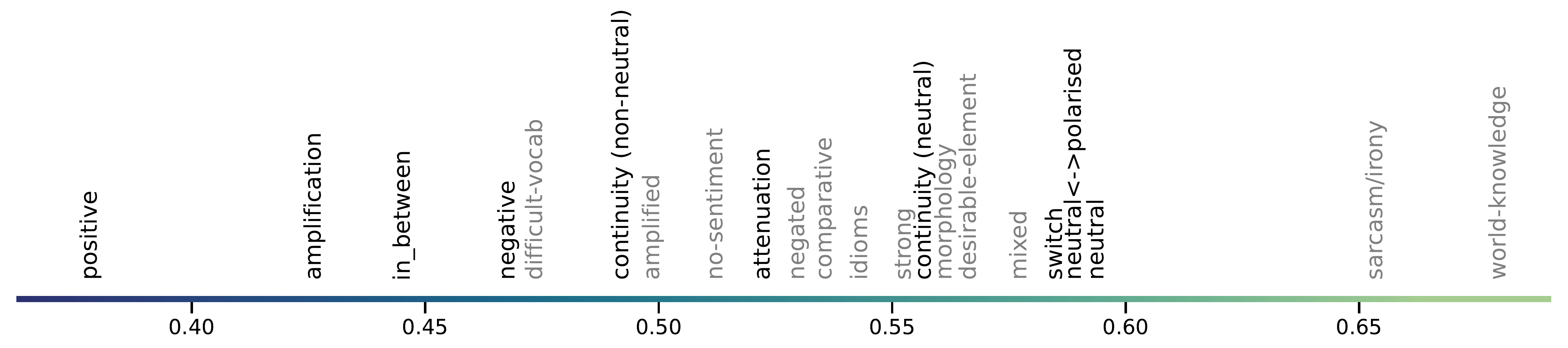}
    \caption{BCM-PP, Hidden dimensionality bottleneck}
    \end{subfigure}
    \begin{subfigure}[b]{\textwidth}\centering
    \includegraphics[width=0.81\textwidth]{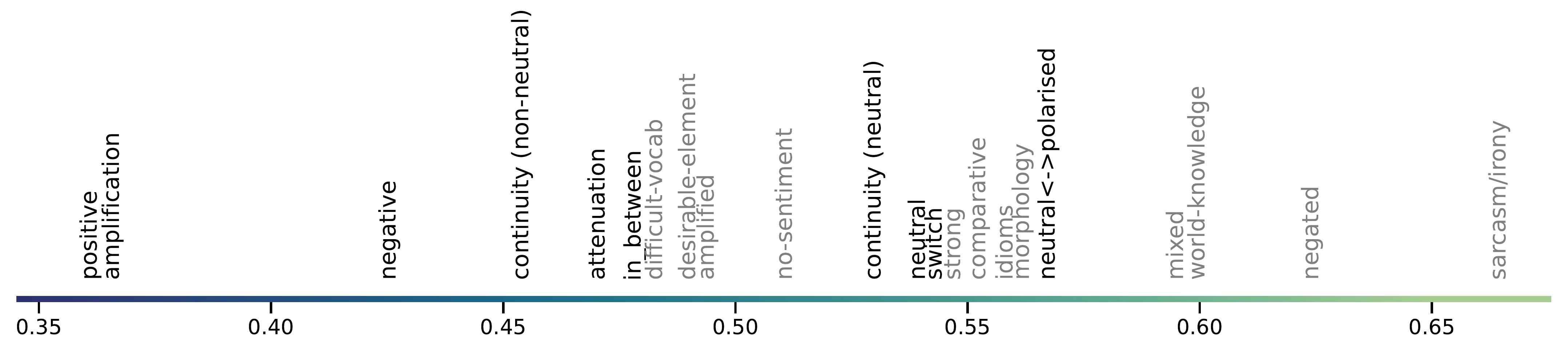}
    \caption{BCM-PP, DVIB bottleneck}
    \end{subfigure}
    \begin{subfigure}[b]{\textwidth}\centering
    \includegraphics[width=0.81\textwidth]{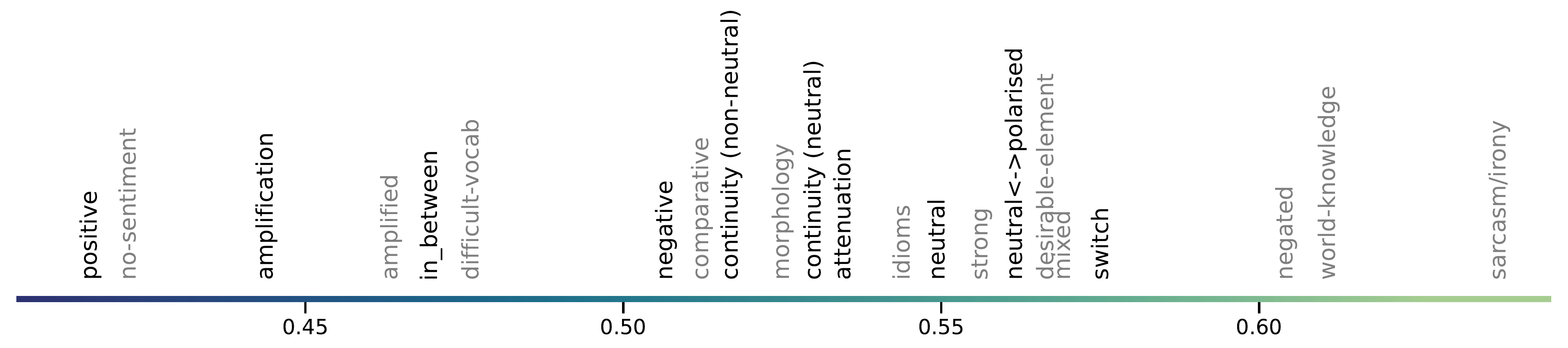}
    \caption{BCM-PP, Dropout bottleneck}
    \end{subfigure}
    \begin{subfigure}[b]{\textwidth}\centering
    \includegraphics[width=0.81\textwidth]{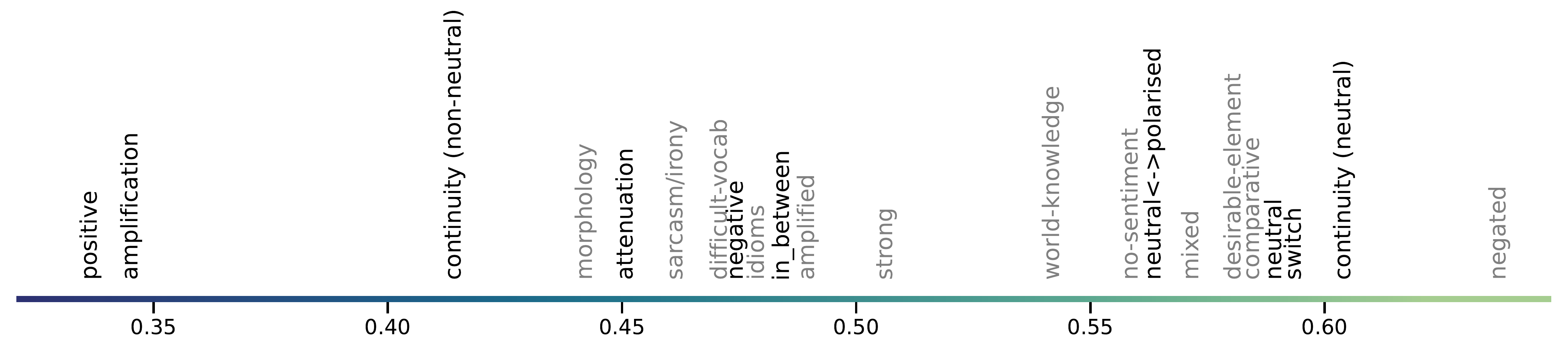}
    \caption{BCM-TT, Hidden dimensionality bottleneck}
    \end{subfigure}
    \begin{subfigure}[b]{\textwidth}\centering
    \includegraphics[width=0.81\textwidth]{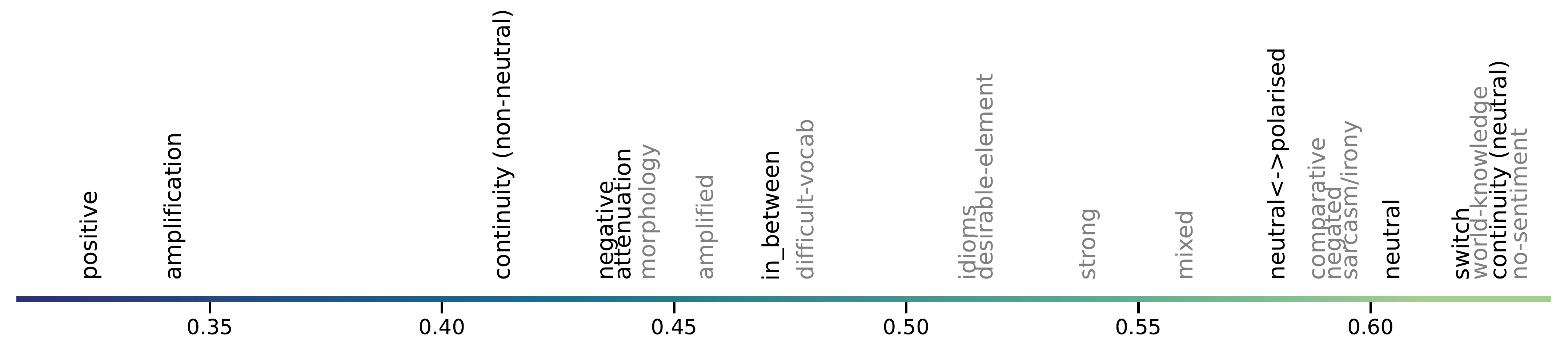}
    \caption{BCM-TT, DVIB bottleneck}
    \end{subfigure}
    \begin{subfigure}[b]{\textwidth}\centering
    \includegraphics[width=0.81\textwidth]{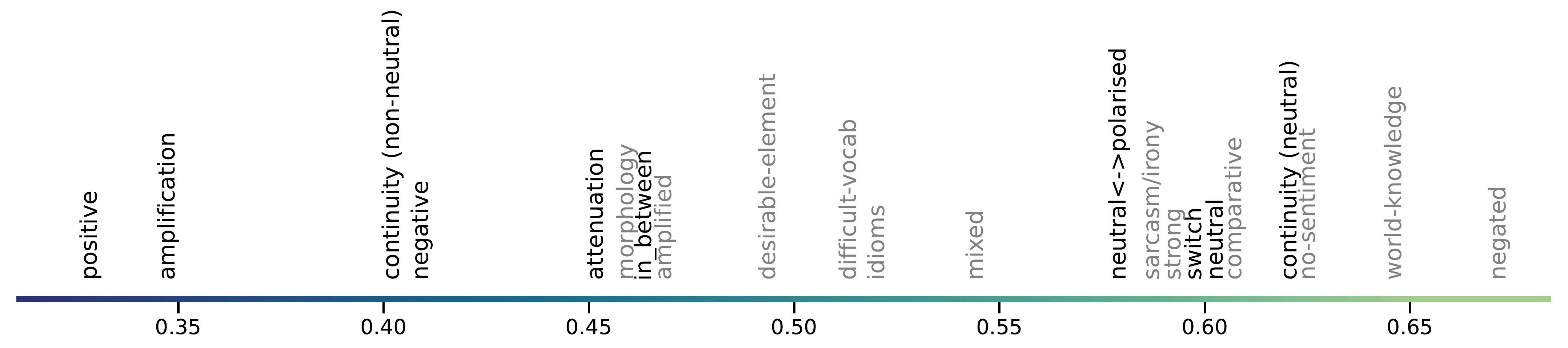}
    \caption{BCM-TT, Dropout bottleneck}
    \end{subfigure}
\caption{Example categories for SST test set examples, and their average position on the compositionality ranking for the hidden dimensionality bottleneck with $d=25$, the dropout bottleneck with $p=0.65$ or the DVIB with $\beta=0.0025$.
The categories in black are assigned by us, and the categories in gray are from \citet{barnes2019sentiment}. Jittering was applied to better visualise overlapping categories.}
\label{fig:sentiment_continuum_all}
\vspace{-0.5cm}
\end{figure*}

\begin{table*}\centering
Secondly, Table~\ref{fig:ranking_examples_all} provides example sentences from different parts of the rankings, randomly sampled. 0 indicates the most compositional examples, and 1 the least compositional ones.\\

    \resizebox{0.88\textwidth}{!}{\centering\small
    \begin{tabular}{ccl}
    \toprule
    \textbf{Relative position} & \textbf{Target} & \textbf{Sentence} \\\midrule\midrule
    \multicolumn{3}{l}{- \textit{BCM-PP, Hidden dim. bottleneck}} \\
    0.07 & 3 & tsai convincingly paints a specifically urban sense of disassociation here .\\
    0.17 & 0 & the film desperately sinks further and further into comedy futility .\\
    0.21 & 3 & a cop story that understands the medium amazingly well .\\
    0.40 & 3 & one scarcely needs the subtitles to enjoy this colorful action farce .\\
    0.42 & 1 & the entire movie is about a boring , sad man being boring and sad .\\
    0.53 & 2 & not everyone will play the dark , challenging tune taught by the piano teacher .\\
    0.66 & 3 & this is the stuff that disney movies are made of .\\
    0.75 & 3 & daughter from danang sticks with its subjects a little longer and tells a deeper story\\
    0.86 & 2 & not kids , who do n't need the lesson in repugnance .\\
    0.98 & 0 & lacks heart , depth and , most of all , purpose .\\\midrule
    \multicolumn{3}{l}{- \textit{BCM-PP, Dropout bottleneck}} \\
    0.09 & 4 & it 's a cool event for the whole family .\\
    0.14 & 0 & done in mostly by a weak script that ca n't support the epic treatment .\\
    0.20 & 3 & some movies are like a tasty hors-d'oeuvre ; this one is a feast .\\
    0.37 & 4 & my oh my , is this an invigorating , electric movie .\\
    0.41 & 3 & its director 's most substantial feature for some time .\\
    0.51 & 4 & the modern master of the chase sequence returns with a chase to end all chases\\
    0.62 & 3 & like its bizarre heroine , it irrigates our souls .\\
    0.78 & 1 & sadly , ` garth ' has n't progressed as nicely as ` wayne . '\\
    0.86 & 2 & you 're too conscious of the effort it takes to be this spontaneous .\\
    0.95 & 0 & a film of empty , fetishistic violence in which murder is casual and fun .\\\midrule
    \multicolumn{3}{l}{- \textit{BCM-PP, DVIB}} \\
    0.07 & 4 & highly recommended viewing for its courage , ideas , technical proficiency and great acting .\\
    0.18 & 3 & it manages to squeeze by on angelina jolie 's surprising flair for self-deprecating comedy .\\
    0.25 & 1 & goes on and on to the point of nausea .\\
    0.38 & 3 & the best part about `` gangs '' was daniel day-lewis .\\
    0.44 & 2 & a dopey movie clothed in excess layers of hipness .\\
    0.51 & 2 & a perplexing example of promise unfulfilled , despite many charming moments .\\
    0.64 & 0 & the entire film is one big excuse to play one lewd scene after another .\\
    0.79 & 1 & the problematic characters and overly convenient plot twists foul up shum 's good intentions .\\
    0.82 & 3 & the jabs it employs are short , carefully placed and dead-center .\\
    0.92 & 2 & then nadia 's birthday might not have been such a bad day after all .\\\midrule
    \multicolumn{3}{l}{- \textit{BCM-TT, Hidden dim. bottleneck}} \\
    0.00 & 3 & the story is smart and entirely charming in intent and execution .\\
    0.02 & 2 & for single digits kidlets stuart little 2 is still a no brainer .\\
    0.17 & 1 & outer-space buffs might love this film , but others will find its pleasures intermittent .\\
    0.29 & 3 & although shot with little style , skins is heartfelt and achingly real .\\
    0.37 & 3 & a knowing look at female friendship , spiked with raw urban humor .\\
    0.50 & 3 & there 's an energy to y tu mamá también .\\
    0.57 & 2 & a piquant meditation on the things that prevent people from reaching happiness .\\
    0.63 & 3 & too daft by half ... but supremely good natured .\\
    0.71 & 1 & the feature-length stretch ... strains the show 's concept .\\
    0.88 & 1 & a thriller without thrills and a mystery devoid of urgent questions .\\
    0.97 & 1 & the movie does n't generate a lot of energy .\\ \midrule
    \multicolumn{3}{l}{- \textit{BCM-TT, Dropout bottleneck}} \\
    0.03 & 3 & an extremely funny , ultimately heartbreaking look at life in contemporary china .\\
    0.14 & 1 & the pretensions -- and disposable story -- sink the movie .\\
    0.22 & 4 & that rara avis : the intelligent romantic comedy with actual ideas on its mind .\\
    0.39 & 2 & admirable , certainly , but not much fun to watch .\\
    0.50 & 1 & the end result is a film that 's neither .\\
    0.59 & 2 & throwing it all away for the fleeting joys of love 's brief moment .\\
    0.63 & 4 & kids should have a stirring time at this beautifully drawn movie .\\
    0.77 & 3 & the obnoxious title character provides the drama that gives added clout to this doc .\\
    0.87 & 0 & stitch is a bad mannered , ugly and destructive little **** .\\
    0.98 & 0 & there is no pleasure in watching a child suffer .\\\midrule
    \multicolumn{3}{l}{- \textit{BCM-TT, DVIB}} \\
    0.03 & 4 & quite simply , a joy to watch and -- especially -- to listen to .\\
    0.14 & 1 & a fake street drama that keeps telling you things instead of showing them .\\
    0.22 & 1 & it tries too hard , and overreaches the logic of its own world .\\
    0.34 & 1 & plays less like a coming-of-age romance than an infomercial .\\
    0.48 & 4 & it 's one of the most honest films ever made about hollywood .\\
    0.52 & 3 & a low-key labor of love that strikes a very resonant chord .\\
    0.69 & 3 & it 's never laugh-out-loud funny , but it is frequently amusing .\\
    0.71 & 3 & a meditation on faith and madness , frailty is blood-curdling stuff .\\
    0.81 & 1 & a beautifully shot but dull and ankle-deep ` epic . '\\
    0.91 & 2 & it uses the pain and violence of war as background material for color .\\
    \bottomrule
    \end{tabular}}
    \caption{Examples from across the six rankings, randomly sampled.}
    \label{fig:ranking_examples_all}
\end{table*}

\begin{figure*}\centering
Thirdly, Figure~\ref{fig:increasing_all} illustrates how training on different subsets of the ranking leads to different performance on the test set. In general, it is better to train on the compositional examples.

\begin{subfigure}[b]{0.67\columnwidth}
    \centering
    \includegraphics[width=0.9\textwidth]{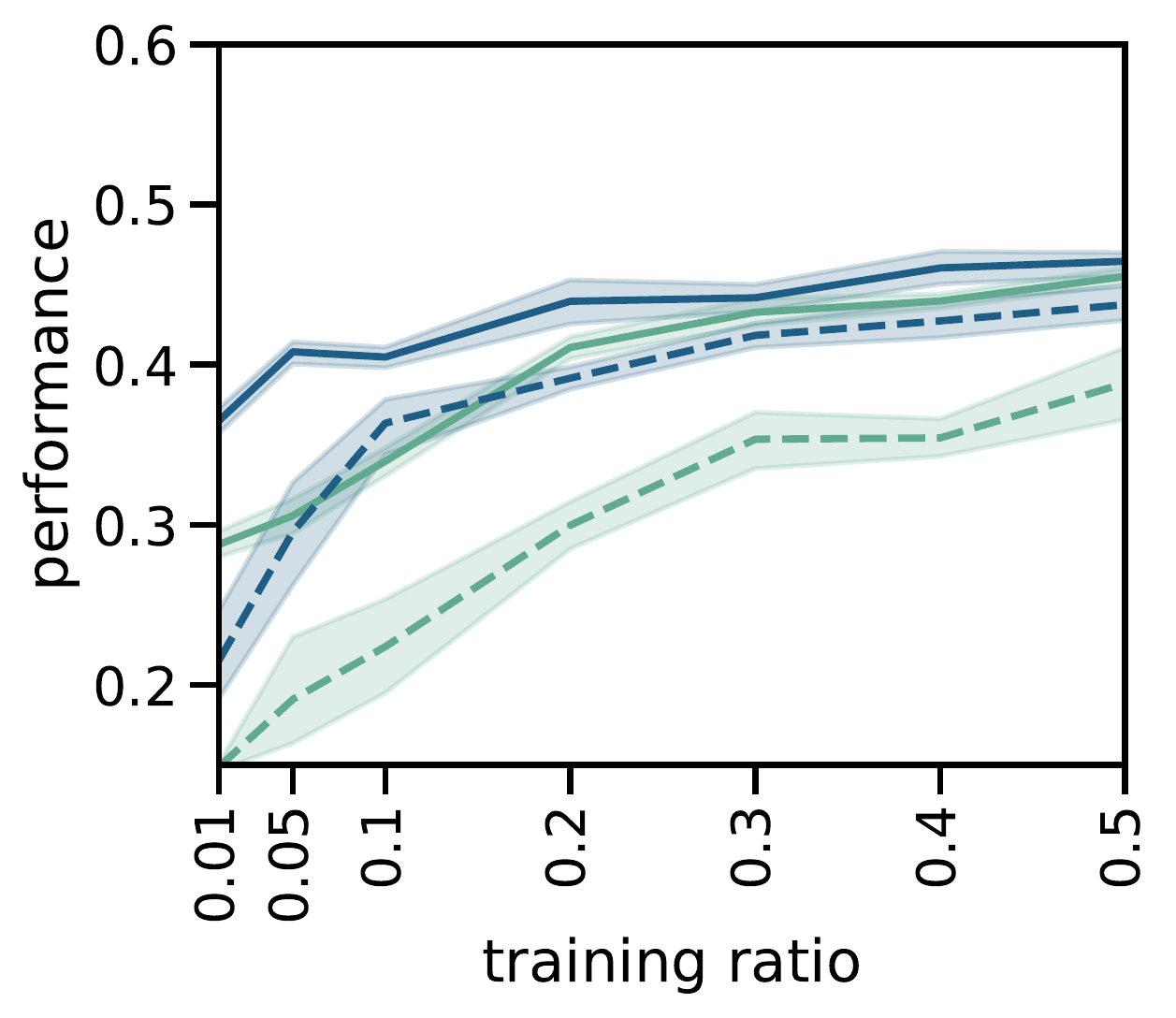}
    \caption{LSTM, Hidden dim., BCM-PP}
\end{subfigure}
\begin{subfigure}[b]{0.67\columnwidth}
    \centering
    \includegraphics[width=0.9\textwidth]{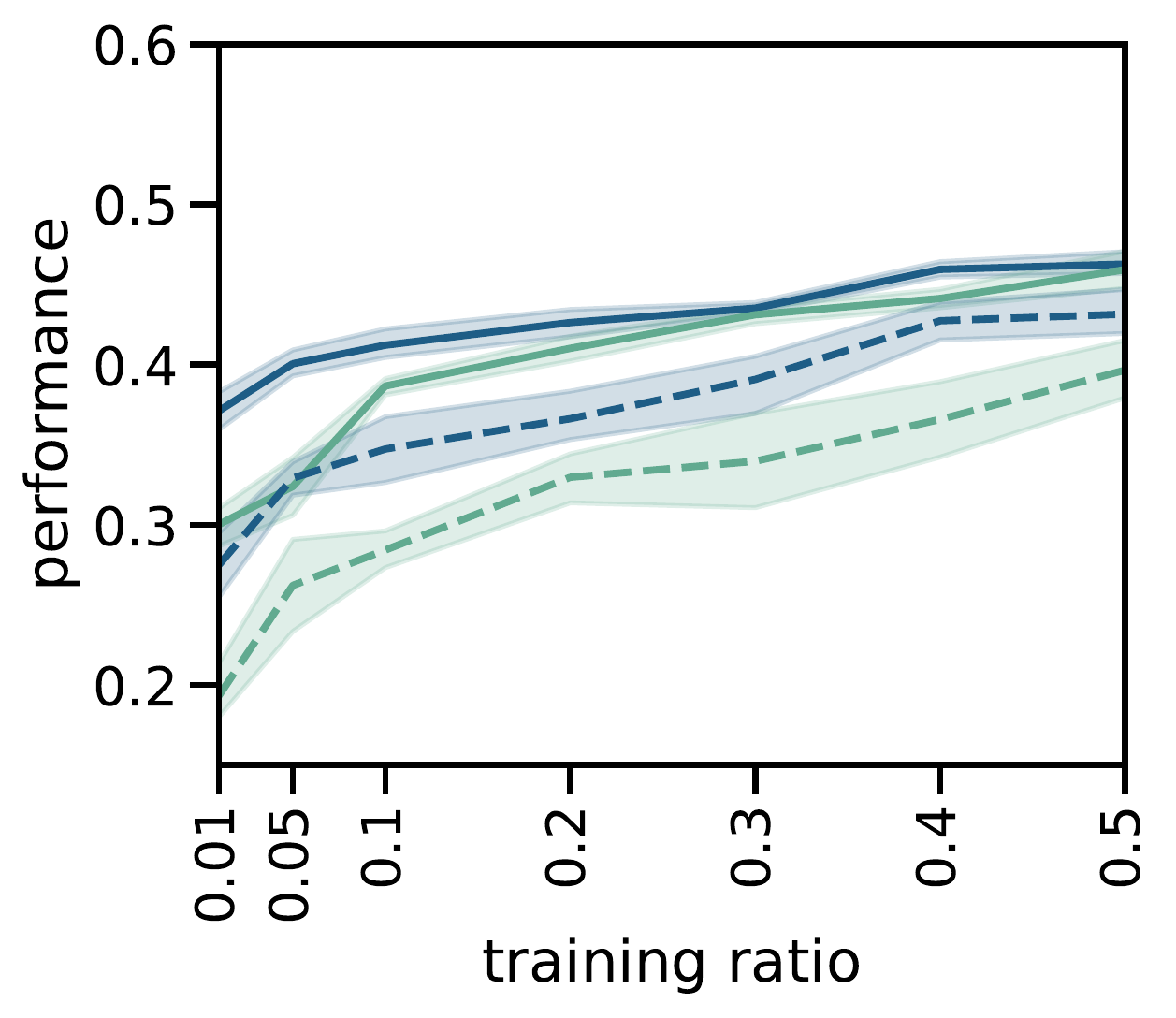}
    \caption{LSTM, dropout, BCM-PP}
\end{subfigure}
\begin{subfigure}[b]{0.67\columnwidth}
    \centering
    \includegraphics[width=0.9\textwidth]{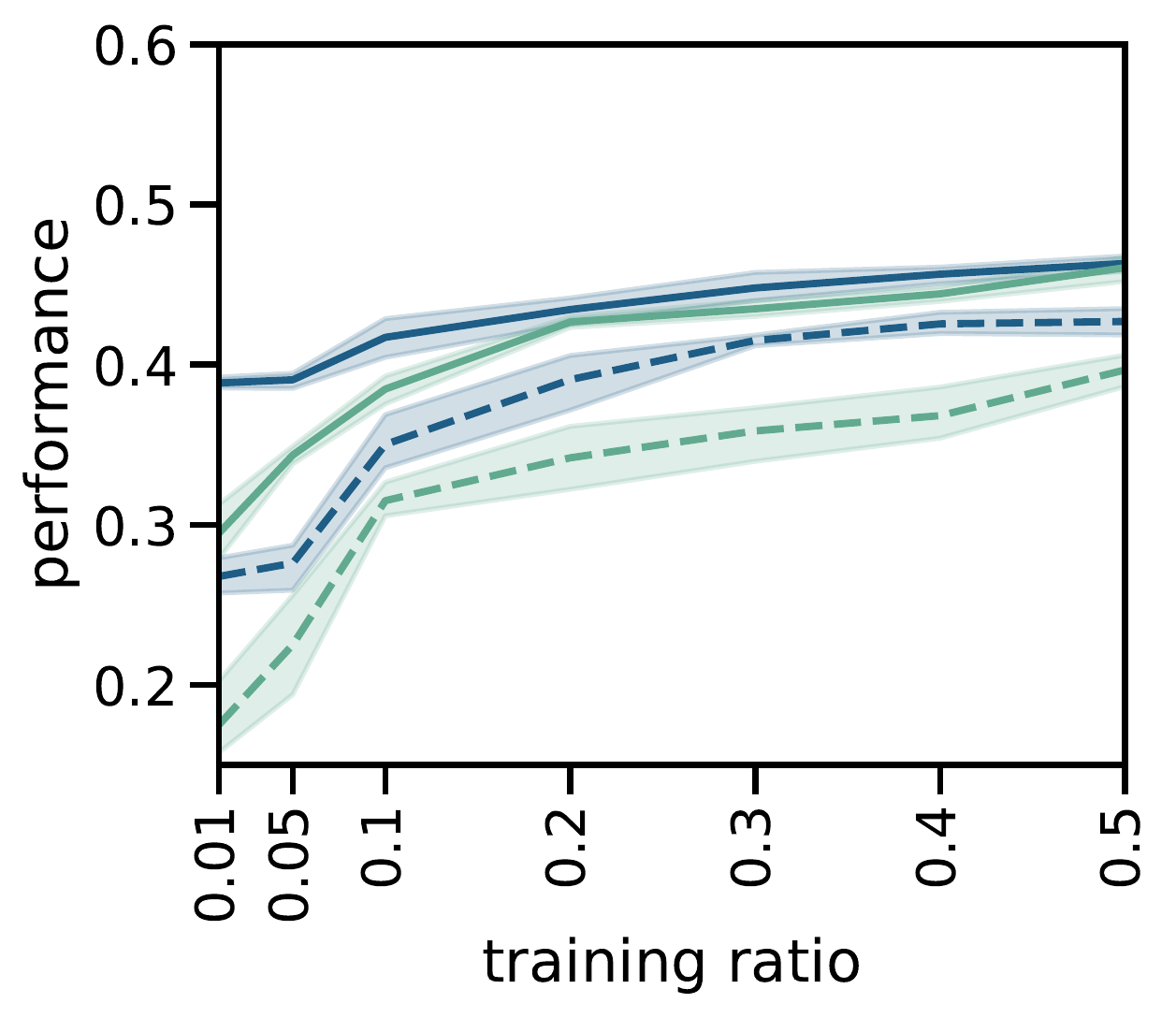}
    \caption{LSTM, DVIB, BCM-PP}
\end{subfigure}
\begin{subfigure}[b]{0.67\columnwidth}
    \centering
    \includegraphics[width=0.9\textwidth]{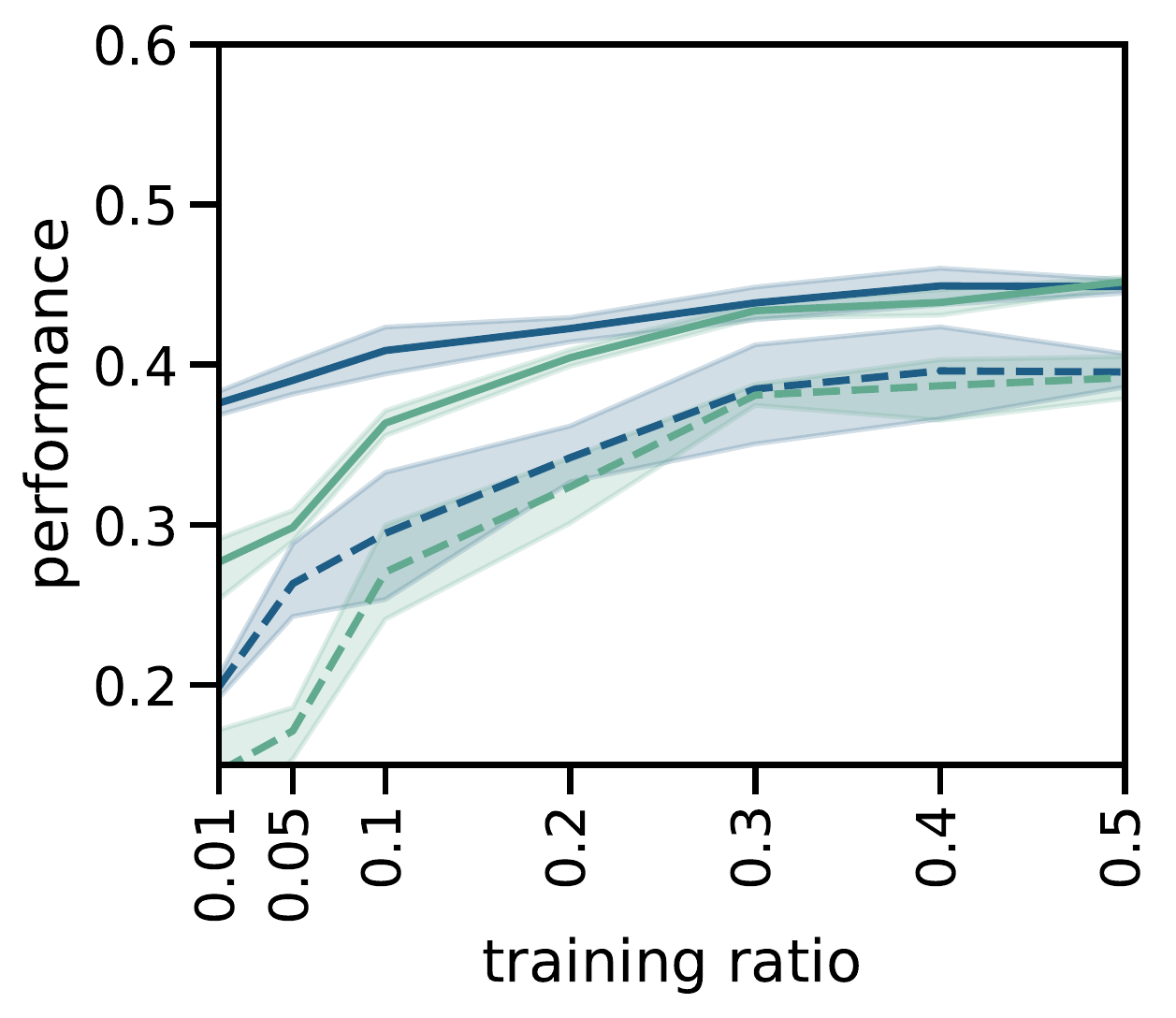}
    \caption{LSTM, Hidden dim., BCM-TT}
\end{subfigure}
\begin{subfigure}[b]{0.67\columnwidth}
    \centering
    \includegraphics[width=0.9\textwidth]{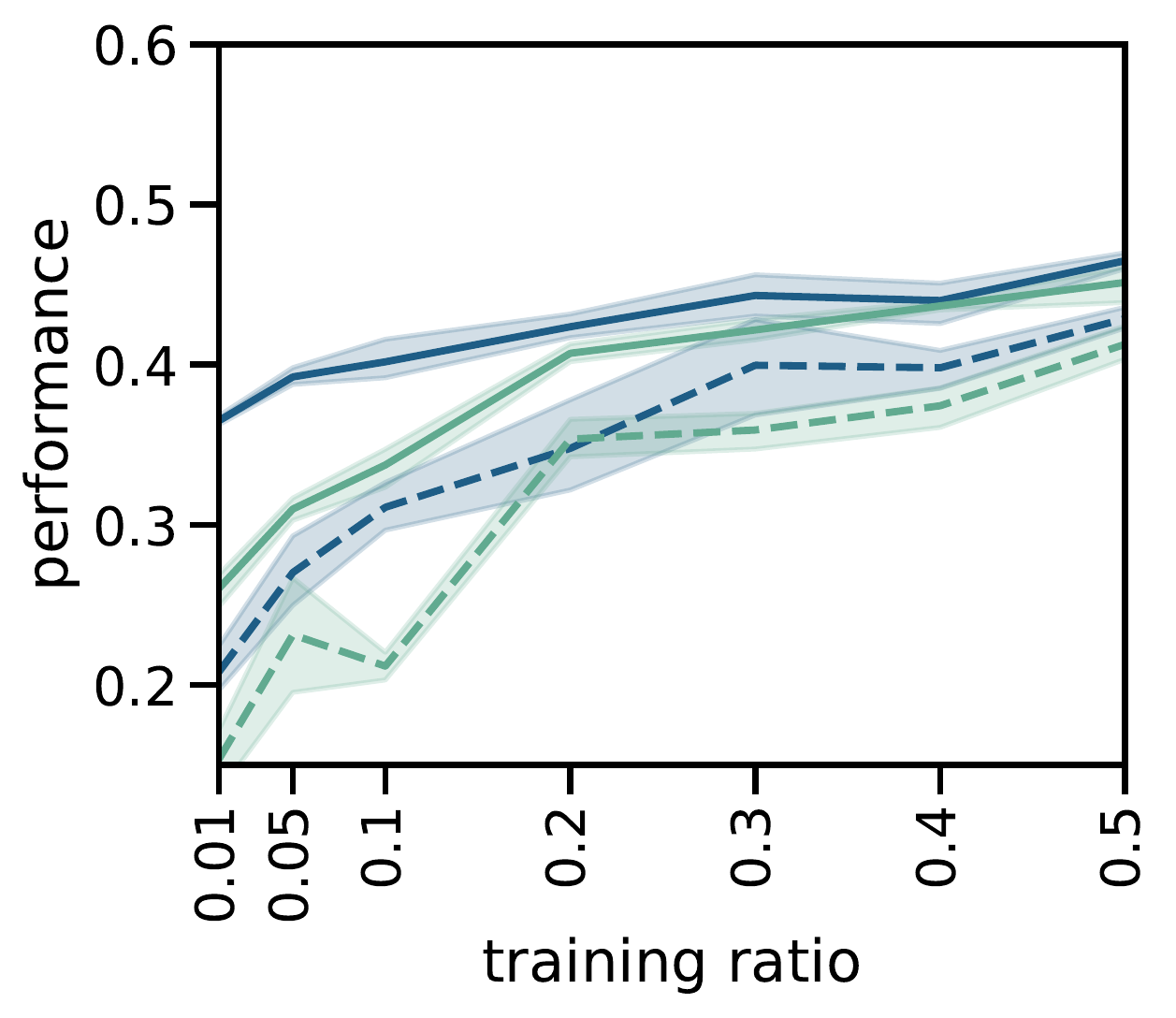}
    \caption{LSTM, dropout, BCM-TT}
\end{subfigure}
\begin{subfigure}[b]{0.67\columnwidth}
    \centering
    \includegraphics[width=0.9\textwidth]{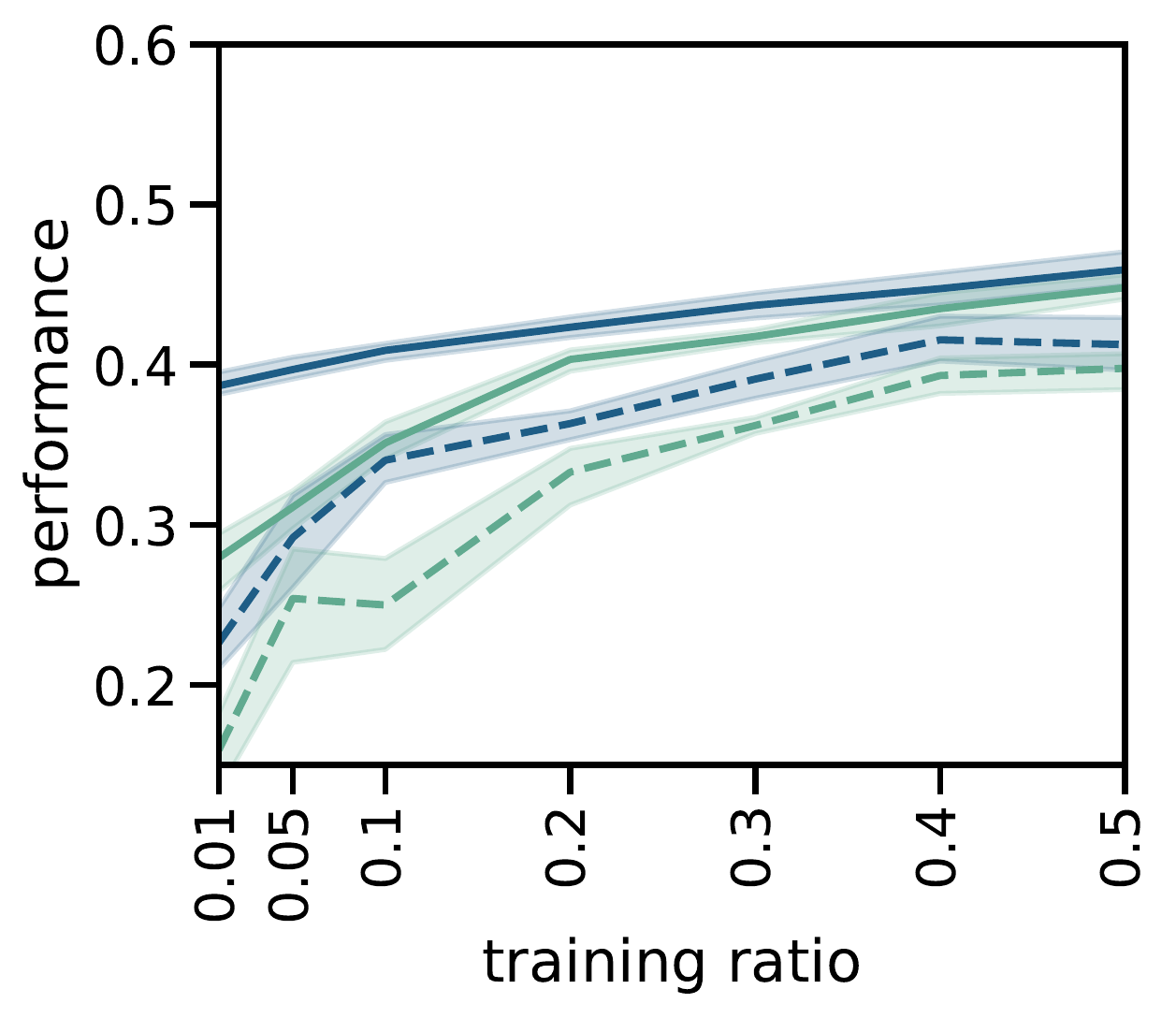}
    \caption{LSTM, DVIB, BCM-TT}
\end{subfigure}

\begin{subfigure}[b]{0.67\columnwidth}
    \centering
    \includegraphics[width=0.9\textwidth]{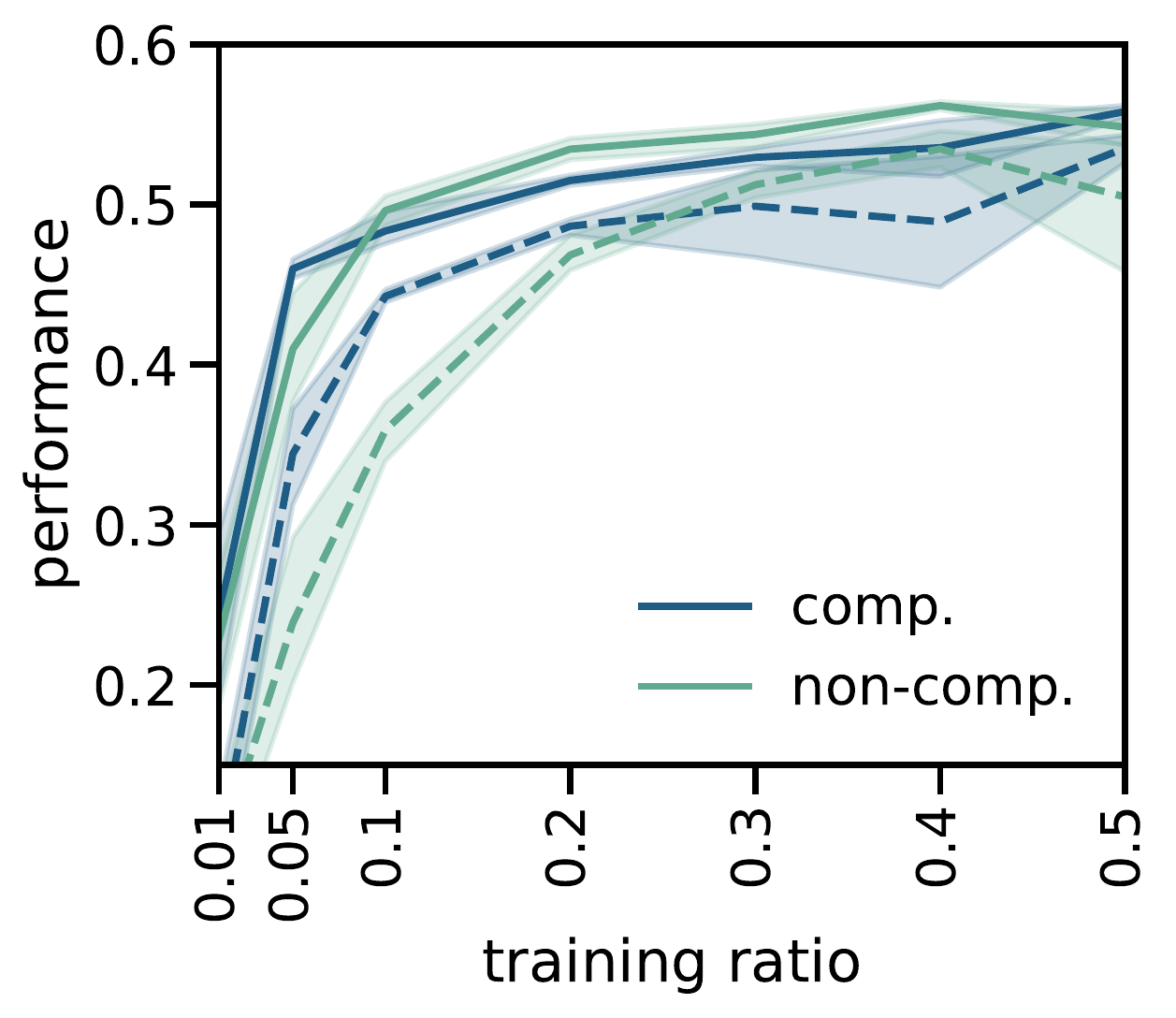}
    \caption{Roberta, Hidden dim., BCM-PP}
\end{subfigure}
\begin{subfigure}[b]{0.67\columnwidth}
    \centering
    \includegraphics[width=0.9\textwidth]{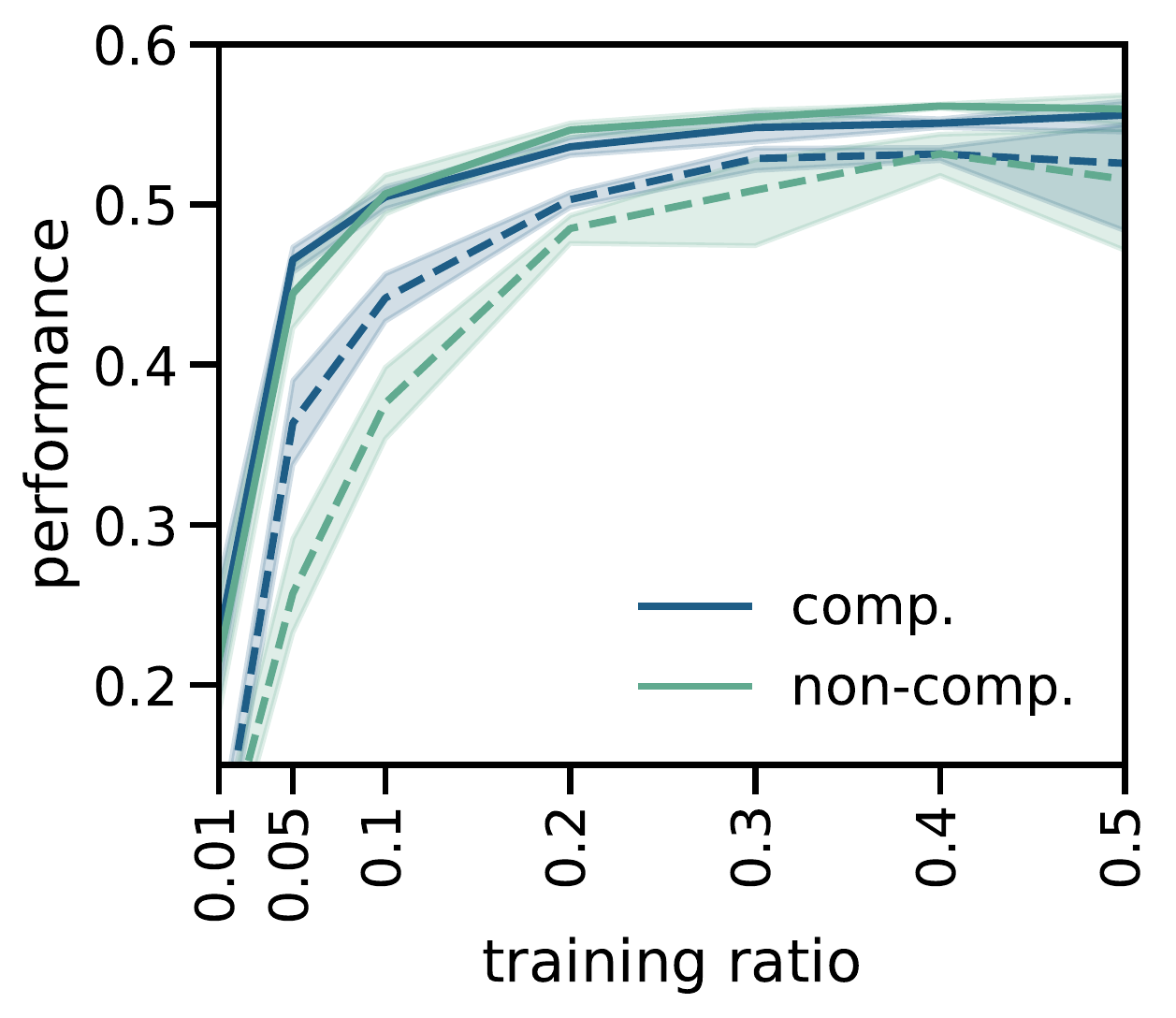}
    \caption{Roberta, dropout, BCM-PP}
\end{subfigure}
\begin{subfigure}[b]{0.67\columnwidth}
    \centering
    \includegraphics[width=0.9\textwidth]{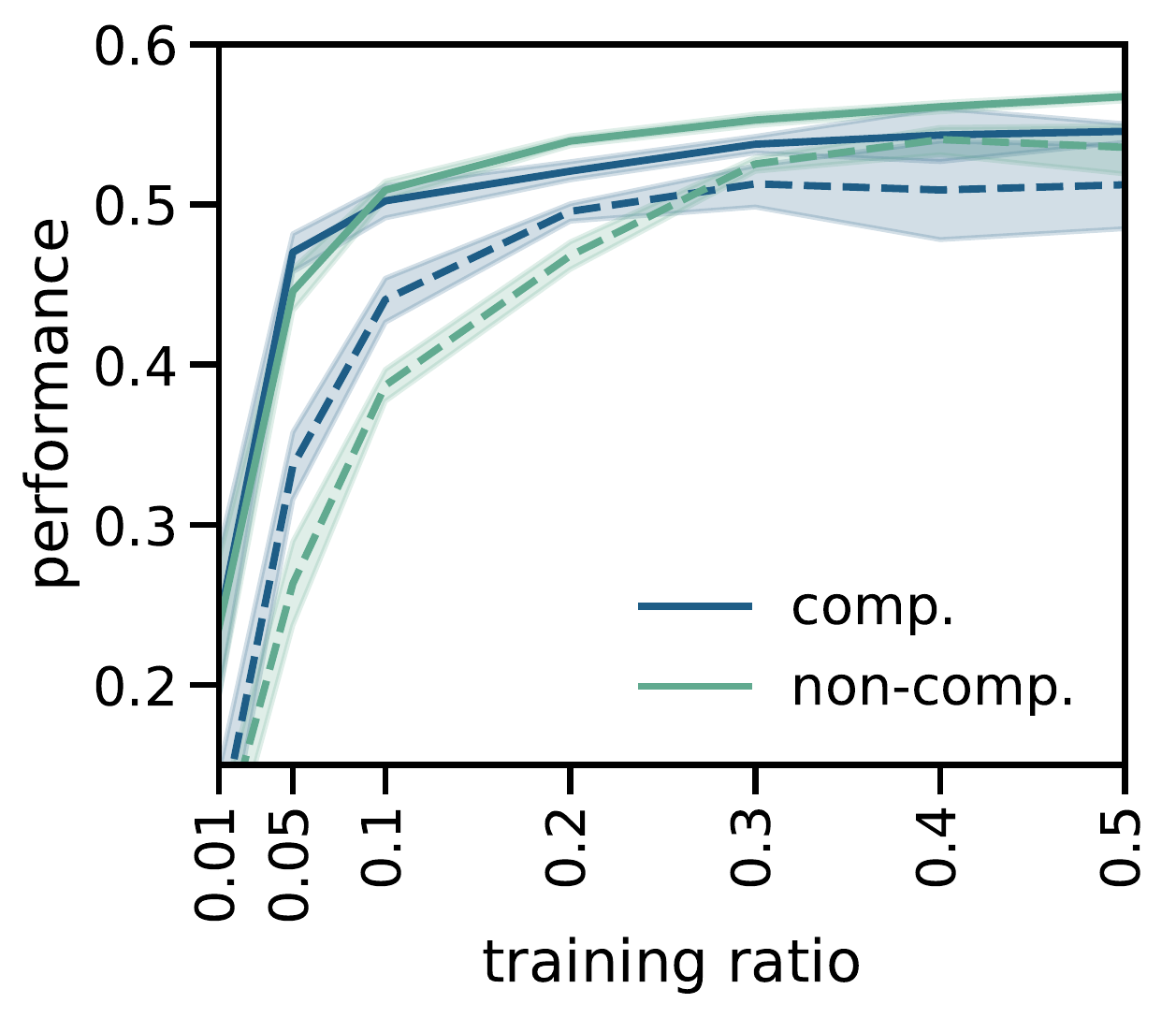}
    \caption{Roberta, DVIB, BCM-PP}
\end{subfigure}
\begin{subfigure}[b]{0.67\columnwidth}
    \centering
    \includegraphics[width=0.9\textwidth]{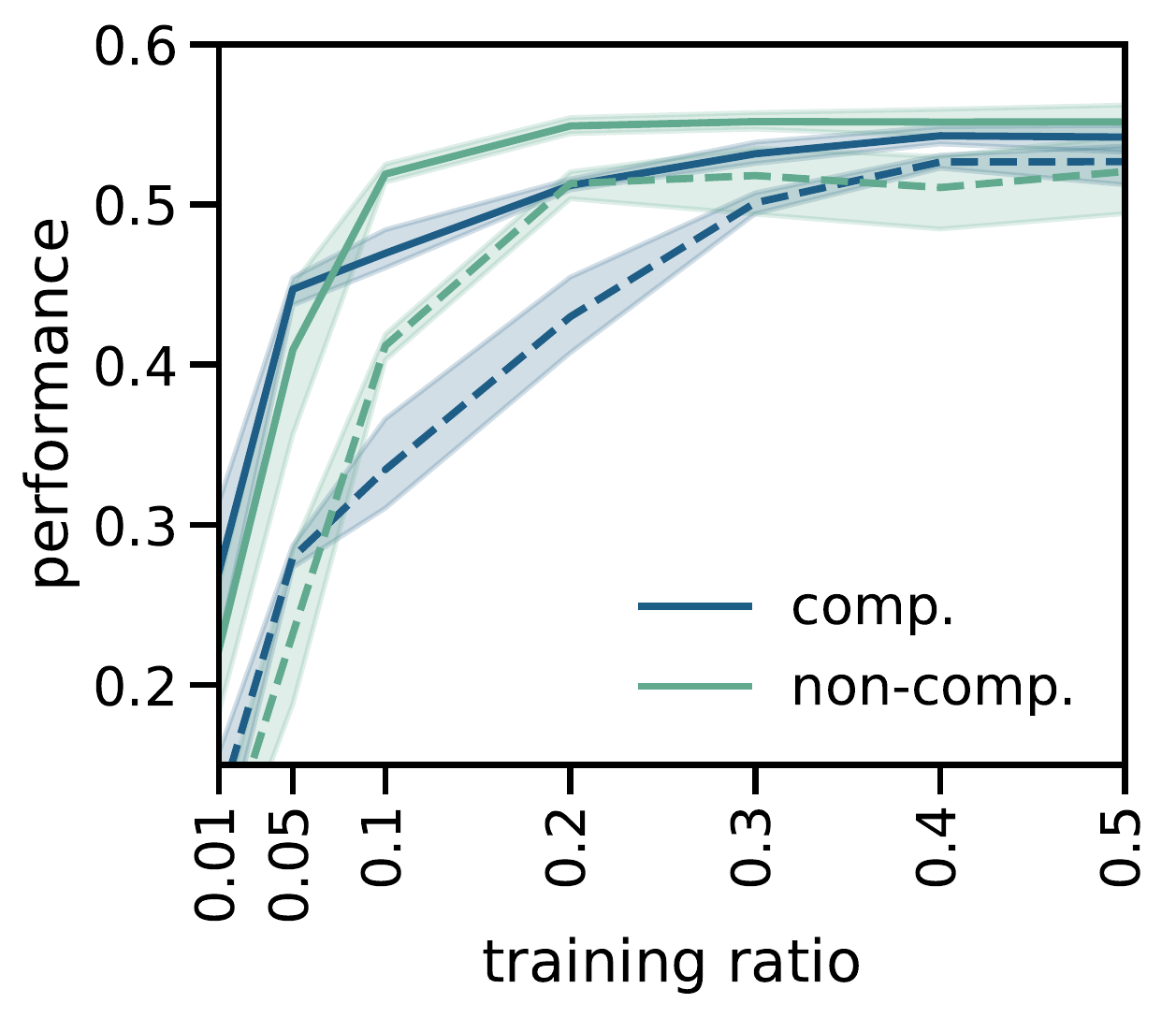}
    \caption{Roberta, hidden dim., BCM-TT}
\end{subfigure}
\begin{subfigure}[b]{0.67\columnwidth}
    \centering
    \includegraphics[width=0.9\textwidth]{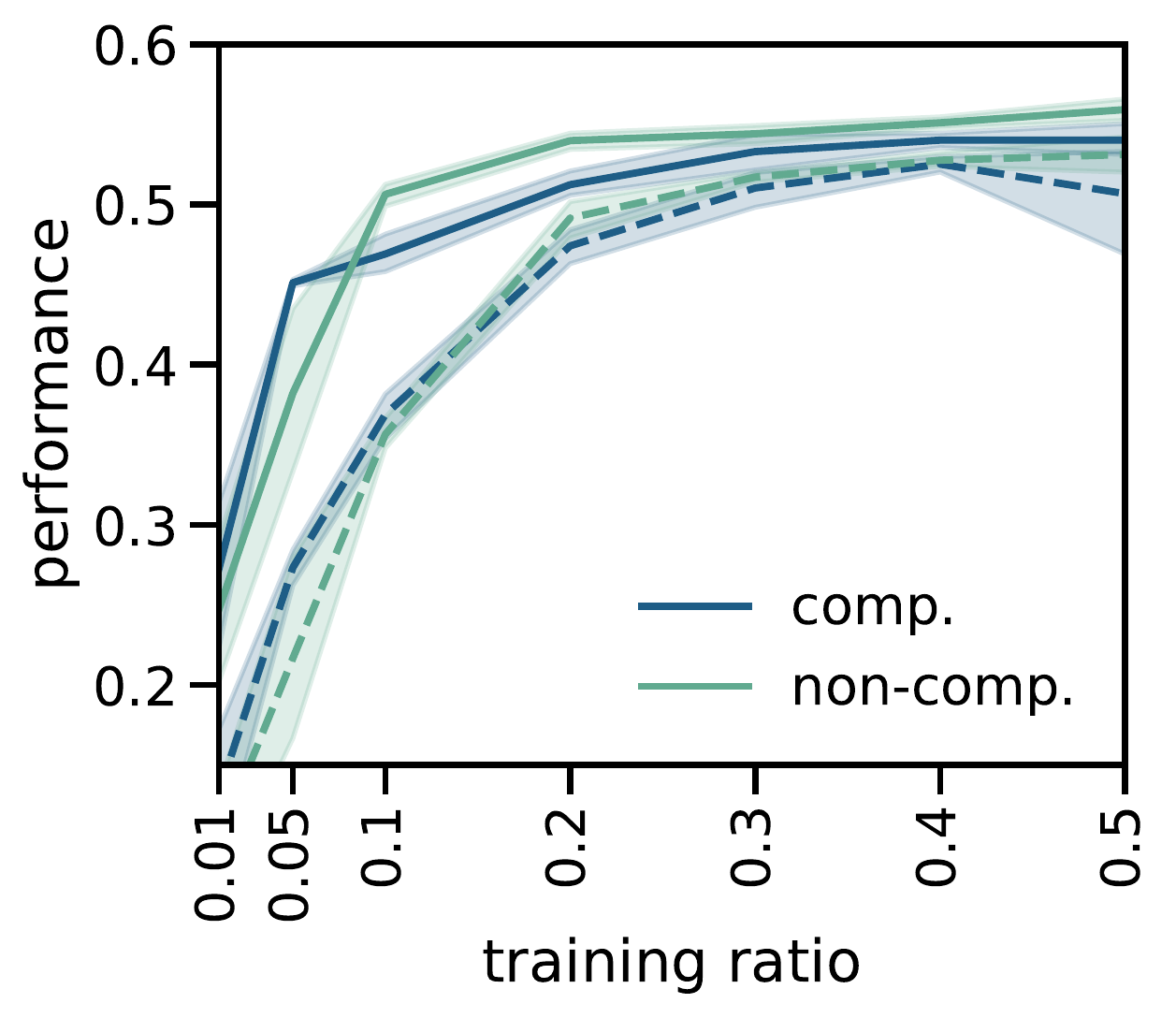}
    \caption{Roberta, dropout, BCM-TT}
\end{subfigure}
\begin{subfigure}[b]{0.67\columnwidth}
    \centering
    \includegraphics[width=0.9\textwidth]{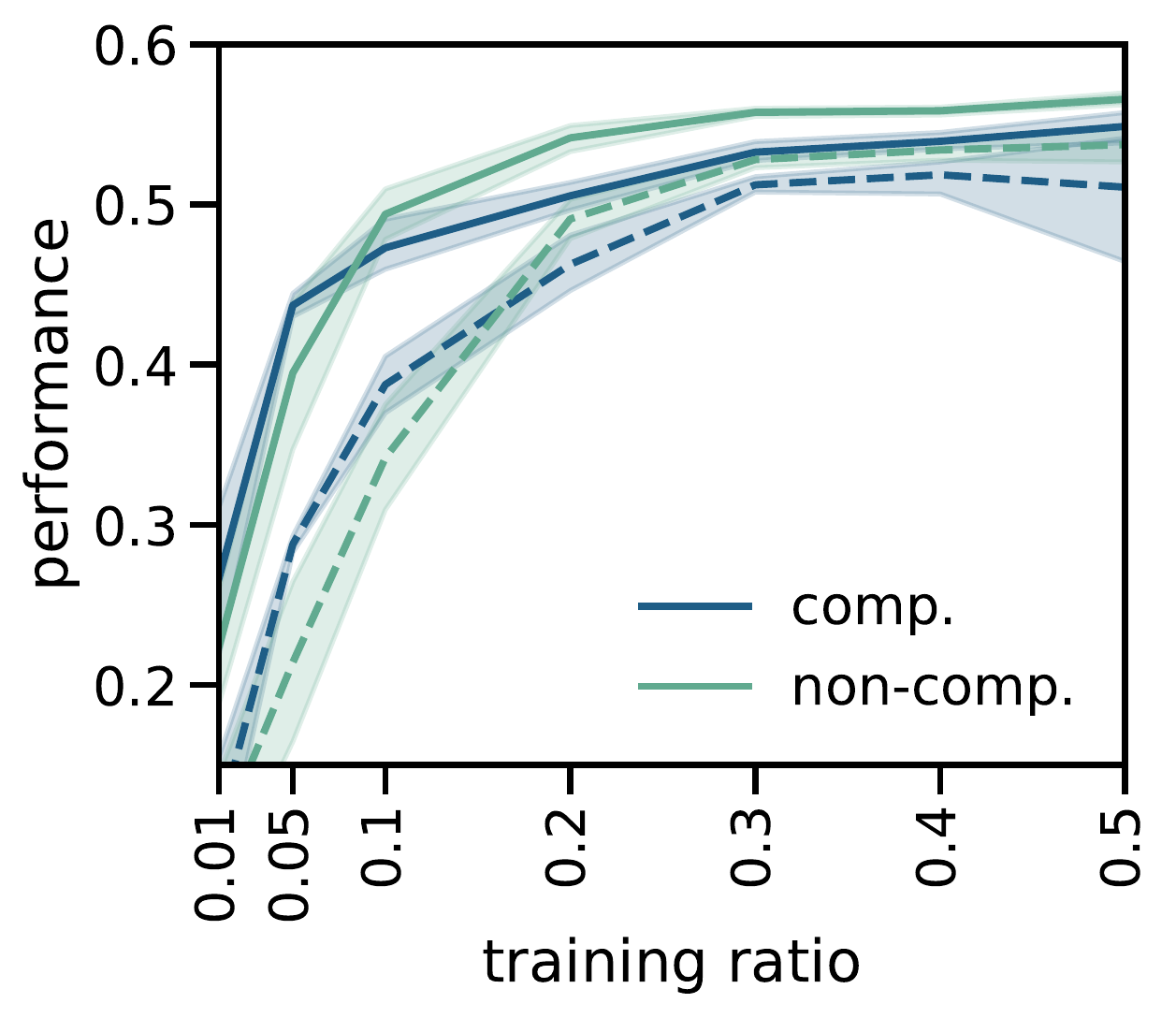}
    \caption{Roberta, DVIB, BCM-TT}
\end{subfigure}
\caption{Change in SST test set accuracy (solid) and macro-averaged $F_1$-score (dashed) as the training set size increases, for LSTM and Roberta models.}
\label{fig:increasing_all}
\vspace{-0.3cm}
\end{figure*}

\clearpage

\section{Reproducibility details}
\label{ap:reproducibility}
Visit \url{https://github.com/vernadankers/bottleneck_compositionality_metric} for the code and data.
Below we collect an overview of the settings involved in the model training and model evaluation:

\begin{itemize}[itemsep=1pt]
    \item \textbf{Model specifications}: Following \citet{tai2015improved} we use 150 hidden dimensions, and 300-dimensional word embeddings for the sentiment analysis task. For consistency, we adopt the 150 dimensions in the arithmetic task, as well, but reduce the size of the word embeddings to 150. The arithmetic base models have 426k trainable parameters, and the sentiment base models have 514k trainable parameters (excluding the frozen word embeddings).
    \item \textbf{Training procedure}: The sentiment and arithmetic models are trained for 10 and 50 epochs, respectively, based on model convergence. These numbers are fixed across models, to ensure that when multiple models are combined in the compositionality metric, they have been trained for the same amount of time. For sentiment, a batch size of 4 was used, and batch sizes \{1, 4, 8\} were experimented with. For arithmetic, batch sizes \{16, 32, 64\} were experimented with across 5 seeds, where 32 was selected. Selection is based on validation performance across five seeds and training speed (e.g. while 1 slightly outperformed 4 for sentiment analysis, we opted for 4 for computational efficiency). For both tasks, learning rates \{0.001, 2e-4, 1e-4\} were experimented with, and 2e-4 was selected based on model performance on the validation set across 5 seeds. In these hyperparameter trials, the accuracy was used for sentiment analysis and the MSE was used for the arithmetic task.
    \item \textbf{Hidden dimensionality bottleneck}: To increase compression, the dimensionality should clearly become smaller, so we manually select 8 values to run ranging from 150 (the standard dimension) to 5.
    \item \textbf{Dropout bottleneck}: For dropout, the closer to 1, the more compression there is. We manually select 8 values to run from 0 (the standard amount) to 0.9.
    \item \textbf{DVIB}: For DVIB, preliminary experiments indicated that $\beta > 1$ for arithmetic or $\beta > 0.1$ for sentiment debilitates the model. We manually selected 8 values to run for $\beta$ based on that information. We use the implementation of \citet{li2019specializing} for the DVIB, available at: \url{https://github.com/XiangLi1999/syntactic-VIB}.
    \item \textbf{Evaluation metrics}: The evaluation metric used for arithmetic is the MSE. The evaluation metrics for sentiment analysis are the accuracy of the predicted sentiment class, and the $F_1$-score, macro-averaged.
    \item \textbf{Number of runs \& run time}: The results for both tasks are averaged over ten seeds. Training one model with one seed on CPU lasts up to 40 minutes for the sentiment analysis models, and up to 30 minutes for the arithmetic task. The TRE-training setup typically takes twice as long. We utilise CPUs from the \href{https://docs.hpc.cam.ac.uk/hpc/user-guide/icelake.html}{\texttt{icelake} partition of the CSD3 cluster}.
\end{itemize}

\noindent For the datasets used, the following are relevant details in terms of their size and preprocessing:
\begin{itemize}[itemsep=1pt]
    \item \textbf{Arithmetic task}: The arithmetic task was generated using the implementation of \citet{hupkes2018visualisation}, available at \url{https://github.com/dieuwkehupkes/processing_arithmetics}. We augment the data with the ambiguous examples ourselves.
    The training data consist of 14903 expressions with 1 to 9 numbers. We test on expressions with lengths 5 to 9, using 5000 examples per length.
    2100 additional examples are used as validation data, to track the model's behaviour during training.
    \item \textbf{Stanford Sentiment Treebank}: We collect the SST data from the \texttt{pytreebank} package (\url{https://github.com/JonathanRaiman/pytreebank}). For the (Tree-)LSTM models, we further preprocess the inputs by lowercasing the sentences. 
    We use SST-5, that classifies sentences using five classes ranging from very negative to very positive. The standard train, validation and test subsets have 8544, 1101 and 2210 examples, respectively.
    Each node in the input trees has its own sentiment label.
\end{itemize}

For the experiments contained in Section~\ref{sec:sentiment_applications}, we did not run an extensive hyperparameter search. Those results are averaged over five seeds, and the models were trained on one NVIDIA A100-SXM-80GB, where training one seed lasted up to 15 minutes.

\end{document}